%% file: 0_main.tex
\documentclass{article} 
\usepackage{iclr2027_conference,times}

\input{math_commands.tex}

\usepackage{duckuments}
\usepackage{booktabs}
\usepackage{multirow}
\usepackage{graphicx}
\usepackage{url}
\usepackage{subcaption}
\usepackage[pagebackref=true]{hyperref}
\usepackage{amsmath}
\usepackage{amssymb}
\usepackage{mathtools}
\usepackage{amsthm}
\usepackage[capitalize,noabbrev,nameinlink]{cleveref}
\usepackage{lipsum}
\usepackage{duckuments}
\usepackage{xifthen}
\usepackage{array}
\usepackage{xparse}
\usepackage{wrapfig}
\usepackage{textcomp, gensymb}
\usepackage[ruled]{algorithm2e}
\usepackage{pifont}
\usepackage[table]{xcolor}
\usepackage{etoc}
\usepackage{enumitem}

\definecolor{red}{HTML}{E6194B}
\definecolor{orange}{HTML}{E0925C}
\definecolor{green}{HTML}{3CB44B}
\definecolor{blue}{HTML}{6474CF}
\definecolor{lightblue}{HTML}{B4BEF0}
\definecolor{purple}{HTML}{9169C2}
\definecolor{lightpurple}{HTML}{D3BEEA}
\definecolor{AffAccent}{HTML}{B4BEF0}
\definecolor{AffMuted}{HTML}{606875}
\definecolor{AffCheck}{HTML}{465365}
\definecolor{AffCross}{HTML}{A1A8B3}

\hypersetup{
    colorlinks=true,
    linkcolor=blue,
    citecolor=blue,
    urlcolor=blue
}

\newif\ifmcp
\mcpfalse

\crefname{appendix}{Appendix}{Appendices}
\Crefname{appendix}{Appendix}{Appendices}
\etocsetstyle{section}
    {\par\addvspace{1.0em}}
    {}
    {\noindent
     \textbf{\etocnumber\quad\etocname}\par}
    {}

\etocsetstyle{subsection}
    {}
    {}
    {\noindent\hspace{2.3em}
     \etocnumber\quad\etocname
     \dotfill
     \etocpage\par}
    {}

\title{
    Blending experimental electron density information to protein-ligand interaction
}

\author{\textbf{Seonghyun Park}$^{1}$ \quad
\textbf{Shuan Chen}$^{2}$ \quad
\textbf{Jun Jin Choong}$^{2}$ \quad
\textbf{Noboru Kuno}$^{2}$ \quad
\textbf{Gyubok Lee}$^{2}$ \\[2pt]
\textbf{Kaushalya Madhawa}$^{2}$ \quad
\textbf{Keisuke Ozawa}$^{2}$ \quad
\textbf{Hikaru Shindo}$^{2}$ \quad
\textbf{Sungsoo Ahn}$^{1}$ \\[5pt]
$^{1}$KAIST\quad$^{2}$SB Intuitions \\[3pt]
\texttt{\{hyun26,sungsoo.ahn\}@kaist.ac.kr}
}

\newcommand{\std}[1]{\ensuremath{\,{\scriptstyle\pm\,#1}}}
\newcommand{\cmark}{\textcolor{green}{\ding{51}}}
\newcommand{\xmark}{\textcolor{red}{\ding{55}}}
\newcommand{\ours}{\textsc{CoDE}}
\newcommand{\oursC}{\textsc{CoE}}
\newcommand{\map}{\textnormal{2mF\textsubscript{o}-DF\textsubscript{c}}}

\iclrfinalcopy 
\begin{document}

\maketitle

\begin{abstract}
    Machine learning models for protein--ligand interactions commonly rely on atomic coordinates, which are constructed by fitting molecular models to experimental electron-density maps. However, most models represent each complex as a single fitted structure, which can omit information about conformational heterogeneity, positional uncertainty, and local density support for individual atoms at binding interfaces. To fill this gap, we present \textbf{\ours{}}, a \textbf{\underline{Co}}ordinate and \textbf{\underline{D}}ensity \textbf{\underline{E}}ncoder, pre-trained on experimental density maps. We represent each binding site on a shared three-dimensional voxel grid combining atomic occupancy, experimental electron density, and its gradient magnitude. Using approximately 100K curated complexes, we pre-train a channel-wise vision transformer to jointly reconstruct masked atomic and density channels, learning a representation that integrates structural and experimental information. 
    We evaluate the frozen encoder on two downstream tasks: (i) binding-affinity regression under leakage-controlled splits and (ii) pocket-conditioned ligand generation. Across these settings, \ours{} improves over an identically pre-trained coordinate-only encoder and remains competitive with task-specific baselines, showing that density complements fitted atomic coordinates for protein--ligand interaction modeling and that the density-blended representation is useful for structure-based ligand generation.
\end{abstract}

\addtocontents{toc}{\protect\setcounter{tocdepth}{-1}}
\input{ICLR2027/1_intro}
\input{ICLR2027/2_related_works}
\input{ICLR2027/3_method}
\input{ICLR2027/4_experiment_results}
\input{ICLR2027/5_conclusion}

\subsection*{AI use statement}


We used generative AI tools, including OpenAI Codex and Claude Code, to assist with data processing, code implementation, model training, and evaluation, preparation of figures and tables, and drafting and polishing text. These tools were used under specific instructions reflecting the authors' ideas and intentions. We did not use generative AI tools to originate the core research hypothesis or methodology, formulate scientific claims, construct conclusions, or construct mathematical proofs. All AI-assisted work was reviewed by the authors: generated code and configurations were checked, experimental outputs and analysis were verified, and figures, tables, and text suggested were reviewed by the authors. The authors take full responsibility for the final content of this work.


\subsection*{Ethics statement}


This work focuses on computational modeling of protein-ligand interactions using structural and experimental electron-density information. As with other computational methods for molecular modeling and drug discovery, the proposed approach could potentially be applied to the design or evaluation of harmful compounds. We therefore encourage responsible use of the method and independent validation of computational predictions before experimental or practical use. All datasets used in this work are subject to their respective licenses and terms of use.

\subsection*{Reproducibility statement}


The code for data processing, pre-training, and downstream evaluation will be publicly released at \url{https://github.com/seonghyun26/CoDE}, with configuration described in the appendices. Additionally, the model architecture, input representation, preprocessing procedures, training settings, evaluation protocols, and hyperparameters used for the experiments are reported in the appendices. We specify all evaluation procedures used for downstream tasks. Reproducing experiments may additionally require access to the corresponding original data sources and compliance with their respective licensing conditions.



\bibliography{ref}
\bibliographystyle{iclr2027_conference}

\appendix
\crefalias{section}{appendix}
\crefalias{subsection}{appendix}
\etocdepthtag.toc{appendix}

\newpage
\etocsettagdepth{appendix}{subsection}
\renewcommand{\contentsname}{Appendix}
\tableofcontents

\newpage
\input{ICLR2027/appx/1_data}
\newpage
\input{ICLR2027/appx/2_method}
\newpage
\input{ICLR2027/appx/3_evaluation}
\newpage
\input{ICLR2027/appx/4_additional_result}

\end{document}

%% file: math_commands.tex
\usepackage{amsmath,amsfonts,bm}

\def\eqref#1{equation~\ref{#1}}

\def\1{\bm{1}}

\DeclareMathAlphabet{\mathsfit}{\encodingdefault}{\sfdefault}{m}{sl}
\SetMathAlphabet{\mathsfit}{bold}{\encodingdefault}{\sfdefault}{bx}{n}



%% file: ICLR2027/1_intro.tex
\section{Introduction}
\label{sec:intro}

Modern machine learning for proteins primarily relies on amino acid sequences and three-dimensional atomic coordinates. Sequence models learn evolutionary and functional patterns~\citep{rives2021biological,lin2023evolutionary}, while structure-based models capture molecular geometry and spatial interactions from atomic coordinates~\citep{jumper2021highly,abramson2024accurate}. Their widespread use is supported by large-scale resources such as UniProt~\citep{uniprot2019uniprot} and the Protein Data Bank~\citep{berman2000protein}. However, these models often use only a single set of coordinates fitted to X-ray diffraction data, which does not convey how strongly the data support each fitted position or whether alternative conformations are present. This gap is consequential at protein--ligand interfaces, where binding depends on local structure.

Crystallographic electron-density maps offer a complementary view of the same complex, reflecting how well the data constrain each atom. A \map{} map combines observed diffraction amplitudes with model-dependent phase estimates and weights~\citep{read1986improved}. Its local pattern can reflect occupancy, atomic displacement, and unresolved alternative conformations, alongside resolution and model error~\citep{kuzmanic2014xray,lang2014protein}. For instance, greater atomic displacement or disorder, reflected in a higher B factor, spreads the local density over a larger volume \citep{sun2019utility}. The density map at a protein--ligand interface may therefore help a model interpret both a proposed contact and the structural ambiguity around it~\citep{van2018qfit}.

Yet, experimental electron-density maps remain underused for learning protein--ligand interactions, while established applications mostly concern crystallographic structure determination \citep{kowiel2019automatic}. Existing molecular models often construct density-like inputs from coordinates through occupancy grids~\citep{ragoza2022generating,pinheiro2024structure} or quantum-chemical calculations~\citep{isert2024exploring,lin2025tokenizing}, without incorporating additional crystallographic observations. Recent work evaluates electron density maps as alternatives to atom-based inputs for affinity prediction~\citep{suriana2025beyond}. It remains unclear whether experimental electron-density maps complement the fitted coordinates in representing protein--ligand interactions.

\textbf{Contribution.} To fill this gap, we empirically examine the value of experimental electron density as a complementary input for modeling protein--ligand interactions. We present \ours{}, a \textbf{\underline{Co}}ordinate and \textbf{\underline{D}}ensity \textbf{\underline{E}}ncoder, pre-trained by jointly reconstructing masked atomic occupancy, density, and density gradient magnitude on a shared voxel grid, encouraging the representation to encode structural ambiguity or disorder. Curated PLINDER complexes~\citep{durairaj2024plinder} paired with experimental \map{} maps support pre-training, with a matched coordinate-only encoder and channel ablations providing controlled comparisons.
We evaluate the frozen representations on two downstream tasks: (i) binding-affinity regression, which probes protein--ligand interaction modeling, and (ii) pocket-conditioned ligand generation, a representative structure-based drug-design task.
Both tasks use the reference's experimental density map for the target of interest, allowing us to assess the utility of the representations in scenarios where crystallographic data are available.

\input{ICLR2027/resource/figure/method-overview}

In our experiments, we make two observations:
\begin{itemize}[leftmargin=*]
    \item Electron-density map information is indeed predictive of protein--ligand interactions. \ours{} improves over the matched coordinate-only encoder, CoE, in binding-affinity regression and pocket-conditioned ligand generation. When used to condition VoxBind, it also improves docking scores over the original generator and remains competitive with task-specific baselines, while its molecule-quality metrics stay close to VoxBind. Additionally, it produces more consistent local geometry, with gains growing on large rigid fragments.
    \item The benefit of density is associated with structural disorder. Without being trained to predict B-factors, the frozen \ours{} encoder more accurately decodes these values at the complex, residue, and atom levels than CoE. Its affinity-ranking advantage over CoE is larger in higher complex contact B-factors, consistent with density helping most where coordinates are less certain.
\end{itemize}

%% file: ICLR2027/resource/figure/method-overview.tex
\begin{figure}[t]
\centering
    \includegraphics[width=.98\textwidth]{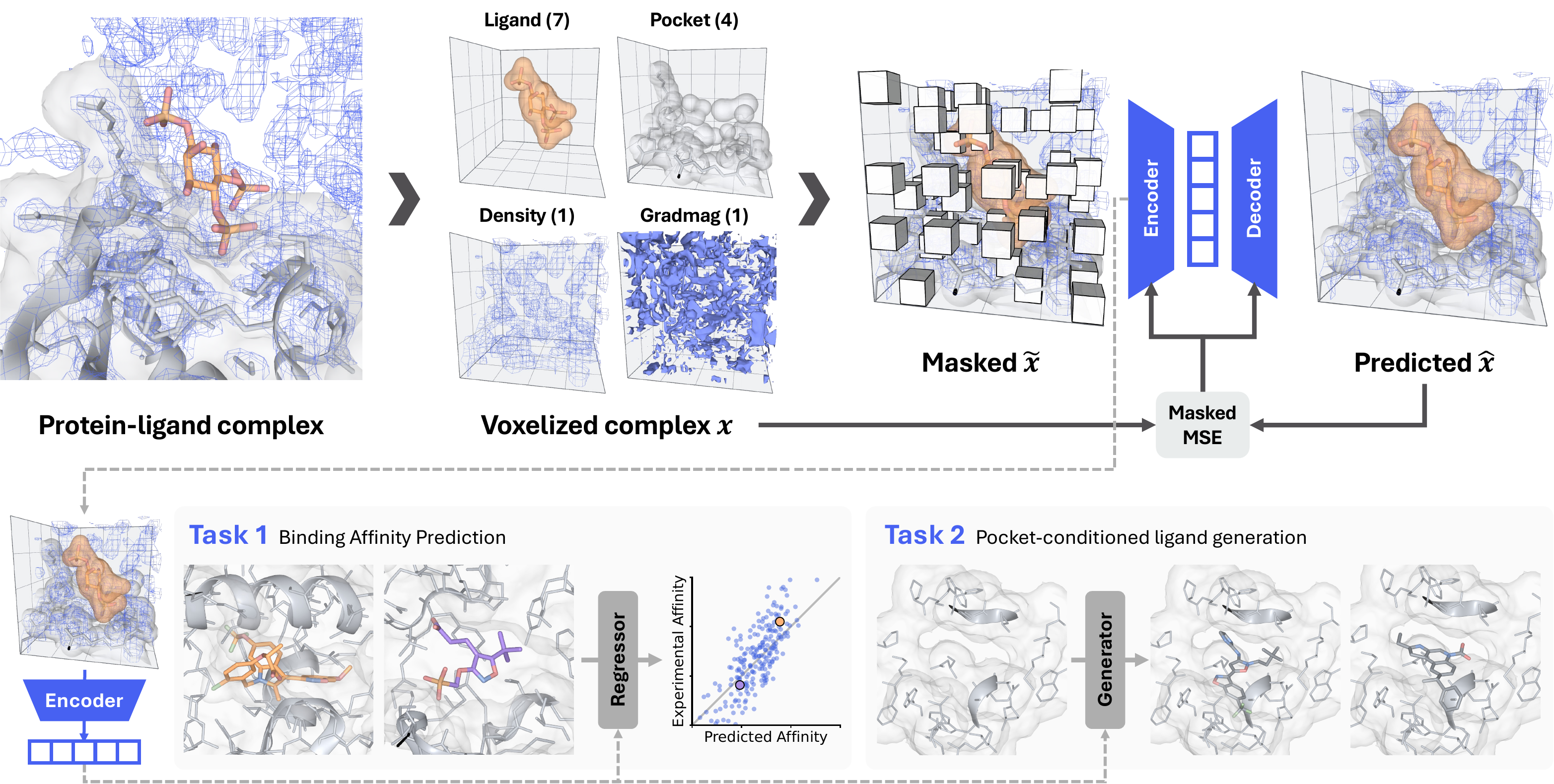}
    \caption{
        \textbf{Overview of pre-training and downstream transfers.}
        Each protein--ligand complex is voxelized into 13 channels: seven ligand and four pocket heavy-atom occupancy channels, plus experimental density and its gradient magnitude. A Channel-ViT is trained to reconstruct masked regions. For downstream transfer, the frozen encoder is used with a regressor for binding-affinity regression, or fused with a generator for pocket-conditioned ligand generation.
    }
    \label{fig:method-overview}
    \vspace{-.1in}
\end{figure}

%% file: ICLR2027/2_related_works.tex
\section{Related Works}
\label{sec:related-work}

\textbf{Experimental electron-density maps.} 
An experimental electron-density map is reconstructed from the diffraction intensities of an X-ray experiment, conventionally as a likelihood-weighted \map{} map \citep{read1986improved}. Since the experiment measures no phases, the map combines the observed amplitudes \textnormal{F\textsubscript{o}} with model-derived amplitudes \textnormal{F\textsubscript{c}} and phases, weighted by the figure of merit m and the model error D. Because it retains \textnormal{F\textsubscript{o}}, the map preserves evidence of partial occupancy and alternative conformations that the fitted model omits \citep{lang2010automated,lang2014protein}. Prior works use such maps mainly to support the fitting itself, e.g., ligand identification \citep{kowiel2019automatic,karolczak2025ligand} and local model assessment \citep{miyaguchi2021machine}. Only recently, maps have been used as an input for affinity prediction \citep{suriana2025beyond}, attempting to replace atomic inputs rather than complementing them. \citet{lin2025tokenizing} shares a similar approach, tokenizing electron clouds computed from coordinates by quantum chemistry and using hierarchical codebooks. In contrast, we learn directly from experimentally measured maps.

\textbf{Binding affinity regression.}
Binding-affinity regression predicts the interaction strength of a protein--ligand complex from experimentally measured binding constants \citep{wang2005pdbbind}, with methods differing in the structural detail they consume. \textit{Sequence}-based models use a ligand SMILES and protein sequences without 3D geometry \citep{ozturk2018deepdta,huang2021moltrans,shenoy2026nesso1}, whereas \textit{structure}-based models use atomic coordinates through voxel CNNs~\citep{stepniewska2018development}, graph neural networks \citep{valsson2025narrowing,kong2026hbgsa}, or equivariant architectures \citep{kong2023generalist,huang2024protein}. Since labeled data are limited, self-supervised methods pre-train with denoising \citep{jin2023dsmbind} or interaction-based objectives \citep{feng2024protein,gao2024self}. Since data are scarce and redundant, which distorts evaluation, leakage-controlled splits and benchmarks have been proposed, removing test complexes resembling the training set \citep{graber2025resolving,li2026leak,wei2026honestaffinity}.

\textbf{Pocket-conditioned ligand generation.}
Structure-based drug design uses target structures to identify or optimize ligands \citep{blundell1996structure,anderson2003process}. Recent pocket-conditioned generators build 3D ligands directly inside a given target pocket. Autoregressive models place atoms or fragments sequentially \citep{luo20213d,peng2022pocket2mol}, diffusion models denoise atom types and coordinates \citep{guan20233d-5e4,schneuing2024structure}, and voxel models denoise a continuous atom occupancy grid \citep{pinheiro20233d-81b,pinheiro2024structure}. A separate line of approaches supplies the generator with extra information about the target site; DecompDiff uses arm and scaffold priors estimated from a reference ligand \citep{guan2024decompdiff}, while others use ligand-shaped electron density and grow molecules into it \citep{wang2022pocket,li2025electron}. Whereas \citet{wang2022pocket} use experimental density maps for task-specific ligand generation, we ask whether a representation pre-trained jointly on coordinates and density transfers to both binding-affinity regression and ligand generation.

%% file: ICLR2027/3_method.tex
\section{Method}
\label{sec:method}

\ours{} represents a protein--ligand complex as a voxel grid combining atomic occupancy with experimental electron density. We pre-train an encoder on this grid by masked reconstruction and transfer its frozen representation to regression and generation tasks. \Cref{fig:method-overview} illustrates the full pipeline.

\textbf{Experimental electron-density map.}
Our input signal is the experimental electron-density map used to refine each X-ray crystal structure.
For each complex in PLINDER \citep{durairaj2024plinder}, we fetch its \map{} map through PDBe \citep{armstrong2020pdbe} by the PDB ID and normalize it. Though the map is phased with the deposited model, it retains the observed amplitudes and thus evidence of what the model omits. To ensure the ligand is resolved in the map, we keep complexes with a ligand real-space correlation coefficient \citep[RSCC]{branden1990between,jones1991improved} of at least $0.8$ \citep{fusani2020exploring}. We also remove every downstream validation and test complex to prevent leakage. Further details are in \cref{appx:data}.

\textbf{Voxel representation of protein--ligand complexes.}
To place atoms and density in a common frame, we build a binding-site grid centered at the ligand centroid~\citep{pinheiro2024structure}. The grid has $G^{3}$ voxels with $G=64$ at $0.25$\,\AA{} spacing, covering a $16$\,\AA{} cube. It contains the ligand heavy atoms and the protein heavy atoms only within $10$\,\AA{} of the ligand, cropped to the cube. Each complex then becomes a $13$-channel tensor $\mathbf{x}$ as follows:
\begin{equation}
    \mathbf{x}
    =
    \operatorname{Concat}(\mathbf{a}, \rho,\lVert\nabla\rho\rVert)
    \in\mathbb{R}^{G^3\times13},
    \quad
    a_{v,c}
    =
    1-\prod_{i\in\mathcal{A}_c}(1-g_{v,i}),
    \quad
    g_{v,i}
    =
    \exp\left(
    -\frac{\Vert\mathbf{r}_v-\mathbf{r}_i\Vert^{2}}
    {(0.93R_{i})^{2}}
    \right)\,.
\end{equation}
The occupancy $\mathbf{a}\in\mathbb{R}^{G^3 \times 11}$ has seven ligand (C, O, N, S, F, Cl, P) and four pocket (C, O, N, S) atom-type channels. Each channel $c$ is the combination of Gaussian blobs $g_{v,i}$ over its heavy atoms $i\in\mathcal{A}_c$ \citep{orlando2022pyuul}. A blob $g_{v,i}$ decays with the distance from the atom center $\mathbf{r}_{i}$ to the voxel center $\mathbf{r}_{v}$, at a width set by the element-specific van der Waals radius $R_{i}$. The density channel $\rho$ is the map tri-linearly interpolated at each voxel center. Its gradient magnitude $\lVert\nabla\rho\rVert$ is computed on the same grid \citep{suriana2025beyond}. Further details are in \cref{appx:data}.

\textbf{Masked pre-training.}
We then learn representations of the grid without labels, following masked image modeling \citep{he2022masked,xie2022simmim}. A Channel-ViT \citep{bao2023channel-24a} encoder $f_{\theta}$ splits the grid into $(G/P)^3$ blocks of $P^{3}$ voxels, with $P=8$ corresponding to a $2$\,\AA{} cube. We mask $75\%$ of the blocks and broadcast the mask to a voxel-level mask $\mathbf{M}\in\{0,1\}^{G^{3}}$. The encoder only sees the visible input $\widetilde{\mathbf{x}}=\mathbf{x}\odot(1-\mathbf{M})$, with masked blocks zeroed rather than dropped. A lightweight reconstruction MLP head $h_{\psi}$ then reconstructs all channels with a loss $\mathcal{L}$ as follows:
\begin{equation}
    \mathcal{L}(\theta,\psi)=
    \frac{1}{|\mathcal{M}|C}
    \sum_{v\in\mathcal{M}}
    \sum_{c=1}^{C}
    w_{c}\omega_{c,v}
    \left(\widehat{x}_{v,c}-x_{v,c}\right)^{2}\,,
    \qquad
    \widehat{\mathbf{x}}=h_{\psi}\!\left(f_\theta(\widetilde{\mathbf{x}})\right)
    \,,
\end{equation}
where $\mathcal{M}=\{v:M_{v}=1\}$ denotes the masked voxels. The channel weights $w_{c}$ balance sparse element frequencies and density channels, while the voxel weight $\omega_{c,v}$ emphasizes occupied voxels. Additionally, we augment the entire grid by random rotation for robustness \citep{pinheiro2024structure}.

\input{ICLR2027/resource/table/result-binding-affinity}

\textbf{Downstream task transfer.}
After pre-training, we freeze the encoder $f_{\theta}$ including its voxel projection, and discard the reconstruction head $h_{\psi}$. For affinity regression, we mean-pool the final patch tokens and fit an MLP regressor on them. For pocket-conditioned ligand generation, the voxel generator is unchanged and fuses the encoder's voxel output with VoxBind's features through a zero-initialized convolution, before the denoising U-Net. Only the generator and this fusion layer are trained, while the encoder stays frozen. For further details, refer to \cref{appx:method}.


%% file: ICLR2027/resource/table/result-binding-affinity.tex
\begin{table}[!t]
    \centering
    \caption{
        \textbf{Binding-affinity prediction on LP-PDBBind \textit{test} and \textit{CL3-filtered test} sets.} Seq. and Struc. denote sequence and structure inputs; structural inputs include coordinates and their chemical features. Results are averaged over five seeds. Affinity-leaked cases are marked with $\dagger$, included for context only. \textbf{Bold} and \underline{underlining} indicate the best and second-best results; overlapping intervals are marked together. N/A denotes unavailable metrics, and \textbf{Pre.} denotes pre-training.
    }
    \label{tab:result-binding-affinity}
    \setlength{\tabcolsep}{4.5pt}
    \renewcommand{\arraystretch}{1.08}
    \setlength{\heavyrulewidth}{1pt}
    \setlength{\lightrulewidth}{0.4pt}
    \resizebox{.98\textwidth}{!}{%
        \begin{tabular}{llccccccc}
            \toprule
            & & & \multicolumn{3}{c}{\textbf{Test (N=1320)}} & \multicolumn{3}{c}{\textbf{CL3 filtered Test (N=733)}} \\
            \cmidrule(lr){4-6}\cmidrule(lr){7-9}
            \textbf{Input} & \textbf{Method} & \textbf{Pre.} & \textbf{Pearson \textit{r} $\uparrow$} & \textbf{Spearman $\rho$ $\uparrow$} & \textbf{RMSE $\downarrow$} & \textbf{Pearson \textit{r} $\uparrow$} & \textbf{Spearman $\rho$ $\uparrow$} & \textbf{RMSE $\downarrow$} \\
            \midrule
            \multirow{4}{*}{Seq.}
            & DeepDTA & \xmark & 0.535\std{0.022} & 0.500\std{0.023} & 1.510\std{0.026} & 0.513\std{0.026} & 0.481\std{0.022} & 1.505\std{0.028} \\
            & MolTrans & \xmark & 0.494\std{0.030} & 0.461\std{0.026} & 1.567\std{0.056} & 0.415\std{0.131} & 0.404\std{0.116} & 1.585\std{0.076} \\
            & HonestAffinity & \cmark & 0.472\std{0.040} & 0.459\std{0.046} & 1.582\std{0.034} & 0.466\std{0.030} & 0.436\std{0.030} & 1.563\std{0.034} \\
            & Nesso-1$^\dagger$ & \cmark & 0.670 & 0.663 & 1.410 & 0.673 & 0.659 & 1.367 \\
            \midrule
            \multirow{13}{*}{Struc.}
            & HBGSA & \xmark & 0.567\std{0.016} & 0.542\std{0.022} & 1.703\std{0.191} & 0.568\std{0.015} & 0.548\std{0.011} & 1.586\std{0.201} \\
            & EGNN & \xmark & 0.593\std{0.008} & 0.583\std{0.007} & 1.433\std{0.014} & 0.595\std{0.004} & 0.561\std{0.005} & 1.403\std{0.010} \\
            & EGNN\textsubscript{\scriptsize TargetDiff} & \xmark & 0.595\std{0.007} & 0.584\std{0.007} & 1.430\std{0.013} & 0.590\std{0.003} & 0.556\std{0.005} & 1.410\std{0.004} \\
            & GET & \xmark & 0.593\std{0.006} & 0.587\std{0.006} & 1.430\std{0.008} & 0.594\std{0.003} & 0.559\std{0.002} & 1.400\std{0.006} \\
            & CheapNet & \xmark & \underline{0.628\std{0.003}} & \underline{0.604\std{0.003}} & \underline{1.417\std{0.014}} & 0.592\std{0.024} & 0.551\std{0.024} & 1.462\std{0.041} \\
            & IPNet & \xmark & 0.593\std{0.022} & 0.569\std{0.024} & 1.437\std{0.041} & 0.556\std{0.032} & 0.526\std{0.020} & 1.440\std{0.037} \\
            & IPNet$^\dagger$ & \cmark & 0.586\std{0.006} & 0.554\std{0.006} & 1.477\std{0.019} & 0.584\std{0.024} & 0.533\std{0.028} & 1.433\std{0.026} \\
            & AEV-PLIG & \xmark & 0.552\std{0.017} & 0.527\std{0.015} & 1.593\std{0.032} & 0.515\std{0.016} & 0.476\std{0.020} & 1.574\std{0.031} \\
            & DSMBind & \cmark & 0.308 & 0.363 & N/A & 0.464 & 0.480 & N/A \\
            & BindNet & \cmark & 0.526\std{0.025} & 0.507\std{0.019} & 1.583\std{0.105} & 0.487\std{0.022} & 0.463\std{0.018} & 1.567\std{0.039} \\
            & ProFSA & \cmark & {0.623\std{0.005}} & 0.594\std{0.006} & 1.518\std{0.012} & \underline{0.645\std{0.008}} & \underline{0.609\std{0.009}} & \underline{1.370\std{0.016}} \\
            \cmidrule(lr){2-9}
            \rowcolor{AffAccent!12} \cellcolor{white} & & \xmark & 0.563\std{0.023} & 0.553\std{0.012} & 1.524\std{0.058} & 0.567\std{0.032} & 0.545\std{0.017} & 1.484\std{0.058} \\
            \rowcolor{AffAccent!12} \cellcolor{white} & \multirow{-2}{*}{\oursC{}} & \cmark & {0.593\std{0.009}} & 0.560\std{0.008} & {1.478\std{0.032}} & {0.620\std{0.010}} & 0.572\std{0.011} & {1.401}\std{0.030} \\
            \midrule
            \rowcolor{AffAccent!28} Struc.\,, & & \xmark & 0.567\std{0.017} & 0.557\std{0.009} & 1.500\std{0.042} & 0.570\std{0.022} & 0.545\std{0.013} & 1.457\std{0.043} \\
            \rowcolor{AffAccent!28} Density & \multirow{-2}{*}{\ours{} (Ours)} & \cmark & \textbf{0.654\std{0.005}} & \textbf{0.641\std{0.004}} & \textbf{1.364\std{0.012}} & \textbf{0.667\std{0.005}} & \textbf{0.638\std{0.006}} & \textbf{1.314\std{0.017}} \\
            \bottomrule
        \end{tabular} %
    }
\end{table}

%% file: ICLR2027/4_experiment_results.tex
\section{Experiment results}
\label{sec:exp-results}

We validate the pre-trained representations with downstream tasks on protein--ligand complexes, spanning both discriminative and generative settings: binding affinity regression and pocket-conditioned ligand generation. Both tasks use a reference density map of each target, which is a mild assumption since it exists for targets of interest, even when bound to a different ligand. Throughout, we compare against a coordinate-only variant of \ours{}, coined \oursC{}, that is identical except for the density and its gradient magnitude input, which isolates the contribution of the density information. For details on experiments and additional results, refer to \cref{appx:data,appx:method,appx:baseline,appx:eval,appx:results}.

\subsection{Binding affinity regression}
\label{subsec:exp-affinity}

\textbf{Setup.}
We first use protein--ligand binding affinity regression as a controlled test of whether experimental density adds information beyond atomic coordinates. The goal of this experiment is not to establish a practical affinity predictor, but rather to determine whether density encodes complementary information. We use complex data from PDBBind v2020 \citep{liu2015pdb}, keeping only those with a crystallographic density map and affinity measured as $K_{d}$ or $K_{i}$. We exclude validation and test complex data from the PLINDER-based pre-training set. We adopt the LP-PDBBind split \citep{li2026leak} to reduce leakage from complex-level interaction similarity, and report results on both the full test set and its CL3 subset, restricting evaluation to higher-quality complexes.

\textbf{Evaluation.}
We measure linear and rank agreement by the Pearson ($r$) and Spearman ($\rho$) correlations, and the predictive error by RMSE. Results are averaged over five seeds, except for zero-shot pre-trained scorers with single predictions. Other results come from our own runs under identical protocols: pre-trained encoders use the same frozen-probe protocol as \ours{} with their own probing heads, while the remaining methods are trained end-to-end with their own heads. Some baselines pre-train on data that overlaps with the LP-PDBBind test split, e.g., the co-folding model Nesso-1~\citep{shenoy2026nesso1} and affinity-supervised IPNet \citep{huang2024protein}. We mark them with $\dagger$ and include them for context rather than as leakage-controlled comparisons.

\textbf{Main results.}
\cref{tab:result-binding-affinity} shows that \ours{} improves over \oursC{} on both test sets, and the pre-trained \oursC{} itself is competitive with structure-based baselines. \ours{} is comparable to the strongest structure-based baseline, ProFSA \citep{gao2024self}. The sequence-based co-folding model~Nesso-1 \citep{shenoy2026nesso1} reaches similar correlation without any structural input, though its training data overlaps with the test split of LP-PDBBind. Without pre-training, \ours{} and \oursC{} perform comparably to weaker structure-based baselines, indicating that the pre-training provides transferable representations. The larger gain of \ours{} further supports that density provides complementary information. Overall, we read these results as evidence that experimental electron density carries transferable information on binding.

\input{ICLR2027/resource/figure/result-motivation}

\textbf{Density helps in uncertain structures.}
We attribute the density gain to structural uncertainty, which we quantify through crystallographic B factors as a proxy for positional uncertainty (higher values indicate less certain atomic positions). To test what the representation encodes, we keep the encoder frozen and train lightweight MLPs to decode B factors at the complex, residue, and atom levels. In \cref{fig:result-motivation-decoding}, \ours{} decodes B factors more accurately than \oursC{} and all baselines at every level. This is partly expected, since B factors are refined against the map, but it confirms that\ours{} retains this information. \Cref{fig:result-motivation-difference} relates this to binding affinity, where \ours{} ranks better than \oursC{} in the two higher B-factor bins while the lowest bin shows no difference. Density therefore helps most where the fitted coordinates are least certain, though this association is exploratory rather than causal. Further details are described in \cref{appx:method}.

\input{ICLR2027/resource/table/result-binding-affinity-design}
\textbf{Design choices.}
We also isolate the density-aware representations on the CL3 subset in \cref{tab:result-binding-affinity-design}. Both density channels contribute, and neither alone recovers the full gain; removing either the density or its gradient magnitude falls below \ours{} while staying above \oursC{}. The gain also requires the experimental signal at the input. Supplying density only as a reconstruction target, replacing the experimental map with random values, and using gradient magnitudes derived from atom blobs ($\lVert\nabla a \rVert$) all stay below \ours{}. Together, these results suggest that the benefit comes from the experimental density, providing information complementary to coordinates.

\input{ICLR2027/resource/table/result-drug}
\input{ICLR2027/resource/figure/result-drug-pocketwise}

\subsection{Pocket-conditioned ligand generation}
\label{subsec:exp-drug}

\textbf{Setup.}
Next, we evaluate pocket-conditioned ligand generation on the CrossDocked2020 benchmark \citep{francoeur20203d}, augmenting VoxBind \citep{pinheiro2024structure} with representations from \ours{}. We pass the reference complex based on the PDB ID to the frozen encoder, producing features per $P^{3}$-voxel block. These features are fused with VoxBind's features through a zero-initialized convolution. All other components follow the original VoxBind. We adopt established splits and preprocessing protocols~\citep{luo20213d,guan20233d-5e4,schneuing2024structure} and use 78,512 training data points and 79 test pockets with experimental maps for \ours{}.

\textbf{Evaluation.}
Following \citet{guan20233d-5e4}, we aim to generate $100$ ligands for each pocket with every method and ask three questions of the generated ligands. First, do they bind better to the pocket? Stronger binding is the goal of pocket-conditioned ligand generation; therefore, we dock the ligands with AutoDock Vina \citep{eberhardt2021autodock} and compare Vina scores and sample quality. Second, do the gains reflect new molecules rather than copies of the reference ligand? Since \ours{} conditions on the reference complex, higher scores would be trivial if it simply replayed the reference, so we measure the novelty and nearest-neighbor similarity against the training set. Third, are the molecules physically plausible? A better docking score is meaningful only for a realistic pose; we check molecular geometry with PoseCheck \citep{harris2023posecheck}, validity with PoseBusters~\citep{buttenschoen2024posebusters}, and geometry of the rigid fragments \citep{guan20233d-5e4}. For details and additional results on each evaluation, refer to \cref{appx:eval,appx:results}.

\input{ICLR2027/resource/figure/result-drug-qualPeratomNovelty}

\textbf{Density conditioning improves docking scores.}
As shown in \cref{tab:result-drug-design}, conditioning on experimental electron density results in ligands that dock better across pockets, while keeping sample quality close to VoxBind, which we report for context rather than as a ranking \citep{guan20233d-5e4}. This gain is broad rather than driven by a few pockets: \cref{fig:result-drug-pocketwise} shows a better mean and median docking score on a majority of pockets. As an example, \cref{fig:result-drug-example-qual} visualizes the best-scoring ligand of each method for \texttt{5LIU}. Also, \ours{}'s gain is not a size effect; while \ours{} generates slightly larger molecules (\# atoms/mol in \cref{tab:result-drug-design}), \cref{fig:result-drug-scoreperatom} shows that larger molecules do not always dock better in VoxBind, yet \ours{}'s gain holds across heavy atom counts. Notably, \oursC{} repeatedly generates the same ligands, leaving fewer distinct molecules within the same attempt budget ($n$ in \cref{tab:result-drug-design}), whereas density conditioning avoids this.

\textbf{Generated ligands do not copy the reference.}
We further show that \ours{} does not simply replay the ligand from the reference, similar to DecompDiff \citep{guan2024decompdiff}. Against the CrossDocked2020 training split, we report novelty of the generated ligands at two levels: (i) the fraction of canonical SMILES \citep{weininger1988smiles} and Bemis-Murcko scaffolds~\citep{bemis1996properties} that never appear in training \citep{polykovskiy2020molecular,brown2019guacamol}, and (ii) the similarity to the nearest neighbor (SNN) \citep{walters2020assessing} by the Tanimoto similarity~\citep{bajusz2015tanimoto} of each generated molecule to its closest training molecule. \Cref{fig:result-drug-reference-novelty} shows that every method maintains high molecule/scaffold novelty and low SNN. Despite conditioning on the reference complex, \ours{} remains at least as novel as VoxBind, indicating that density information supplies pocket context rather than a template to copy. We additionally find low similarity between each generated ligand and the reference ligand of the test pocket; see \cref{appx:novelty} for details.

\input{ICLR2027/resource/figure/result-drug-posecheck}

\textbf{Docking improved with realistic geometry.}
Since better docking scores are meaningful with realistic poses, we check the generated geometry with PoseCheck \citep{harris2023posecheck} following \citet{guan20233d-5e4,pinheiro2024structure}. \Cref{fig:result-drug-posecheck} reports three pose-level geometric measures, where \ours{} shows comparable numbers to VoxBind, sometimes better. \textit{Strain energy}, the universal force-field \citep[UFF]{rappe1992uff} relaxed energy a pose has above a stable conformer pose, detects internally distorted molecules. \Cref{fig:result-drug-posecheck-strain-ecdf} shows that \ours{} matches VoxBind and places ligands at lower strain than other baselines. This holds at fixed ligand size, i.e., the number of heavy atoms, as shown in \cref{fig:result-drug-posecheck-strain-per-atom}. \textit{Steric clashes}, the number of ligand atoms overlapping with protein atoms, detect poses the pocket cannot accommodate. \ours{} is comparable to VoxBind and the reference ligands, while DiffSBDD~\citep{schneuing2024structure} and TargetDiff \citep{guan20233d-5e4} show much heavier tails. \textit{Redocking RMSD}, how far Vina moves a pose when docking the molecule, measures the stability of the generated pose. As shown in \cref{fig:result-drug-posecheck-redock}, \ours{} attains the lowest median among the methods and is closest to the reference.

Beyond individual poses, we also compare the generated molecules against the reference distribution with the Jensen-Shannon distance \citep[JSD]{lin1991divergence} over atom, bond, and ring statistics \citep{guan20233d-5e4}. \Cref{fig:result-drug-posecheck-jsd} shows that \ours{} broadly follows the reference, although a lower JSD indicates closeness to the reference rather than a better molecule. Interestingly, \ours{} generates more rings than the reference ligands, even relative to their size, which raises its ring-count JSD.

\input{ICLR2027/resource/figure/result-drug-consistency}

\textbf{Rigid fragments keep their relaxed geometry.}
\ours{} reproduces the local geometry of its rigid fragments as well as VoxBind, and is best at the largest fragments. We follow \citet{guan20233d-5e4} and focus on rigid fragments, whose geometry is fixed by chemistry rather than by conformational choice, and report the RMSD between its generated pose and its force-field-relaxed pose, where lower is more physically plausible. \Cref{fig:result-drug-rigid-consistency} shows that most methods place small fragments close to their relaxed geometry, and RMSD grows with fragment size. In contrast, \ours{} stays with VoxBind while most baselines drift upward, and the margin widens along the rigid fragment size. While the improved consistency of large rigid fragments does not imply improved macrocycle generation, we additionally provide a macrocycle analysis in \cref{appx:macro}.

\input{ICLR2027/resource/figure/result-drug-posebusters}
\textbf{Comparable chemical validity to VoxBind.}
Furthermore, conditioning on experimental electron density yields a comparable chemical validity to VoxBind. We screen the generated ligands without any relaxation with PoseBusters \citep{buttenschoen2024posebusters} in its \texttt{dock} configuration, and report the fraction that passes all checks. \Cref{fig:result-drug-posebusters} shows that \ours{} is comparable to VoxBind and passes most categories at near-reference rates, with ring pucker as the main exception. Our post hoc analysis suggests that bond-order and hydrogen assignment in the OpenBabel-based reconstruction pipeline \citep{o2011open} contribute to many ring non-flatness failures. Reassigning eligible rings as aromatic reduces the ring non-flatness failure rate while preserving heavy atom coordinates; further details are provided in \cref{appx:pb}.

%% file: ICLR2027/resource/figure/result-motivation.tex
\begin{figure}
    \centering
    \includegraphics[width=.96\linewidth]{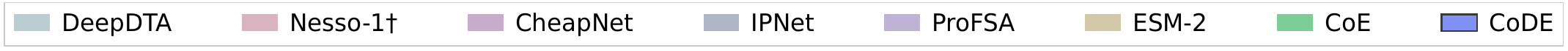}
    \begin{subfigure}[t]{0.64\linewidth}
        \centering
        \includegraphics[width=\linewidth]{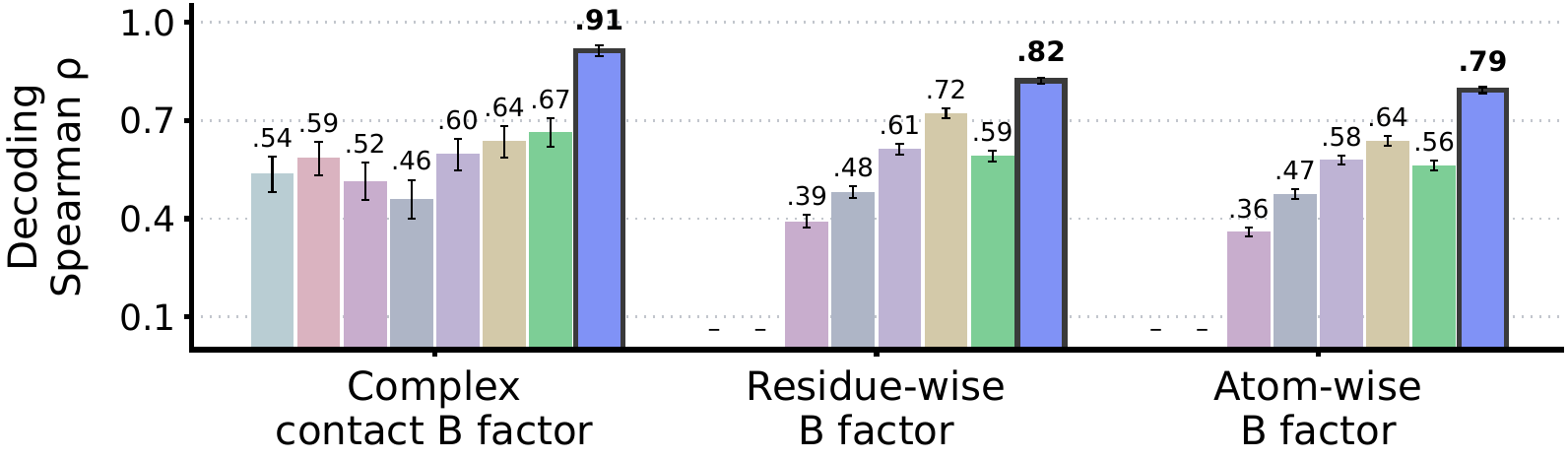}
        \caption{Decoding structural uncertainty from representations}
        \label{fig:result-motivation-decoding}
    \end{subfigure}
    \hfill
    \begin{subfigure}[t]{0.32\linewidth}
        \centering
        \includegraphics[width=\linewidth]{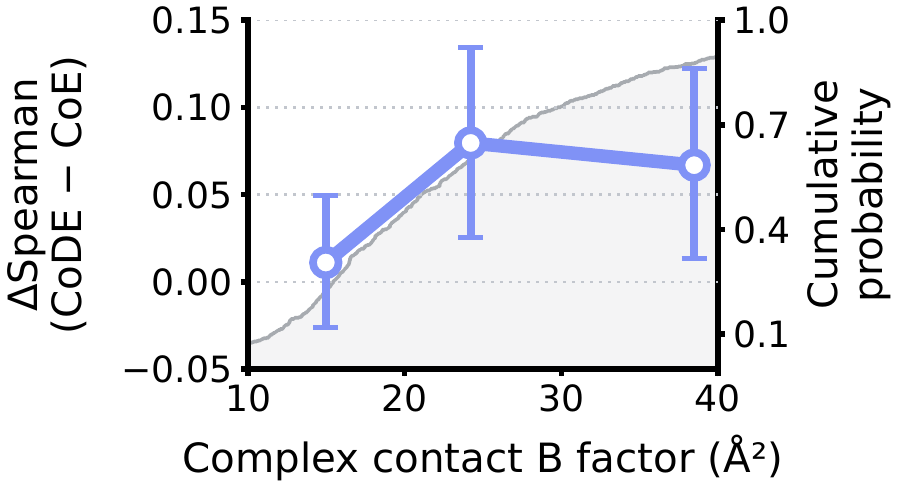}
        \caption{Binding-affinity improvement}
        \label{fig:result-motivation-difference}
    \end{subfigure}
    \caption{
        \textbf{Structural uncertainty in representations and affinity prediction.}
        \textbf{\subref{fig:result-motivation-decoding} Decoding structure uncertainty.}
        B-factors at complex, residue, and atom levels are decoded from the frozen representation of the CL3 test complexes. DeepDTA and Nesso-1 appear only at the complex level, since their representations are not tied to residue and atom levels.
        \textbf{\subref{fig:result-motivation-difference} Binding-affinity improvement.} 
        We split the test complexes into three complex contact B-factor bins ($<20$, $[20,30)$, $\geq 30$\,\AA$^2$), and show the CoDE$-$CoE Spearman gain. Results are averaged over five seeds, and points are placed at bin medians. The gray curve is the cumulative distribution of the complex contact B-factor.
    }
    \label{fig:result-motivation}
\end{figure}

%% file: ICLR2027/resource/table/result-binding-affinity-design.tex
\begin{wraptable}{r}{0.44\textwidth}
    \centering
    \vspace{-.15in}
    \caption{
        \textbf{Design choice results}. \textbf{Bold} marks best, \textsc{D} and \textsc{G} denote density and gradmag.
    }
    \label{tab:result-binding-affinity-design}
    \setlength{\tabcolsep}{4.5pt}
    \renewcommand{\arraystretch}{1.04}
    \setlength{\heavyrulewidth}{1pt}
    \setlength{\lightrulewidth}{0.4pt}
    \resizebox{.98\linewidth}{!}{%
        \begin{tabular}{l|ccc}
            \toprule
            \textbf{Design choices} & \textbf{Pearson \textit{r} $\uparrow$} & \textbf{Spearman $\rho$ $\uparrow$} & \textbf{RMSE $\downarrow$} \\
            \midrule
            \rowcolor{AffAccent!28} \ours{} (Ours)
            & \textbf{0.667\std{0.005}} & \textbf{0.638\std{0.006}} & \textbf{1.314\std{0.017}} \\
            \midrule
            \rowcolor{AffAccent!12} \;- w.o. \text{D}, \textsc{G} (\oursC{})
            & {0.620}\std{0.010} & 0.572\std{0.011} & {1.401}\std{0.030} \\
            \;- \oursC{} + \textsc{D} recon.
            & 0.597\std{0.021} & 0.562\std{0.015} & 1.421\std{0.048} \\
            \;- w.o. \textsc{D} channel
            & {0.652}\std{0.005} & {0.621}\std{0.005} & {1.326}\std{0.010} \\
            \;- w.o. \textsc{G} channel
            & {0.644\std{0.008}} & 0.605\std{0.006} & 1.347\std{0.016} \\
            \;- w. noisy \textsc{D,G}
            & 0.627\std{0.008} & 0.586\std{0.010} & 1.380\std{0.035} \\
            \;- w. $\lVert\nabla a \rVert$
            & 0.638\std{0.015} & 0.603\std{0.015} & 1.377\std{0.059} \\
        \end{tabular}
    }
    \vspace{-.15in}
\end{wraptable}

%% file: ICLR2027/resource/table/result-drug.tex
\begin{table}[t]
    \centering
    \caption{
        \textbf{Pocket-conditioned ligand generation results on the CrossDocked2020 benchmark} for 79 test pockets with experimental density maps. Vina evaluation metrics are taken over all of the generated molecules as in prior works. High aff. denotes the per-pocket percentage of the generated molecules with a better Vina docking score than the reference, aggregated over pockets, and $n$ denotes the number of valid molecules, counted as distinct per pocket for VoxBind variants.
    }
    \label{tab:result-drug-design}
    \setlength{\tabcolsep}{4.5pt}
    \renewcommand{\arraystretch}{1.04}
    \setlength{\heavyrulewidth}{1pt}
    \setlength{\lightrulewidth}{0.4pt}
    \resizebox{.98\textwidth}{!}{%
        \begin{tabular}{lccccccccccccccccc}
            \toprule
            & & \multicolumn{8}{c}{\textbf{Vina evaluation}} & \multicolumn{8}{c}{\textbf{Sample quality}} \\
            \cmidrule(lr){3-10}\cmidrule(lr){11-18}
            & & \multicolumn{2}{c}{Score $\downarrow$} & \multicolumn{2}{c}{Min $\downarrow$} & \multicolumn{2}{c}{Dock $\downarrow$} & \multicolumn{2}{c}{High aff.\,\% $\uparrow$} & \multicolumn{2}{c}{QED $\uparrow$} & \multicolumn{2}{c}{SA $\uparrow$} & \multicolumn{2}{c}{Div. $\uparrow$} & \multicolumn{2}{c}{\shortstack{\# atoms / mol}} \\
            \cmidrule(lr){3-4}\cmidrule(lr){5-6}\cmidrule(lr){7-8}\cmidrule(lr){9-10}\cmidrule(lr){11-12}\cmidrule(lr){13-14}\cmidrule(lr){15-16}\cmidrule(lr){17-18}
            \textbf{Method} & $n$ & Avg. & Med. & Avg. & Med. & Avg. & Med. & Avg. & Med. & Avg. & Med. & Avg. & Med. & Avg. & Med. & Avg. & Med. \\
            \midrule
            Reference & 79 & -6.21 & -6.10 & -6.66 & -6.15 & -7.32 & -6.99 & - & - & 0.46 & 0.46 & 0.72 & 0.75 & - & - & 22.8 & 21.0 \\
            \midrule
            AR & 7,655 & -5.76 & -5.58 & -6.16 & -5.85 & -6.73 & -6.54 & 41.8 & 37.0 & 0.50 & 0.50 & 0.63 & 0.62 & 0.71 & 0.71 & 17.2 & 16.0 \\
            Pocket2Mol & 7,772 & -5.06 & -4.78 & -6.47 & -5.90 & -7.27 & -6.79 & 51.2 & 51.0 & \textbf{0.57} & \textbf{0.57} & \textbf{0.74} & \textbf{0.75} & 0.72 & \underline{0.78} & 18.5 & 16.0 \\
            DiffSBDD & 7,720 & -2.70 & -4.62 & -4.89 & -5.41 & -6.71 & -6.78 & 41.3 & 36.4 & 0.49 & 0.50 & 0.63 & 0.62 & \textbf{0.80} & \textbf{0.80} & 20.2 & 20.0 \\
            DecompDiff & 6,427 & -5.00 & -4.97 & -5.90 & -5.70 & -6.89 & -6.77 & 49.8 & 46.8 & 0.49 & 0.50 & 0.67 & 0.65 & 0.71 & 0.72 & 20.9 & 21.0 \\
            TargetDiff & 7,287 & -5.27 & -5.92 & -6.42 & -6.46 & -7.48 & -7.51 & 54.8 & 52.7 & 0.49 & 0.50 & 0.61 & 0.60 & 0.75 & 0.73 & 22.2 & 22.0 \\
            FuncBind & 7,895 & -5.75 & -5.57 & -6.40 & -6.09 & -7.27 & -7.13 & 49.4 & 46.0 & 0.50 & 0.51 & 0.66 & 0.65 & 0.70 & 0.71 & 19.0 & 18.0 \\
            VoxBind\textsubscript{\scriptsize $\sigma$=0.9} & 7,887 & -6.15 & \underline{-6.83} & \underline{-7.12} & \underline{-7.22} & \underline{-7.97} & \underline{-8.05} & \underline{61.7} & 65.7 & \underline{0.54} & \underline{0.55} & \underline{0.70} & \underline{0.71} & \underline{0.76} & 0.76 & 24.0 & 25.0 \\
            \midrule
            \rowcolor{AffAccent!12} \;+ \oursC{} & 5,911 & \underline{-6.18} & -6.07 & -6.98 & -6.80 & -7.73 & -7.58 & 58.5 & \underline{66.0} & 0.43 & 0.44 & 0.58 & 0.58 & 0.56 & 0.54 & 24.4 & 25.0 \\
            \rowcolor{AffAccent!28}  \;+ \ours{} & 7,873 & \textbf{-6.58} & \textbf{-7.20} & \textbf{-7.64} & \textbf{-7.70} & \textbf{-8.48} & \textbf{-8.51} & \textbf{68.1} & \textbf{78.0} & 0.52 & 0.53 & 0.68 & 0.68 & 0.71 & 0.70 & 24.9 & 25.0 \\
            \bottomrule
        \end{tabular}
    }
\end{table}

%% file: ICLR2027/resource/figure/result-drug-pocketwise.tex
\begin{figure}[t]
\centering
    \includegraphics[width=.98\textwidth]{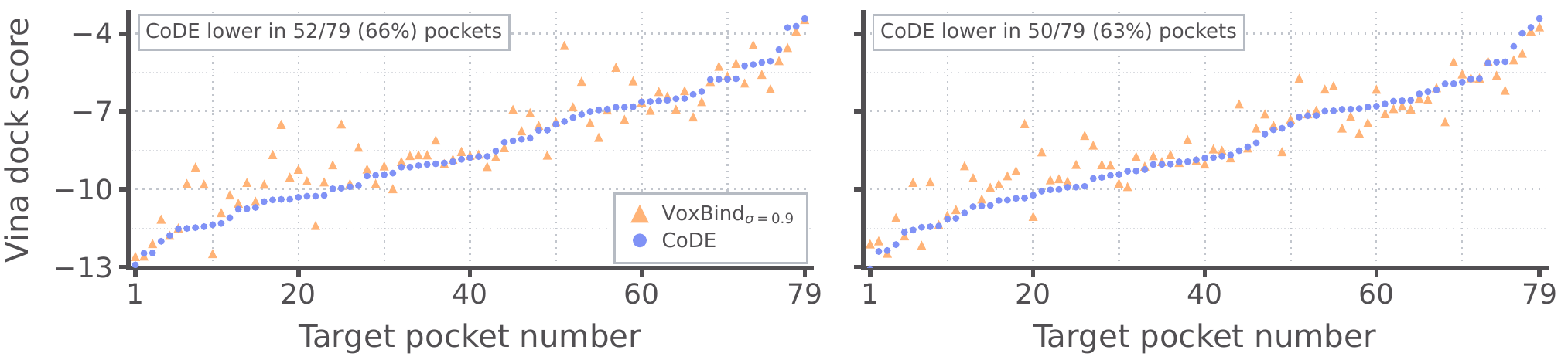}
    \caption{
        \textbf{Mean (left) and median (right) Vina docking scores} for generated molecules over 79 test pockets with experimental density maps, sorted by the \ours{} Vina docking score. \ours{} achieves better mean and median Vina docking scores than VoxBind for 66\% and 63\% of the pockets.
    }
    \label{fig:result-drug-pocketwise}
    \vspace{-.1in}
\end{figure}

%% file: ICLR2027/resource/figure/result-drug-qualPeratomNovelty.tex
\begin{figure}[t]
    \centering
    \begin{subfigure}[t]{0.53\textwidth}
        \vspace{0in}
        \centering
        \includegraphics[width=\linewidth]{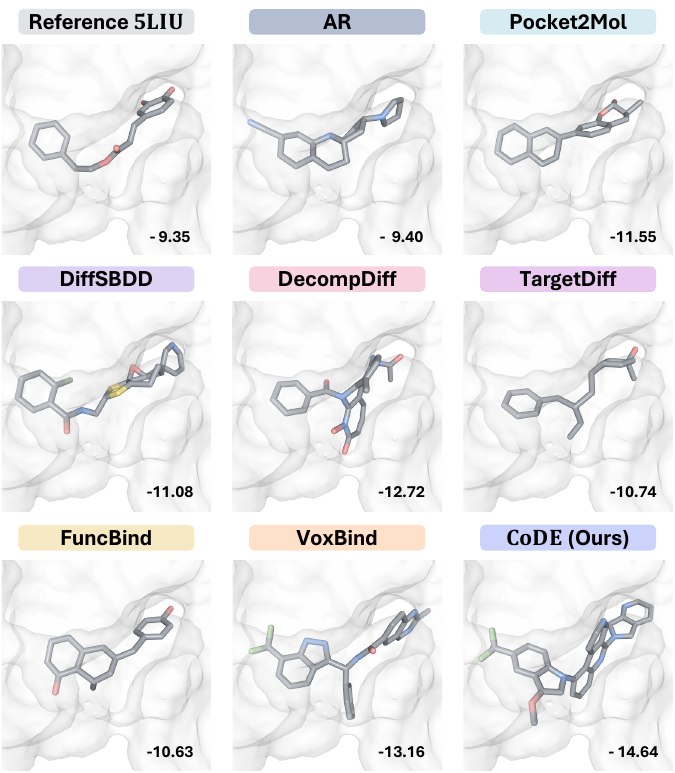}
        \caption{Visualization of generated ligands}
        \label{fig:result-drug-example-qual}
    \end{subfigure}
    \hfill
    \begin{minipage}[t]{0.45\textwidth}
        \vspace{0in}
        \begin{subfigure}[t]{\linewidth}
            \centering
            \includegraphics[width=.98\linewidth]{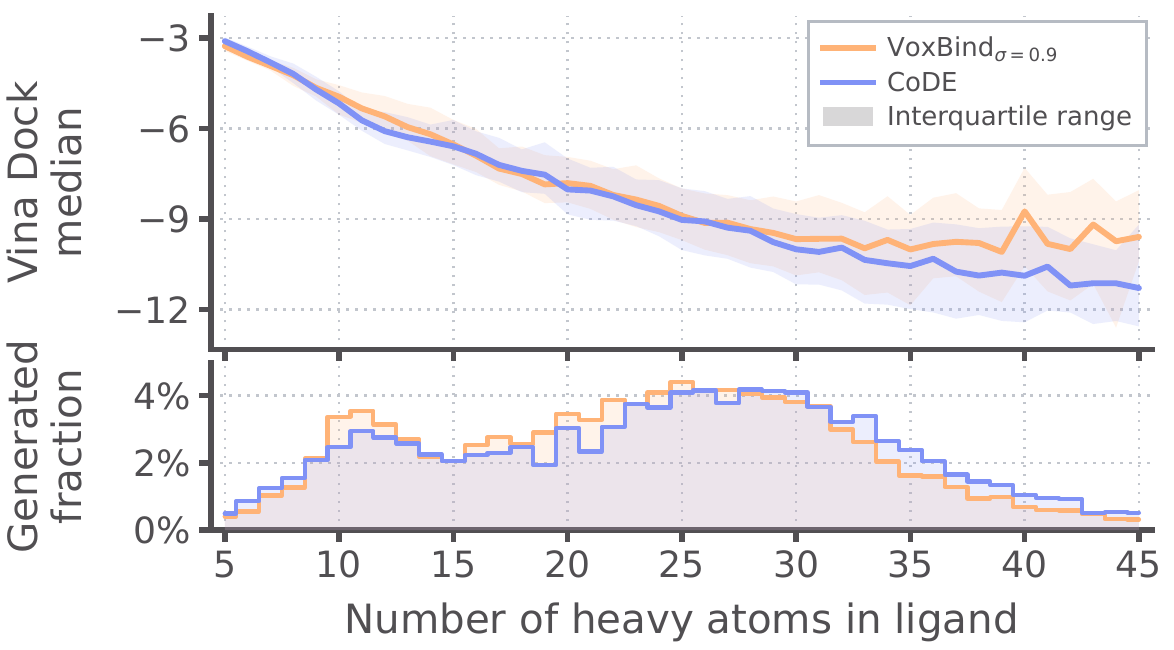}
            \caption{Docking score by number of heavy atoms}
            \label{fig:result-drug-scoreperatom}
        \end{subfigure}
        
        \vspace{.15in}
        
        \begin{subfigure}[t]{\linewidth}
            \centering
            \setlength{\tabcolsep}{4.5pt}
            \renewcommand{\arraystretch}{1.04}
            \setlength{\heavyrulewidth}{1pt}
            \setlength{\lightrulewidth}{0.4pt}
            \resizebox{.79\linewidth}{!}{%
            \begin{tabular}{lccc}
                \toprule
                & \multicolumn{2}{c}{\textbf{Novelty $\uparrow$}} & \multirow{2}{*}[-0.05in]{\textbf{SNN $\downarrow$}} \\
                \cmidrule(lr){2-3}
                \textbf{Method} & Mol.\,\% & Scaf.\,\% & \\
                \midrule
                AR & \textbf{99.1} & 79.3 & 0.297 \\
                Pocket2Mol & 96.5 & 69.7 & 0.349 \\
                DiffSBDD & 98.6 & 79.6 & \underline{0.286} \\
                DecompDiff & 96.9 & 76.2 & 0.335 \\
                TargetDiff & 98.0 & \underline{82.3} & \textbf{0.281} \\
                FuncBind & 93.8 & 70.7 & 0.339 \\
                VoxBind\textsubscript{\scriptsize $\sigma$=0.9} & 97.2 & 80.6 & 0.350 \\
                \midrule
                \rowcolor{AffAccent!12} \;+ \oursC{} & 98.3 & 82.2 & 0.343 \\
                \rowcolor{AffAccent!28} \;+ \ours{} & \underline{98.8} & \textbf{84.8} & 0.301 \\
                \bottomrule
            \end{tabular} %
            }
            \caption{Novelty against reference ligand}
            \label{fig:result-drug-reference-novelty}
        \end{subfigure}
    \end{minipage}
    \caption{
        \textbf{\subref{fig:result-drug-example-qual} Qualitative examples of the generated ligands}. For pocket \texttt{5LIU}, the best Vina-scoring ligand from each method is shown, with its docking score at the bottom right.
        \textbf{\subref{fig:result-drug-scoreperatom} Median Vina docking score by the number of heavy atoms}, with the distribution of generated ligand sizes at the bottom right. Docking scores do not simply become better proportionally to the molecule size, yet VoxBind + \ours{} lowers them across atom counts.
        \textbf{\subref{fig:result-drug-reference-novelty} Novelty against the reference ligands}, measured by molecule-level and scaffold-level novelty and nearest-neighbor similarity.
    }
\end{figure}

%% file: ICLR2027/resource/figure/result-drug-posecheck.tex
\begin{figure}[t]
    \centering
    \includegraphics[width=.96\linewidth]{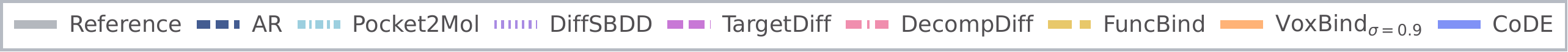}
    \begin{subfigure}[t]{0.49\linewidth}
        \centering
        \includegraphics[width=\linewidth]{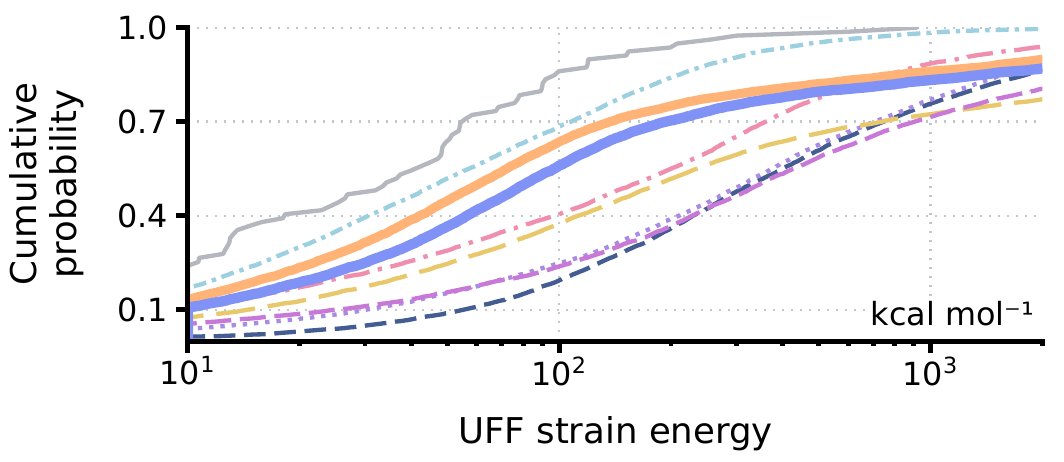}
        \caption{Cumulative distribution of strain energy}
        \label{fig:result-drug-posecheck-strain-ecdf}
    \end{subfigure}
    \hfill
    \begin{subfigure}[t]{0.49\linewidth}
        \centering
        \includegraphics[width=\linewidth]{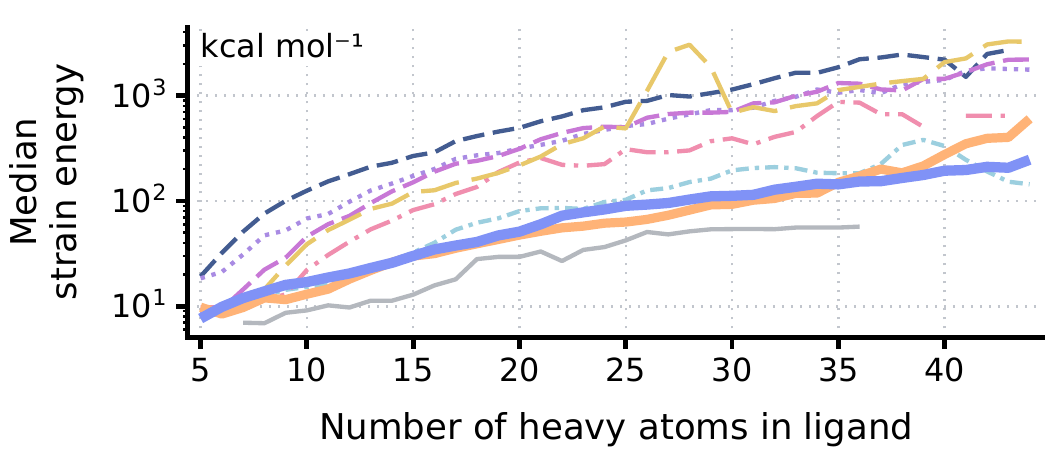}
        \caption{Strain energy per number of heavy atoms}
        \label{fig:result-drug-posecheck-strain-per-atom}
    \end{subfigure}
    \vspace{0.1in}
    \begin{subfigure}[t]{0.49\linewidth}
        \centering
        \includegraphics[width=\linewidth]{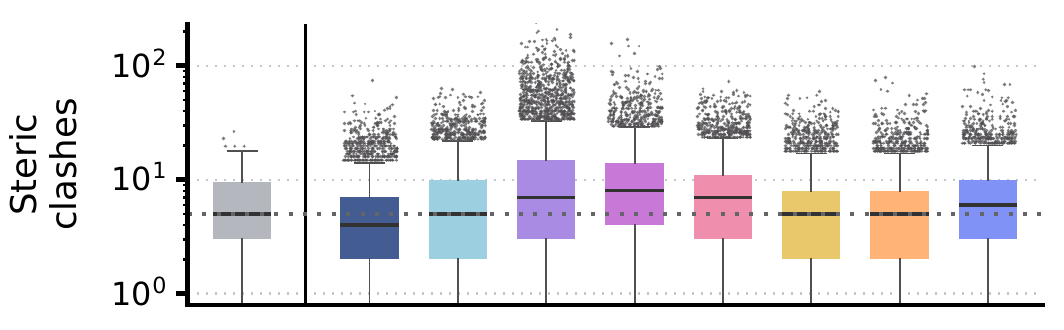}
        \caption{Number of steric clashes}
        \label{fig:result-drug-posecheck-steric-clash}
    \end{subfigure}
    \hfill
    \begin{subfigure}[t]{0.49\linewidth}
        \centering
        \includegraphics[width=\linewidth]{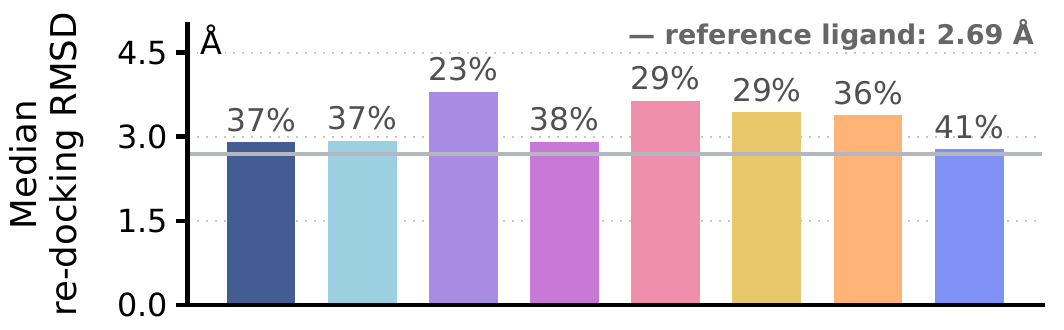}
        \caption{Median RMSD of re-docking}
        \label{fig:result-drug-posecheck-redock}
    \end{subfigure}
    \vspace{0.1in}
    \begin{subfigure}[t]{\linewidth}
        \centering
        \includegraphics[width=.98\linewidth]{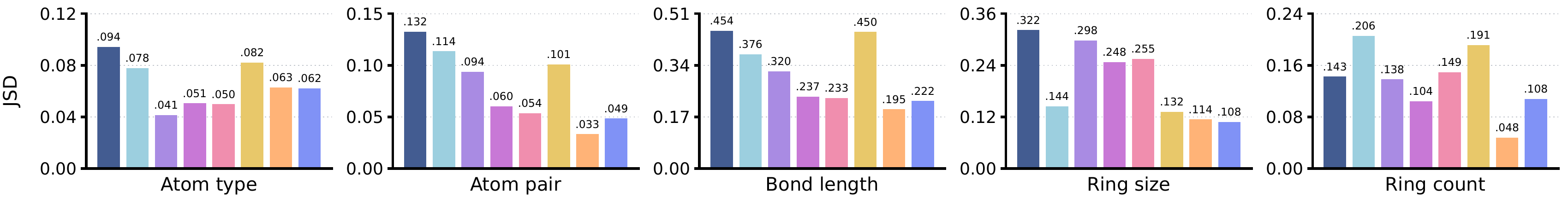}
        \caption{JSD to reference ligands regarding molecular properties}
        \label{fig:result-drug-posecheck-jsd}
    \end{subfigure}
    \caption{
        \textbf{PoseCheck evaluation of generated molecule poses.}
        Method colors are consistent across panels, with reference ligands shown in gray.
        \textbf{\subref{fig:result-drug-posecheck-strain-ecdf} Strain energy distributions} shown by the empirical cumulative distribution function.
        \textbf{\subref{fig:result-drug-posecheck-strain-per-atom} Median strain energy} as a function of the number of heavy atoms.
        \textbf{\subref{fig:result-drug-posecheck-steric-clash} Number of steric clashes.} Boxes denote the interquartile range, center lines denote the medians, whiskers extend to 1.5 times the interquartile range, and outliers are shown as individual points.
        \textbf{\subref{fig:result-drug-posecheck-redock} Median re-docking RMSD.}
        Labels above the bars indicate the percentage of poses with RMSD within 2~\AA{}, and the gray horizontal line marks the reference median of 2.69~\AA{}. 
        \textbf{\subref{fig:result-drug-posecheck-jsd} Jensen-Shannon distance (JSD) to reference ligands} for atom types, heavy-atom pair distances, bond lengths, ring sizes, and ring counts.
    }
    \label{fig:result-drug-posecheck}
    \vspace{-.12in}
\end{figure}

%% file: ICLR2027/resource/figure/result-drug-consistency.tex
\begin{figure}[t]
\centering
    \includegraphics[width=.98\textwidth]{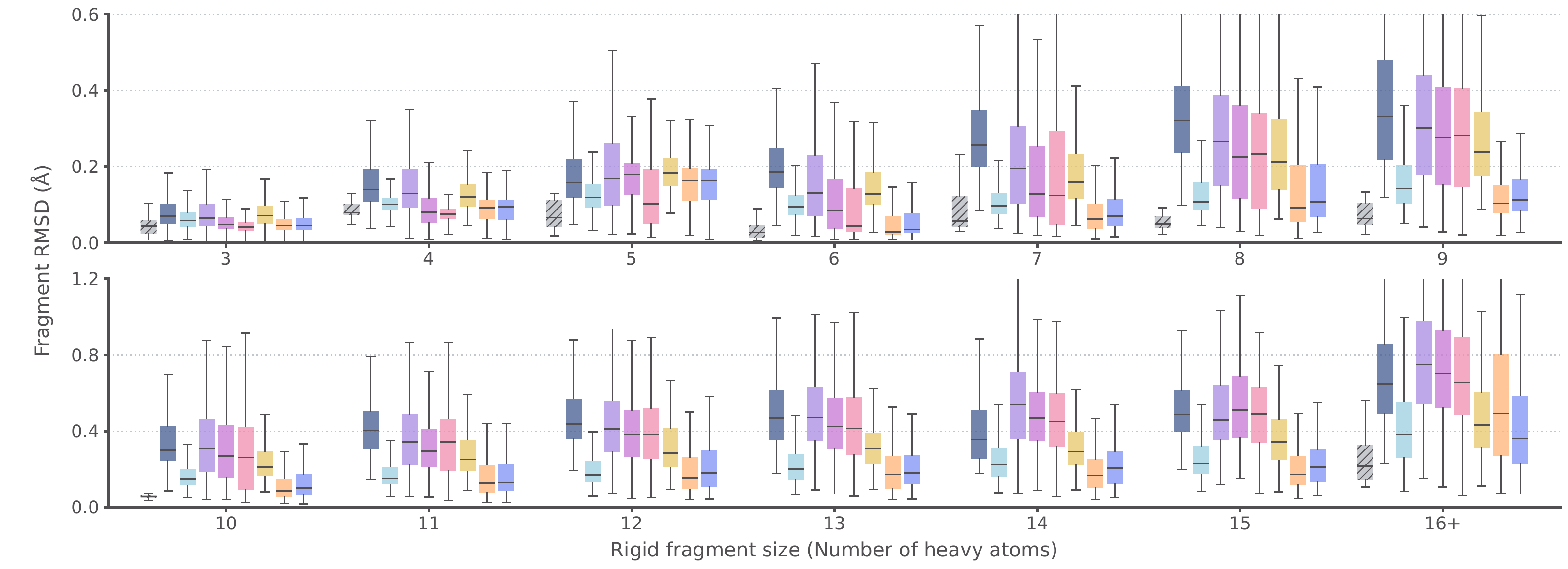}
    \includegraphics[width=.98\textwidth]{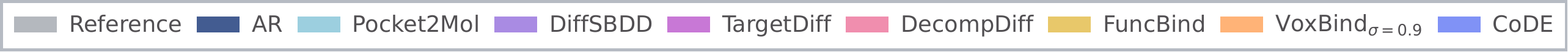}
    \caption{
        \textbf{Rigid fragment consistency results.} Median rigid-fragment RMSD between generated and force-field-relaxed pose after fragment-wise alignment, grouped by the number of heavy atoms. Boxes indicate the interquartile range, horizontal lines denote medians, and whiskers extend to 1.5 times the interquartile range. Lower RMSD indicates local geometry closer to a relaxed pose.
    }
    \label{fig:result-drug-rigid-consistency}
\end{figure}

%% file: ICLR2027/resource/figure/result-drug-posebusters.tex
\begin{wrapfigure}{r}{0.42\textwidth}
    \vspace{-.18in}
    \centering
    \includegraphics[width=.98\linewidth]{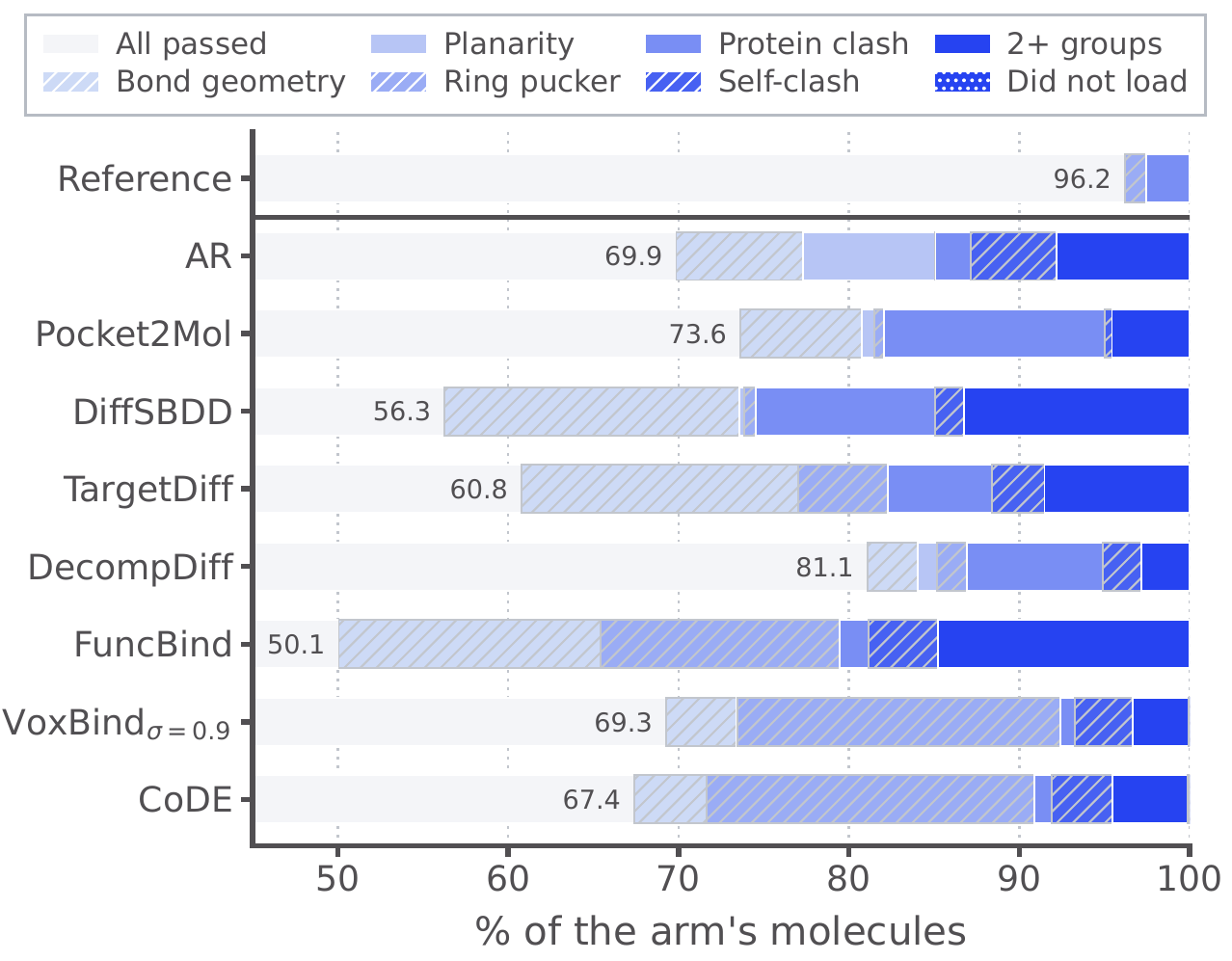}
    \caption{
        \textbf{Chemical validity by PoseBusters}, split by category (colors) with all-passed rate labeled; axis starts at 50\%.
    }
    \label{fig:result-drug-posebusters}
    \vspace{-.2in}
\end{wrapfigure}

%% file: ICLR2027/5_conclusion.tex
\section{Conclusion}
\label{sec:conclusion}

In this work, we introduced \ours{}, a voxel encoder jointly pre-trained on atomic occupancy and experimental \map{} maps by masked reconstruction. Across frozen-probe affinity regression and pocket-conditioned ligand generation, \ours{} improves over a matched coordinate-only encoder and stays competitive with task-specific baselines, and its advantage is largest where the fitted coordinates are least certain. These results support experimental electron density as a transferable complement to coordinates for learning protein--ligand interactions.

%% file: ICLR2027/appx/1_data.tex
\section{Data processing}
\label{appx:data}

\subsection{Dataset sources}

We use five datasets in this work, summarized in \cref{tab:dataset-overview}. PLINDER is used for pre-training, and PDBBind and CASF-2016 for binding-affinity regression with the LP-PDBBind split, and CrossDocked 2020 for pocket-conditioned generation. All electron density maps are obtained from PDBe.

\input{ICLR2027/resource/table/appx/appx-dataset}

\textbf{PLINDER.} 
PLINDER \citep{durairaj2024plinder} is our pre-training source. It curates the Protein Data Bank (PDB) as of 2024-04-09 into 449383 protein--ligand interaction (PLI) systems, each annotated with over 500 fields and with pocket-, ligand-, and interaction-level similarity. It also links 98473 experimental \textit{apo} structures and 205300 AlphaFold2 \citep{jumper2021highly} predictions to their \textit{holo} entries. We use only holo X-ray systems, drawn from all PLINDER splits, since our downstream tasks define their own test sets. Every downstream validation and test complex was excluded.

\textbf{PDBBind 2020.} 
PDBBind \citep{wang2005pdbbind,liu2015pdb} provides our affinity labels for the binding affinity regression downstream task. It pairs protein--ligand crystal structures with measured binding constants ($K_{d}, K_{i}, \mathrm{IC}_{50}$). The 2020 release consists of three sets: (i) the general set consisting of 19443 complexes, (ii) the refined set consisting of 5316 complexes that additionally satisfy stricter criteria such as a crystallographic resolution of 2.5 \AA{} or better and $K_{i}, K_{d}$ only records, and (iii) a core set of 285 complexes. The general, refined, core set was designed around structure quality rather than around similarity control. We therefore do not use the conventional split that trains on the refined set excluding the core set and evaluates on the core set.

\textbf{LP-PDBBind.}
LP-PDBBind \citep{li2026leak} re-splits PDBBind v2020 so that test complexes are dissimilar to training complexes. It showed that the standard sets share highly similar proteins and ligands, so gains on the core set overstate generalization. LP-PDBBind bounds protein and ligand similarity across 11513 training, 2422 validation, and three nested test tiers of 4286 (CL1), 2179 (CL2), and 1348 (CL3) complexes. CL1 removes covalent binders, ligands with rare elements, and complexes with steric clashes; CL2 keeps consistent $K_{d}$/$K_{i}$ measurements; and CL3 keeps only refined- and core-set complexes. We further drop entries that RDKit cannot parse.

\textbf{CASF-2016.}
CASF-2016 \citep{su2019comparative} serves as an additional held-out affinity test. It is built on the PDBBind core set, with 57 target clusters of five ligands that bind the same protein across a wide affinity range. Following \citet{wei2026honestaffinity}, we probe learned representations, restricted to the 92 complexes that have a density map and appear neither in pre-training nor in the LP-PDBBind training or validation split.

\textbf{CrossDocked2020.}
CrossDocked2020 \citep{francoeur20203d} is our pocket-conditioned ligand generation benchmark. It docks ligands into similar binding pockets across the PDB, yielding 22.5 million poses grouped by binding-site similarity. Following \citet{luo20213d,schneuing2024structure,guan20233d-5e4,guan2024decompdiff,pinheiro20233d-81b}, we filter poses by RMSD and sequence identity, leaving 100,000 training pairs and 100 test pockets from held-out clusters. For each receptor, we retrieve the density map of the crystal structure it derives from by its PDB ID and align the receptor to that frame. The map therefore shows the native crystal ligand of that entry rather than the docked ligand pose, which is computational and has no density of its own.

\textbf{Electron density maps.}
All maps are the likelihood-weighted \map{} maps that PDBe serves as pre-computed CCP4 grids, paired with the deposited coordinates:
\begin{center}
    \fcolorbox{gray!60}{gray!8}{\texttt{https://www.ebi.ac.uk/pdbe/entry-files/\{pdb\_id\}.ccp4}}
\end{center}
We use them as deposited, without re-refinement, re-phasing, or resolution and $B$-factor cutoffs.

\subsection{Preprocessing procedure}
\label{appx:data-preprocess}

\input{ICLR2027/resource/table/appx/appx-data-processing}

\textbf{PLINDER filtering.}
We keep X-ray ligand instances whose density resolves the ligand. Each instance must have a proper, RDKit-loadable, non-covalent ligand with 6--50 heavy atoms, no unresolved ligand heavy atoms, and a ligand RSCC of at least 0.8 \citep{fusani2020exploring}. We impose no resolution or pocket-RSCC cutoff, since the ligand RSCC already gates map quality at the binding site. Of the 112733 instances that pass map retrieval and grid construction, we use 112000 for training and 100 for validation. Atoms outside the 11 atom-type channels are omitted during voxelization rather than rejecting the complex.

\textbf{Leakage control.}
No downstream validation or test complex appears in pre-training. We remove the held-out PDB entries of CrossDocked2020 and PDBBind by exact PDB ID. We then verify that none of the LP-PDBBind validation (817) or test (1,320) complexes, nor the 92 CASF-2016 complexes, is in the pre-training set. 250 of the 3,850 LP-PDBBind training complexes do appear, which does not affect evaluation since pre-training uses no affinity labels.

\textbf{Frame alignment.}
CrossDocked2020 receptors must be moved into the crystal frame of their maps. They are re-docked copies in an arbitrary frame, so the density and atoms would otherwise disagree by several \AA{}ngstr\"om. We superpose each receptor onto its deposited structure by a residue-matched Kabsch fit \citep{kabsch1976solution} and apply the same transform to the ligand. Complexes with too few matched atoms or a fit RMSD above 2\,\AA{} are dropped. PLINDER and PDBBind use deposited coordinates and need no alignment.

\textbf{Grids.}
Each complex is a $64^3$ grid at 0.25\,\AA{} spacing, a 16\,\AA{} cube centered on the ligand centroid, trilinearly resampled from the CCP4 map. The pocket consists of protein heavy atoms within 10\,\AA{} of any ligand heavy atom, cropped to this cube. For pre-training, we instead cache a $116^3$ box (29\,\AA{}) per complex, stored in 8 bits over $[-2.2,5.0]$. Its 6.5\,\AA{} margin per side lets rotation augmentation resample the map at load time rather than zero-fill a fixed crop.

\textbf{Density normalization.}
We map raw density values onto a fixed, heavy-tail-compressed scale shared by all complexes:
\begin{equation}
    \rho=
    \frac{\operatorname{asinh}(\rho_{\mathrm{raw}}/0.5)-\mu_{\rho}}
         {\sigma_{\rho}},
    \qquad
    \mu_{\rho}=-0.01544,\quad \sigma_{\rho}=0.50848,
\end{equation}
where $\mu_{\rho}$ and $\sigma_{\rho}$ are estimated on PLINDER pockets. The gradient magnitude $\lVert\nabla\rho\rVert$ is computed on the normalized, resampled grid and standardized within each complex.

\textbf{Affinity and generation datasets.}
Both downstream sets keep only complexes with a density map. For affinity regression, we keep voxelizable PDBBind complexes with ligand and pocket RSCCs of at least 0.8, a non-covalent ligand, an LP-PDBBind split assignment, and a $K_d$ or $K_i$ label. This yields 3,850/817/1,320 train/validation/test complexes, plus the 92 CASF-2016 complexes. For generation, maps exist for 78,612 CrossDocked2020 training pairs, split into 78,512 for training and 100 for validation, and for 79 of the 100 test pockets. \Cref{tab:data-preprocessing} lists the cumulative filters.

%% file: ICLR2027/resource/table/appx/appx-dataset.tex
\begin{table}[!ht]
    \centering
    \caption{
        \textbf{Datasets used in this work.} Split counts are reported as train/validation/test.
    }
    \label{tab:dataset-overview}
    \small
    \resizebox{.98\textwidth}{!}{%
    \begin{tabular}{@{}llll@{}}
        \toprule
        \textbf{Dataset} & \textbf{Source size} & \textbf{Used after preprocessing} & \textbf{Role} \\
        \midrule
        PLINDER
            & 449,383 systems (1,357,906 instances)
            & 112,000 / 100 / -
            & Encoder pre-training \\
        PDBBind v2020
            & 19,443 complexes
            & 12,674 voxelized complexes
            & Source of affinity-labelled complexes \\
        LP-PDBBind
            & 18,795 split-assigned complexes
            & 3,850 / 817 / 1,320
            & Affinity regression \\
        CASF-2016
            & 285 complexes
            & - / - / 92
            & Clean held-out affinity evaluation \\
        CrossDocked2020
            & 22.5M poses (processed into 99,881 / / 100)
            & 78,512 / 100 / 79
            & Pocket-conditioned ligand generation \\
        \bottomrule
    \end{tabular}}
\end{table}

%% file: ICLR2027/resource/table/appx/appx-data-processing.tex
\begin{table}[!ht]
    \centering
    \caption{\textbf{Dataset preprocessing.} Counts are cumulative. PLINDER counts
    refer to ligand instances, whereas PDBBind counts refer to complexes.}
    \label{tab:data-preprocessing}
    \small
    \setlength{\tabcolsep}{4pt}
    \begin{tabular}{@{}lr@{}}
        \toprule
        \multicolumn{2}{c}{\textbf{PLINDER}} \\
        \midrule
        All annotated ligand instances & 1357906 \\
        X-ray diffraction & 1046290 \\
        Proper ligand; excluding ions, artifacts, and cofactors & 300276 \\
        Valid and RDKit-loadable & 289986 \\
        6--50 ligand heavy atoms & 259509 \\
        Non-covalent & 197369 \\
        Downstream validation/test PDB entries excluded & 166448 \\
        No unresolved ligand heavy atoms & 153270 \\
        Ligand RSCC $\geq$ 0.8 & 113976 \\
        Final grid-eligible manifest & 112733 \\
        Encoder train / validation subset & 112000 / 100 \\
        \midrule
        \multicolumn{2}{c}{\textbf{PDBBind / LP-PDBBind}} \\
        \midrule
        All PDBBind v2020 complexes & 19443 \\
        Valid atomic grid and experimental map & 12674 \\
        Ligand RSCC computable & 11601 \\
        Ligand and pocket RSCCs $\geq$ 0.8 & 10926 \\
        Non-covalent & 10424 \\
        Assigned to an LP-PDBBind split & 10128 \\
        $K_d$ or $K_i$ label & 5987 \\
        Final train / validation / test split & 3850 / 817 / 1320 \\
        \bottomrule
    \end{tabular}
\end{table}

%% file: ICLR2027/appx/2_method.tex
\newpage
\section{\ours{} details}
\label{appx:method}

In this section, we explain the details of the \ours{} encoder and how it is used. 

\begin{itemize}
    \item Encoder configuration and model hyperparameters: \cref{appx:method-arch}
    \item Pre-training configuration: \cref{appx:method-pretrain}
    \item Downstream task configuration: \cref{appx:method-downstream}
    \item Motivational analysis: \cref{appx:method-motivation}
    \item Ablation experiment configuration: \cref{appx:method-ablation}
\end{itemize}

\subsection{Architecture}
\label{appx:method-arch}

\textbf{Encoder.}
\ours{} is a 99.47M-parameter Channel-ViT \citep{bao2023channel-24a} that tokenizes each channel group separately. The 13 input channels form three groups of sizes $(7,4,2)$: ligand atoms, pocket atoms, and the density and gradient magnitude pair. Each group has its own $8^{3}$ convolution, which embeds the $64^3$ grid into 512 patch tokens of width 640. Every token receives a learnable positional embedding, shared across groups, and a learnable embedding of its group. The resulting 1536 tokens pass through 18 pre-LayerNorm \citep{ba2016layer} Transformer blocks \citep{vaswani2017attention} with 10 heads, an MLP ratio of 4, GELU activations \citep{hendrycks2016gaussian}, and dropout of 0.1, followed by a final LayerNorm. Since all groups attend to one another, the encoder can relate density to atoms at every location.

\textbf{Voxel projection and reconstruction head.}
The encoder includes a linear voxel projection. For each patch, the final tokens of its three groups are averaged and linearly projected to a 16-channel $8^3$ feature cube. A separate two-layer reconstruction MLP head with hidden width 67 maps the 16-channel cube of each patch to the 13 input channels. The reconstruction MLP head is used only for pre-training and discarded afterwards.

\textbf{Input construction.}
Each input has 13 channels: 7 ligand elements (C, O, N, S, F, Cl, P), 4 pocket elements (C, O, N, S), $\rho$, and $\lVert\nabla\rho\rVert$. Atoms are rendered as soft occupancies following \citet{orlando2022pyuul}, with $R_i$ the van der Waals radius of atom $i$ and $\sigma=0.93$:
\begin{equation}
    g_i(v) = \exp\!\left(-\frac{\lVert v - r_i\rVert^2}{\sigma^2 R_i^2}\right),
    \qquad
    a_c(v) = 1 - \prod_{i \in c}\bigl(1 - g_i(v)\bigr),
\end{equation}
so each element channel $c$ is a soft union of its atoms, bounded in $[0,1]$. Pre-training uses van der Waals radii for both ligand and pocket, since a uniform ligand radius would reveal the ligand/pocket split through blob size. The generation model keeps VoxBind's uniform 0.5\,\AA{} ligand radius, so that our conditioning is a drop-in replacement.

\textbf{Density and gradient channels.}
The density and gradient channels form their own group, so removing them leaves the atomic groups unchanged. $\rho$ is the normalized map of \cref{appx:data-preprocess}, and its gradient magnitude highlights atom boundaries and partially occupied regions. We compute the gradient in-model by central differences on the voxel grid,
\begin{equation}
    \label{eq:appx-grad}
    \partial_{x}\rho(v) = \tfrac{1}{2}\bigl(\rho(v+\mathbf{e}_{x}) - \rho(v-\mathbf{e}_{x})\bigr),
    \qquad
    \lVert\nabla\rho\rVert(v) = \sqrt{\partial_{x}\rho(v)^{2} + \partial_{y}\rho(v)^{2} + \partial_{z}\rho(v)^{2}},
\end{equation}
where $\mathbf{e}_{x}$ is a one-voxel step along $x$, $\partial_{y}$ and $\partial_{z}$ are defined likewise, and boundary voxels are replicated. We then standardize the channel within each complex,
\begin{equation}
    \label{eq:appx-zscore}
    \mathcal{Z}[f](v) = \frac{f(v) - \mu_{f}}{\sigma_{f}},
\end{equation}
where $\mu_{f}$ and $\sigma_{f}$ are the mean and standard deviation of $f$ over the grid, so the input channel is $\mathcal{Z}[\lVert\nabla\rho\rVert]$ and the voxel spacing cancels. Since the gradient is recomputed after each augmentation, it always matches the resampled density. Dropping one channel leaves a group of size one, and dropping both recovers \oursC{}.

\newpage
\subsection{Pre-training configuration}
\label{appx:method-pretrain}

\textbf{Masking.}
We mask 75\% of the blocks, biased toward those occupied by atoms. Each of the 512 blocks is scored by its total atomic occupancy plus Gaussian noise with standard deviation $\tau=1$ times the within-complex standard deviation of the scores. The 384 highest-scoring blocks are masked, and the same spatial mask is applied to all channels.

\textbf{Loss weights.}
The channel weights balance sparse elements against the dense density channels. For each atomic channel, $w_c$ is inversely proportional to its occupied-voxel frequency, estimated on 2000 training complexes with an occupancy threshold of 0.05 and capped at a largest-to-smallest ratio of 30. $\rho$ and $\lVert\nabla\rho\rVert$ receive unnormalized weights of 0.1, and all 13 weights are then normalized to sum to 13. The voxel weight is $\omega_{c,v}=10$ where an atomic channel exceeds 0.05 and $1$ otherwise.

\textbf{Optimization.}
We pre-train for 100 epochs with the settings of \cref{tab:appx-hparams} and use the best valid checkpoint for the main affinity results. Atoms and density are rotated and translated jointly, and rotation resamples the cached box at the augmented pose, $\rho_{\text{aug}}(v) = \rho_{\text{box}}(R^{\top}(v - t))$, rather than shifting a fixed crop. All results use the exponential moving average (EMA) of the parameters.

\input{ICLR2027/resource/table/appx/appx-method-hp}

\newpage
\subsection{Downstream task adaptation}
\label{appx:method-downstream}

\textbf{Binding-affinity probing.}
We fit a small MLP on frozen, mean-pooled encoder tokens. The 1536 final tokens are averaged into a 640-dimensional complex embedding, and a two-layer MLP regresses affinity from it with mean-squared error (\cref{tab:appx-hparams}). We keep the checkpoint with the highest validation Spearman correlation and use no test-time rotation averaging.

\textbf{Pocket-conditioned ligand generation.}
We keep the VoxBind \citep{pinheiro2024structure} walk-jump sampler and change only its pocket conditioning. The frozen encoder reads the reference complex with the experimental density. Its patch tokens are averaged over channel groups, and the encoder's linear output layer, trained with it during pre-training, maps each token to the features of the voxels in its patch, restoring the full grid as $x_{\text{dens}}$. This output is fused with VoxBind's ligand and pocket features $x$ through a zero-initialized $1\times1\times1$ convolution, $x \leftarrow x + W[x \Vert x_{\text{dens}}]$. Only the denoiser and $W$ receive gradients and are updated.

\textbf{Generation optimization.}
The denoiser follows VoxBind: a 3D U-Net with base width 32, channel multipliers $[1,2,2,4]$, two residual blocks per level, and attention at the two coarsest levels (111.6M parameters). We train it for 350 epochs from scratch with AdamW, a constant learning rate of $10^{-5}$, weight decay $10^{-2}$, noise level $\sigma=0.9$, and a batch size of 32 per GPU on four H200 or RTX PRO 6000 Blackwell GPUs. Sampling uses walk-jump with $\sigma=0.9$, friction $\gamma=1$, step $\delta=\sigma/2$, and 400 warm-up and 100 sampling steps from a de novo initialization, with the pocket masked out of both noise and score. Grids are decoded to atoms by non-maximum-suppressed peak picking at threshold 0.25, followed by L-BFGS refinement against the predicted grid.

\subsection{Motivational observation configuration}
\label{appx:method-motivation}

\textbf{Complex contact B factor.}
The complex contact B factor summarizes how disordered the protein is where it touches the ligand. It is the mean B factor of protein heavy atoms within 5\,\AA{} of any ligand heavy atom,
\begin{equation}
    B_{\mathrm{complex}} = \frac{1}{|\mathcal{I}|}\sum_{i\in\mathcal{I}} B_i,
\end{equation}
where $\mathcal{I}$ is this set of protein heavy atoms and $B_i$ is the deposited B factor of atom $i$. It is the complex-level decoding target and defines the affinity bins below.

\textbf{Decoding.}
In \cref{fig:result-motivation-decoding}, we decode B factors from frozen representations of 732 CL3 test complexes without fine-tuning the encoders. At the complex level, the target is $B_{\mathrm{complex}}$, predicted from one pooled embedding per complex. At the residue level, the target is the mean B factor of the evaluated atoms in each residue; at the atom level, it is the atom's B factor. These two levels use local tokens from the same 48,591 protein atoms covered by all compared token models. We train MLP decoders with five complex-disjoint folds and average predictions across three seeds. Residue and atom decoders are trained on B factors standardized within each complex, which preserves their within-complex ranks. We report Spearman correlation across complexes for the complex-level target and the mean of within-complex Spearman correlations for residue and atom targets. Error bars are 95\% confidence intervals from 2,000 bootstrap resamples of complexes.

\textbf{Binding affinity grouping.}
In \cref{fig:result-motivation-difference}, we compare the affinity gain within bins of $B_{\mathrm{complex}}$ in the CL3 setting of \cref{tab:result-binding-affinity}, where probes are trained and evaluated on the CL3-filtered LP-PDBBind split (1559/410/733). We split the 733 test complexes into $B_{\mathrm{complex}}<20$, [20,30), and $\geq 30$\,\AA$^2$, which hold 330, 222, and 181 complexes. Within each bin, we compute the Spearman correlation of \ours{} and \oursC{} for each of five probe seeds and average the per-seed differences. The 95\% intervals are percentile intervals from 8000 paired bootstrap resamples of complexes within each bin. The analysis is exploratory, without multiplicity correction.

\newpage
\subsection{Ablation experiments configuration}
\label{appx:method-ablation}

\textbf{Shared recipe.}
All ablations change only the input or target channels of the \ours{} recipe. They share the architecture, PLINDER data, masking ratio of 0.75, atom-biased masking, channel weighting, optimizer, EMA, and seeds. Each row of \cref{tab:result-binding-affinity-design} averages the five-seed CL3 probe means and standard deviations.

\textbf{Channel selection.}
Each variant removes channels or moves them to the target only.
\begin{itemize}
    \item \ours{}: 13 input channels with groups $(7,4,2)$
    \item w.o.\ D, G (\oursC{}): 11 atomic channels with groups $(7,4)$
    \item \oursC{} + D recon.: 11 input channels, but all 13 channels are reconstructed, so density enters only as a target
    \item w.o.\ D input: atomic channels and $\lVert\nabla\rho\rVert$ of the real density, with groups $(7,4,1)$
    \item w.o.\ G input: atomic channels and $\rho$, with groups $(7,4,1)$
\end{itemize}

\textbf{Noisy pseudo-density.}
The noise control keeps the channel count and value distribution of the density but removes its spatial structure. We build one histogram over all voxels of the 112000 pre-training crops and draw every voxel of $\rho$ independently from it. The noise of each complex is fixed by a per-complex seed, so it stays the same across epochs and is never mixed between complexes. $\lVert\nabla\rho\rVert$ is recomputed from the noise with the same function and per-complex standardization as for the real density.

\textbf{Atom-blob gradients.}
This control tests whether a gradient channel helps without experimental density. We replace the density and its gradient with the gradient magnitude of the atomic occupancies,
\begin{equation}
    \lVert\nabla a\rVert(v) = \mathcal{Z}\Bigl[\,\sum_{c=1}^{11} \lVert\nabla a_{c}\rVert\Bigr](v),
\end{equation}
where each $\lVert\nabla a_{c}\rVert$ uses the central differences of \cref{eq:appx-grad} on atomic channel $c$, and $\mathcal{Z}$ is the standardization of \cref{eq:appx-zscore}. The row thus carries coordinates plus $\lVert\nabla a\rVert$ with groups $(7,4,1)$ and no density.

\newpage
\section{Baseline details}
\label{appx:baseline}

\subsection{Binding affinity regression}
We compare against a broad set of baselines covering sequence-based, structure-based, and pre-trained protein--ligand complex representation models. For methods that provide pre-trained encoders, we follow the same probing protocol as our model and train a lightweight prediction head MLP on top of the frozen representations. For methods without publicly available pre-trained representations, we train the model on the LP-PDBbind training split using its original or recommended training objective as possible. The baselines considered in our experiments are as follows:

\begin{itemize}
    \item \textbf{DeepDTA} \citep{ozturk2018deepdta} uses 1D convolutional networks on a ligand SMILES and the amino acid sequence of the target protein.
    \item \textbf{MolTrans} \citep{huang2021moltrans} uses Transformer encoders to model substructural tokens extracted from the ligand SMILES and the full protein sequence, without using 3D coordinates.
    \item \textbf{HonestAffinity} \citep{wei2026honestaffinity} uses frozen ESM-2 650M embeddings \citep{lin2023evolutionary} for the full protein sequence, with a binary indicator specifying pocket residues. It does not use 3D coordinates as input.
    \item \textbf{Nesso-1} \citep{shenoy2026nesso1} takes the full protein sequence and a ligand represented by SMILES or molecular structure, following a co-folding style architecture stemming from \textbf{Boltz-2} \citep{passaro2025boltz}. It does not require a pre-defined pocket or experimentally determined protein coordinates as input.
    \item \textbf{HBGSA} \citep{kong2026hbgsa} uses the 3D protein--ligand binding complex, including explicit coordinates of the protein--ligand hydrogen bonds. It models the spatial topology with a graph neural network and self-attention. Additionally, it uses the binding-site structure rather than the full protein sequence.
    \item \textbf{EGNN} \citep{duval2023hitchhikers-e8b} represents the protein pocket and ligands as atom features and 3D coordinates, and process them using an E(3)-equivariant graph neural network.
    \item \textbf{EGNN + TargetDiff} \citep{guan20233d-5e4} uses the same protein-pocket and ligand atoms with 3D coordinates as above, while initializing the encoder with TargetDiff pre-trained for protein-conditioned 3D ligand generation.
    \item \textbf{GET} \citep{kong2023generalist} represents the 3D protein--ligand complex as a hierarchical geometric graph and applies an E(3)-equivariant Transformer to model intra- and intermolecular interactions. 
    \item \textbf{CheapNet} \citep{lim2025cheapnet} uses the protein binding pocket and ligand with their 3D atomic structures and learns hierarchical atom- and cluster-level representations with cross-molecular interactions.
    \item \textbf{IPNet} \citep{huang2024protein} encodes the 3D protein pocket and ligand with SE(3)-equivariant networks and cross-attention layers, pre-trained with binding affinity supervision on PDBBind 2016 \citep{liu2015pdb}. Additionally, it is used as the interaction prior component of IPDiff.
    \item \textbf{AEV-PLIG} \citep{valsson2025narrowing} uses atoms from the protein binding site and the ligand, along with their 3D coordinates, represented as atomic environment vectors and a protein--ligand interaction graph.
    \item \textbf{GeoSSL} \citep{liu2022molecular} pre-trains molecular representations from 3D atomic coordinates using SE(3)-invariant denoising distance matching. Unlike protein--ligand pre-training methods, its objective operates on individual molecular structures without explicit protein--ligand interactions.
    \item \textbf{DSMBind} \citep{jin2023dsmbind} uses 3D protein--ligand complex and learns a binding-energy fucntion through SE(3) denoising score matching. Its protein representation is structure-based, rather than full amino-acid sequence-based.
    \item \textbf{BindNet} \citep{feng2024protein} uses Uni-Mol \citep{zhou2023uni} encoders for the 3D protein pocket and ligand, taking atom types and coordinates as input and pre-trains on cross-molecular interactions.
    \item \textbf{ProFSA} \citep{gao2024self} encodes the 3D protein pocket with atom types and coordinates, and pre-trains it by aligning pocket representations with Uni-Mol embeddings of nearby protein fragments treated as pseudo-ligands.
\end{itemize}

\subsection{Pocket-conditioned ligand generation}
We compare our model against autoregressive, diffusion-based, and voxel-based pocket-conditioned molecular generation methods. We use the publicly released generated samples or official implementations with their recommended configurations. The baselines considered in our experiments are as follows:

\begin{itemize}
    \item \textbf{AR} \citep{luo20213d} sequentially samples ligand atoms from a learned spatial distribution conditioned on the protein pocket.
    \item \textbf{Pocket2Mol} \citep{peng2022pocket2mol} uses an E(3)-equivariant network to autoregressively generate atom positions, types, and bonds.
    \item \textbf{DiffSBDD} \citep{schneuing2024structure} generates ligand point clouds by jointly denoising atom positions and types conditioned on the pocket.
    \item \textbf{TargetDiff} \citep{guan20233d-5e4} uses an equivariant diffusion model to jointly generate ligand coordinates and atom types through protein--ligand interactions.
    \item \textbf{DecompDiff} \citep{guan2024decompdiff} decomposes ligands into arms and scaffolds, then generates them using separate diffusion priors.
    \item \textbf{VoxBind} \citep{pinheiro2024structure} represents molecules as atomic-density voxel grids, and generates ligands through conditional voxel denoising.
    \item \textbf{FuncBind} \citep{kirchmeyer2026unified} represents molecules as continuous neural fields for target-conditioned all-atom generation, trained on three modalities: small molecules, macrocyclic peptides(MCPs), and antibody complementarity-determining region (CDR) loops.
\end{itemize}

Note that our model is built on top of VoxBind, where the original pocket representation is augmented with the representation obtained from our density-pre-trained encoder. Unless otherwise stated, the remaining generative architecture and sampling procedure follow the original VoxBind implementation.

%% file: ICLR2027/resource/table/appx/appx-method-hp.tex
\begin{table}[!ht]
    \centering
    \caption{\textbf{\ours{} hyperparameters.} Pre-training and affinity-probing settings of the encoder behind the main affinity results.}
    \label{tab:appx-hparams}
    \small
    \setlength{\tabcolsep}{4pt}
    \begin{tabular}{@{}ll@{}}
        \toprule
        \multicolumn{2}{c}{\textbf{Encoder}} \\
        \midrule
        Patch size & $8^3$ voxels (2\,\AA{}), 512 patches per group \\
        Tokens & 1536 (3 groups $\times$ 512) \\
        Width / depth / heads & 640 / 18 / 10 \\
        MLP ratio / activation & 4 / GELU \\
        Dropout & 0.1 \\
        Positional embedding & learnable, shared across groups, plus a learnable group embedding \\
        Voxel projection & group average, linear to 16 channels \\
        Reconstruction MLP head & per patch, 2 layers, hidden 67 \\
        Parameters & 99.47M \\
        \midrule
        \multicolumn{2}{c}{\textbf{Pre-training}} \\
        \midrule
        Data & PLINDER, 112000 train / 100 validation \\
        Mask ratio / strategy & 0.75 / atom-biased, $\tau$ = 1 \\
        Mask block & $8^3$ voxels, shared across channels \\
        Channel weight $w_c$ & inverse occupied-voxel frequency, max ratio 30; 0.1 for $\rho$ and $\lVert\nabla\rho\rVert$ \\
        Voxel weight $\omega_{c,v}$ & 10 where atomic occupancy $>$ 0.05, else 1 \\
        Optimizer & AdamW, $(\beta_1,\beta_2)$ = (0.99, 0.999), $\epsilon$ = 0 \\
        Learning rate / schedule & $10^{-4}$ / constant \\
        Weight decay & 0.05 \\
        Batch size & 128 (4 per GPU $\times$ 8 accumulation steps $\times$ 4 GPUs) \\
        Gradient clipping & 1.0 \\
        Precision & bfloat16 \\
        EMA decay & 0.999 \\
        Epochs / checkpoint & 100 / best validation checkpoint (epoch 25) \\
        Augmentation & random rotation, translation in $[-1,1]$\,\AA{} per axis \\
        Seed & 42 \\
        \midrule
        \multicolumn{2}{c}{\textbf{Affinity probe}} \\
        \midrule
        Pooling & mean over 1536 final tokens (640-d) \\
        Head & MLP 640 $\rightarrow$ 128 $\rightarrow$ 1, SiLU, dropout 0.1 \\
        Optimizer & Adam, learning rate $10^{-3}$, weight decay $10^{-4}$ \\
        Batch size / epochs & 64 / at most 200, early stopping after 30 \\
        Model selection & best validation Spearman correlation \\
        Seeds & 0, 1, 2, 3, 4 \\
        \bottomrule
    \end{tabular}
\end{table}

%% file: ICLR2027/appx/3_evaluation.tex
\section{Evaluation details}
\label{appx:eval}

\subsection{Binding affinity regression}
We evaluate binding affinity prediction by Pearson correlation, Spearman rank correlation, and root mean squared error (RMSE), following \citet{townshend2020atom3d}. Let $y_i$ denote the ground-truth affinity in p$K$ units, $\hat{y}_i$ the predicted affinity, and $N$ the number of protein--ligand complexes.

\begin{itemize}
    \item \textbf{Pearson correlation.}
    The linear correlation between predicted and measured affinities:
    \[
        r =
        \frac{
            \sum_{i=1}^{N}(y_i-\bar{y})(\hat{y}_i-\bar{\hat{y}})
        }{
            \sqrt{\sum_{i=1}^{N}(y_i-\bar{y})^2}
            \sqrt{\sum_{i=1}^{N}(\hat{y}_i-\bar{\hat{y}})^2}
        }\,,
    \]
    where $\bar{y}$ and $\bar{\hat{y}}$ are the mean ground-truth and predicted affinities. Higher is better.

    \item \textbf{Spearman correlation.}
    The monotonic agreement between the rankings of predicted and ground-truth affinities:
    \[
        \rho =
        \mathrm{corr}\left(
            \mathrm{rank}(y),
            \mathrm{rank}(\hat{y})
        \right)
        = 1 - \frac{
            6\sum_{i=1}^{N} d_i^2
        }{
            N(N^2-1)
        }\,,
    \]
    where the second form holds without ties and $d_i$ is the rank difference of the $i$-th complex. Higher is better.

    \item \textbf{Root mean squared error (RMSE).}
    The absolute prediction error:
    \[
        \mathrm{RMSE}
        =
        \sqrt{
            \frac{1}{N}
            \sum_{i=1}^{N}
            (y_i-\hat{y}_i)^2
        }\,.
    \]
    Lower is better.
\end{itemize}

For all probing experiments in the paper, we report the mean and standard deviation over five seeds $\{0, 1, 2, 3, 4\}$, unless otherwise stated. For CASF 2016, we compute 90\% bias-corrected and accelerated (BCa) confidence intervals from $10,000$ bootstrap resamples of the test complexes, using random seed 42 \citep{efron1987better}. For methods trained with multiple seeds, the same bootstrap indices are applied to every seed, and the metric is averaged over seeds with each resample.

\subsection{Vina docking}
\label{appx:eval-vina}

We score every generated molecule with AutoDock Vina 1.2 \citep{trott2010autodock,eberhardt2021autodock}, following \citet{guan20233d-5e4,pinheiro2024structure}. AutoDock configuration is identical for every method as follows:

\begin{itemize}
    \item Version: \texttt{1.2.2}
    \item Scoring function: \texttt{vina} 
    \item Exhaustiveness: \texttt{32}
    \item Min RMSD: \texttt{1.0} \AA{} (default)
    \item Seed: \texttt{0} (random seed per run)
    \item CPU: \texttt{1} per Vina instance, \texttt{20} instances in parallel
    \item Box size: ligand extent plus \texttt{5} \AA{} per axis, $(\mathbf{x}_{\max} - \mathbf{x}_{\min})+5$\AA{}
    \item Box center: center of ligand bounding box, $(\mathbf{x}_{\max} + \mathbf{x}_{\min})/2$
    \item Receptor: full CrossDocked2020 receptor, prepared with \texttt{PDB2PQR 3.6.1} \citep{dolinsky2007pdb2pqr} with \texttt{AMBER} force field \citep{wang2000well} and AutoDockTools \citep{morris2009autodock4}
    \item Ligand: loaded with RDKit \texttt{AddHs(addCoords=True)}, then processed with Meeko \texttt{0.1.dev3} \texttt{MoleculePreparation()} \citep{santos2025meeko} with default configuration
\end{itemize}

\textbf{Configuration.}
The box follows the molecule being scored, the generated pose for a sample, and the crystal pose for the reference ligand. The mean box is $12.1\times12.0\times11.8$ \AA$^{3}$, and the per-axis edge spans $7.5-18.3$ \AA{}. \texttt{PDB2PQR} assigns the standard protonation states at pH 7.0 without PROPKA titration. Afterwards, Gasteiger charges are added, non-polar hydrogens are removed, and waters and non-standard residues are removed without structural repair. Ligand preparation keeps the generated coordinates with no re-embedding, and \text{Meeko} keeps macrocycles and amide bonds rigid. Since a seed of \texttt{0} draws a fresh seed per run, Vina Dock carries small noise while Vina Score and Vina Min are deterministic.

\textbf{Metrics.}
We generate 100 molecules per pocket and report the average and median over all generated molecules. Lower Vina scores indicate more favorable binding.

\begin{itemize}
    \item \textbf{Vina Score}: Energy of the generated pose as is (\texttt{score()}), without search or relaxation
    \item \textbf{Vina Min}: Energy after local minimization on the generated pose (\texttt{optimize()})
    \item \textbf{Vina Dock}: Energy of the pose from a global docking search within the box (\texttt{dock()})
    \item \textbf{High affinity}: Per pocket computed, the fraction of generated molecules with $E^{(i)}_{\mathrm{dock}} < E^{\mathrm{ref}}_{\mathrm{dock}}$, where $E^{\mathrm{ref}_{\mathrm{dock}}}$ is the Vina Dock energy of the reference ligand. We report its average and median over pockets.
\end{itemize}

\subsection{Generation quality}
We evaluate sample quality following the common protocols for structure-based drug design in \citet{guan20233d-5e4}, with all properties computed by RDKit \citep{landrum2025rdkit}. The configuration is as follows:

\begin{itemize}
    \item Version: \texttt{RDKit 2026.03.3}
    \item Valid molecules: sanitized by RDKit, single connected component
    \item Aggregation: within each pocket, then over pockets
\end{itemize}
    
Invalid samples are discarded before every other metric, and $n$ in \cref{tab:result-drug-design} counts the remaining molecules only. VoxBind-based methods use the original VoxBind sampler, which keeps distinct valid molecules per pocket until it collects 100 or reaches 500 reconstruction attempts.

\textbf{Metrics.}
Higher is better for every metric below.

\begin{itemize}
    \item \textbf{Validity}: Fraction of drawn samples that are valid, $N_{\mathrm{valid}}/N_{\mathrm{raw}}$
    \item \textbf{Uniqueness}: Fraction of distinct canonical SMILES among the valid molecules of a pocket, $\lvert\{\mathrm{SMILES}(m_i)\}\rvert/N_{\mathrm{valid}}$
    \item \textbf{QED}: Quantitative estimate of drug-likeness \citep{bickerton2012quantifying}, a score in $[0,1]$ summarizing physicochemical properties
    \item \textbf{SA}: Synthetic accessibility score \citep{ertl2009estimation} in the normalized form, where a higher value means easier to synthesize
    \item \textbf{Diversity}: One minus the mean pairwise Tanimoto similarity among the valid molecules of a pocket,
    \[
        \mathrm{Diversity}
        =
        1-
        \frac{2}{N_{\mathrm{valid}}(N_{\mathrm{valid}}-1)}
        \sum_{i<j}
        \mathrm{Tanimoto}(f_i,f_j)
    \]
\end{itemize}

$N_{\mathrm{raw}}$ and $N_{\mathrm{valid}}$ denote the number of drawn and valid molecules of a pocket, $m_i$ is the $i$-th valid molecule, and $f_i$ is its fingerprint.

\newpage
\subsection{Novelty computation}
\label{appx:novelty}

We measure how much of each method's output replays training chemistry, following \citet{polykovskiy2020molecular,brown2019guacamol,walters2020assessing}, with the following configuration:

\begin{itemize}
    \item Comparison set: CrossDocked2020 training split, \texttt{100,000} ligands
    \item Canonicalization: isomeric SMILES \citep{weininger1988smiles} with RDKit
    \item Scaffold: Bemis-Murcko \citep{bemis1996properties} scaffold, used with the function \texttt{MurckoScaffold.GetScaffoldForMol}
    \item SNN fingerprint: ECFP4 \citep{rogers2010extended}, radius \texttt{2}, \texttt{2048} bits
    \item Aggregation: within each pocket, then over pockets
\end{itemize}

\begin{figure}[!h]
\centering
    \includegraphics[width=.78\textwidth]{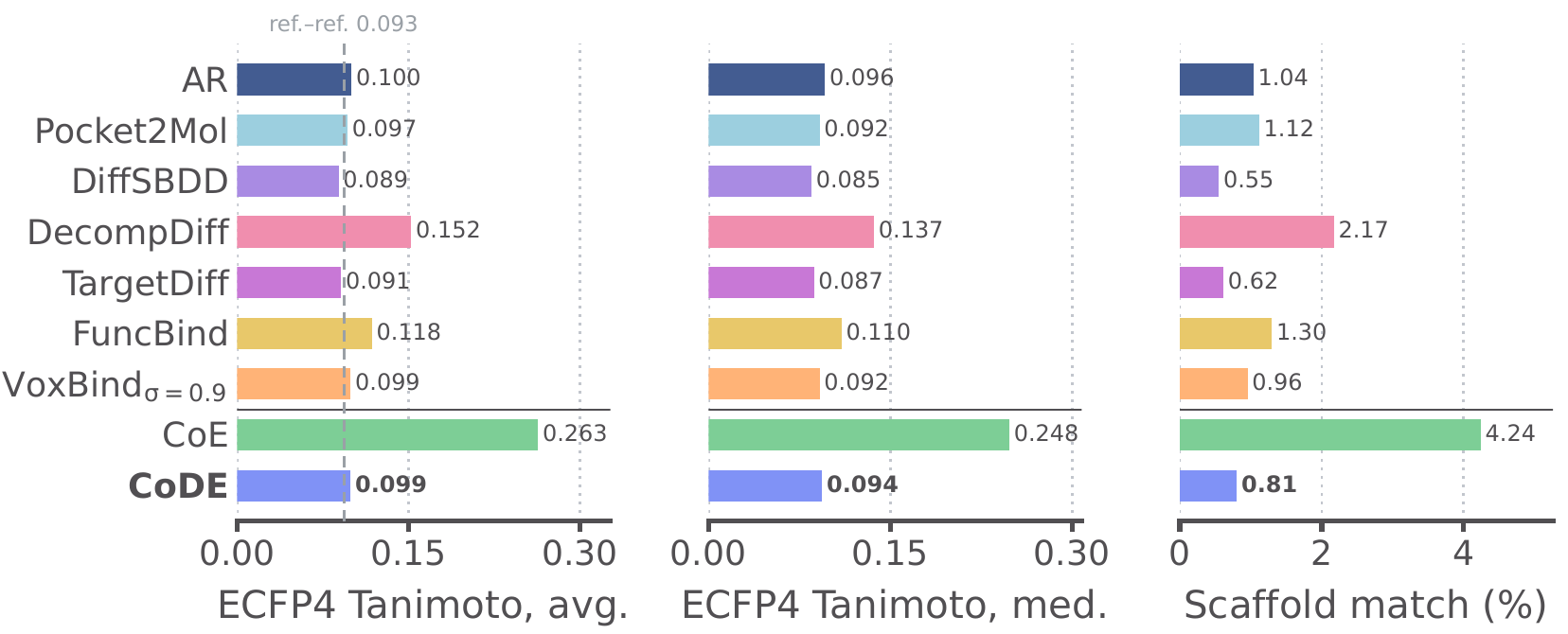}
    \caption{
        \textbf{Similarity to the test reference ligand.}
        Each generated molecule is compared with the crystal ligand of its pocket. The per-pocket mean (left) and median (middle) of the ECFP4 Tanimoto similarity are averaged over pockets. On the right is the percentage of molecules whose Bemis--Murcko scaffold equals the reference's. The dashed line marks 0.093, the mean similarity of the 79 crystal ligands to one another, which is the level of two unrelated ligands on this fingerprint.
    }
    \label{fig:appx-drug-reference-similarity}
\end{figure}

\Cref{fig:result-drug-reference-novelty} reports novelty and SNN against the training split. \Cref{fig:appx-drug-reference-similarity} separately compares each generated molecule with the reference ligand of its corresponding test pocket.

\textbf{Metrics.}
We use the following metrics to measure the distances to the training and the test set reference ligand similarity.

\begin{itemize}
    \item \textbf{Molecule novelty}: Fraction of generated molecules whose canonical SMILES does not exist in the training split.
    \item \textbf{Scaffold novelty}: Same fraction computed with the Bemis-Murcko scaffolds
    \item \textbf{SNN}: Mean over the pocket of the largest Tanimoto similarity \citep{bajusz2015tanimoto} of each generated molecule $g$ to any training ligand $t$, 
    \[
        \mathrm{SNN}
        =
        \frac{1}{\lvert G\rvert}
        \sum_{g \in G}
        \max_{t \in T}
        \mathrm{Tanimoto}\bigl(f(g), f(t)\bigr),
    \]
    which catches the near-duplicates that exact matching misses.
\end{itemize}

$G$ denotes the set of generated molecules of a pocket, $T$ is the comparison set from the training data, and $f(\cdot)$ is the ECFP4 fingerprint.

\newpage
\subsection{PoseCheck}

We use PoseCheck \citep{harris2023posecheck} to check the realistic structures of the generated molecules, with (i) strain energy, (ii) steric clashes, (iii) re-docking, and (iv) protein--ligand interactions of the generated poses. The configuration is as follows:

\begin{itemize}
    \item Version: \texttt{1.3.1}
    \item Receptor: \texttt{10} \AA{} pocket crop, protonated with reduce
    \item Clash tolerance: \texttt{0.5} \AA{} (default)
    \item Strain conformers: \texttt{50} per molecule, embedded with seed \texttt{0}
    \item Strain timeout: \texttt{50} seconds per molecule
    \item Interactions: ProLIF \texttt{2.2.0} \texttt{Fingerprint()} with default configuration
    \item Re-docking: Vina settings stated in \cref{appx:eval-vina}
    \item Re-docking set: \texttt{13} molecules per pockets in generation order, over \texttt{78} pockets
    \item Re-docking RMSD: RDKit \texttt{rdMolAlign.CalcRMS} on heavy atoms, without superposition
\end{itemize}

Water and HETATM records are removed from the pocket before protonation. Molecules whose strain does not converge within the timeout are dropped. One excluded re-docking pocket has a chain gap that \texttt{PDB2PQR} cannot protonate, for all methods. We use PoseCheck \texttt{1.3.1} since prior releases differ in the strain energy definition, which uses the energy difference between the generated pose and the relaxed pose, rather than the relaxed pose to searched conformer poses. Our values are internally consistent, but not comparable with the numbers printed in \citet{harris2023posecheck}.

\textbf{Metrics.}
Lower is better for clashes, strain, and re-docking displacement. For protein--ligand interactions, values similar to the reference are considered better.

\begin{itemize}
    \item \textbf{Steric clashes}: Number of ligand atoms per molecule with ligand and protein atoms overlapping
    \item \textbf{Strain energy}: UFF energy of the relaxed pose relative to the best conformer,
        \[
            E_{\mathrm{strain}}
            =
            E\bigl(\mathrm{UFF}_{\mathrm{constrained}}(x)\bigr)
            -
            \min\Bigl(
            \min_{k=1}^{50} E\bigl(\mathrm{UFF}(c_k)\bigr),\;
            E\bigl(\mathrm{UFF}_{\mathrm{free}}(x)\bigr)
            \Bigr)\,,
        \]
        where $x$ is relaxed under a 0.1 \AA{} per atom constraint and $c_{k}$ are the re-embedded conformers.
    \item \textbf{Protein--ligand interactions}: Number of ProLIF fingerprint bits per molecule for van der Waals contacts, hydrophobic contacts, and hydrogen bond donors and acceptors.
    \item \textbf{Re-docking RMSD}: Heavy atom RMSE between the generated pose and the top Vina pose. It is measured without superposition. Focuses on the displacement in the pocket.
\end{itemize}

$d_{ij}$ denotes the distance between ligand atom $i$ and the receptor atom $j$, and $R^{\mathrm{vdw}}$ denotes the van der Waals radius. Note that the crystal ligands re-docked by the same protocol result in a median of 2.69 \AA{}, with $42.3\%$ within 2\AA{}.

\newpage
\subsection{Molecular structure}

We compare the distributions of atom types, bond lengths, atom-pair distances, and ring statistics against reference ligands, following \citet{guan20233d-5e4,pinheiro2024structure}. For each descriptor, we histogram the generated and reference values on a fixed grid and compute the Jensen-Shannon distance \citep{lin1991divergence} with SciPy \citep{virtanen2020scipy}. The configuration is as follows:

\begin{itemize}
    \item Version: SciPy \texttt{1.15.2}, \texttt{2024.03.06}
    \item Reference: CrossDocked 2020 train and test ligands, \texttt{100,100} files
    \item Atom types: C, N, O, F, P, S, and Cl
    \item Atom-pair distances: all heavy atom pairs below 12 \AA{} and C--C pairs below 2\AA{}, 100 bins each
    \item Ring size: rings of each size from 3 to 9
    \item Rings per molecule: exact counts
    \item Aromaticity; aromatic share of heavy atoms in 20 bins and of rings in 10 bins
\end{itemize}

Only valid, single-component molecules are included. We use all CrossDocked 2020 ligands as the reference, as in \citet{guan20233d-5e4}, since 100 test ligands are too few to estimate the bond-length histogram. 8,765 distinct molecules are used from 100,100 files in total.

\subsection{PoseBusters} 
We use PoseBusters \citep{buttenschoen2024posebusters} to check the physical validity of the generated poses without relaxation. We run its \texttt{dock} configuration, consisting of 22 tests and intended for a pose inside a receptor without a ground-truth ligand. The configuration is as follows:

\begin{itemize}
    \item Version: PoseBusters \texttt{0.6.5}
    \item Mode: \texttt{dock}
    \item Receptor: \texttt{10} \AA{} pocket crop, identical for every method
    \item Distance geometry: bound thresholds \texttt{0.25} for bond lengths, \texttt{0.25} for bond angles, and \textbf{0.3} for internal clashes, ignoring hydrogens
    \item Flatness: maximum deviation \texttt{0.25} \AA{} for aromatic rings and trigonal-planar double bonds
    \item Non-flatness: minimum deviation \texttt{0.05} \AA{} for non-aromatic six-member rings
    \item Internal energy: energy-ratio threshold \texttt{100} against \texttt{50} generated conformers
    \item Minimum distance: clash cutoff \texttt{0.75}, maximum distance \texttt{5.0} \AA{}
    \item Volume overlap: clash cutoff \texttt{0.75}, van der Waals scale \texttt{0.8}, and \texttt{0.5} for inorganic co-factors and waters
\end{itemize}

\texttt{dock} mode tests 22 cases, where two are whether input files are loaded properly, and 20 are graded checks. The pocket crop contains no co-factors or waters, so four co-factor and two water checks pass by construction. The list of test cases is as follows.

\begin{itemize}
    \item \textbf{Overall validity}: Percentage of molecules passing all 22 checks
    \item \textbf{Load (2)}: Valid loading of files for evaluation
    \item \textbf{Length (1)}: Bond lengths within distance-geometry bounds
    \item \textbf{Angle (1)}: Bond angle within distance-geometry bounds
    \item \textbf{Planarity (2)}: Flat aromatic rings and flat double bonds
    \item \textbf{Pucker (1)}: Non-planar non-aromatic six-member rings
    \item \textbf{Valence (4)}: RDKit sanitization, InChI conversion, a single connected component, and no radicals
    \item \textbf{Clash (9)}: Protein proximity, and minimum distance and volume overlap against protein, organic co-factors, inorganic co-factors, and waters
    \item \textbf{Self-clash (2)}: Internal steric clash and internal energy
\end{itemize}

\subsection{Rigid-fragment consistency}

We measure whether rigid fragments keep their force-field geometry, following \citet{guan20233d-5e4}. A rigid fragment, such as a benzene ring, an amide, or a fused bicycle, has one correct shape, so any deviation is a generation error. We relax each molecule with the Merck molecular force field \citep[MMFF]{halgren1996merck} in RDKit, cut it at rotatable bonds, and align each fragment with the Kabsch algorithm \citep{kabsch1976solution}. The configuration is as follows: 

\begin{itemize}
    \item Version: RDKit 2024.03.06
    \item Force Field: \texttt{MMFF}
    \item Fragmentation of every rotatable bond
    \item Alignment per fragment: Kabsch superposition on heavy atoms
    \item Size bins: fragments' heavy atoms
    \item Reference: crystal ligands of the test pockets, through the same pipeline.
\end{itemize}

\textbf{Metrics.}
Lower is better, indicating that the rigid fragment did not move much.

\begin{itemize}
    \item \textbf{Rigid-fragment RMSD}: Heavy-atom RMSD between a fragment's generated coordinates $\mathbf{x}_{i}$ and relaxed coordinates $\mathbf{y}_{i}$ after superposition over rotations $R$ and translation $\mathbf{t}$,
    \[
        \mathrm{RMSD} =
        \min_{R,\mathbf{t}}
        \sqrt{
            \frac{1}{n}
            \sum_{i=1}^{n}
            \lVert R\mathbf{x}_{i} + \mathbf{t} - \mathbf{y}_{i} \rVert^{2}
        }\,,
    \]
    summarized by its median within each size bin.
\end{itemize}

$n$ denotes the number of heavy atoms in the fragment.

%% file: ICLR2027/appx/4_additional_result.tex
\section{Additional results}
\label{appx:results}

\subsection{Binding affinity regression}
We additionally report the CL1- and CL2-filtered test set and CASF 2016 results. $\dagger$ marks the affinity-leaked reference, included for context only. \textbf{Bold} marks the best and \underline{underlining} marks the next-best; methods whose intervals overlap are marked together. N/A denotes unavailable metrics, and \textbf{Pre.} indicates methods with external-data pre-training.

\input{ICLR2027/resource/table/result-binding-affinity-cl12}
\input{ICLR2027/resource/table/result-binding-affinity-casf}


\newpage
\subsection{Vina docking}
We present plots with all the baselines for the Vina docking regarding the target pocket and number of heavy atoms, extended from \cref{fig:result-drug-pocketwise,fig:result-drug-scoreperatom}. \ours{} still shows better performance in a majority of pockets. Also, all methods score worse when there are fewer heavy atoms, while \ours{} scores better in many regions with many heavy atoms.

\input{ICLR2027/resource/figure/appx/result-drug-pocketwise-full}
\input{ICLR2027/resource/figure/appx/result-drug-atomwise-full}

\newpage
Also, we divide \cref{fig:result-drug-pocketwise-all} by \ours{} for better visualization.
\input{ICLR2027/resource/figure/appx/result-drug-pocketwise-expanded}

\newpage
\subsection{Qualitative visualization}
\Cref{fig:result-drug-example-qual-all} extends \cref{fig:result-drug-example-qual} with four additional pockets (\texttt{5TJN}, \texttt{1RS9}, \texttt{4KCQ}, and \texttt{1COY}). For each pocket and method, we visualize the generated ligand with the lowest Vina docking score, annotated at the bottom right. \ours{} yields the best scores, and VoxBind follows it; thus, the two voxel-based models occupy the binding site more fully than other Autoregressive and diffusion baselines. This aligns with the trend in \cref{tab:result-drug-design} and \cref{fig:result-drug-pocketwise}, where density conditioning steers ligands towards poses that fit the pocket. Note that these are all the best-scoring samples per method; they illustrate rather than quantify the gains.

\input{ICLR2027/resource/figure/appx/result-drug-qual-all}

\newpage
\subsection{Macrocycles in ligand generation}
\label{appx:macro}

We ask whether the favorable docking scores of generated macrocycles come with favorable geometry. We define a macrocycle as a molecule with a ring of at least 12 atoms in RDKit's SymmSSSR, and characterize it by the size of that largest ring, which counts ring atoms rather than all heavy atoms in the molecule. \Cref{fig:result-drug-macrocycle-quality} compares TargetDiff, VoxBind, and \ours{} on the same 79 test pockets. At every shared ring size, \ours{} attains lower median strain, fewer clashes, and a higher PoseBusters pass rate than TargetDiff, and its median Vina Dock score is the lowest of the three for ring sizes of 18--21 atoms; against VoxBind the comparison varies with ring size.

These curves are descriptive rather than controlled. The three methods generate macrocycles in different pocket subsets (30, 40, and 60 pockets, with 24 shared by \ours{} and VoxBind), and conditioning on ring size matches neither heavy-atom count nor pocket composition; a comparison matched on both leaves too few strata to support the pooled advantage over VoxBind. We also note that PoseCheck and PoseBusters assess the generated poses whereas Vina Dock scores follow re-docking, so passing PoseBusters does not validate the re-docked pose.

\input{ICLR2027/resource/figure/appx/result-drug-macrocycle}

\newpage
\subsection{PoseCheck}
We present additional results for PoseCheck on (i) strain energy against the number of rotatable bonds, and (ii) protein--ligand interactions. In \cref{fig:result-drug-posecheck-strain-rot}, FuncBind shows exceptionally high strain energy when the number of rotatable bonds increases. AR, DiffSBDD, and TargetDiff also show relatively high strain energy when the number of rotatable bonds increases, compared to Pocket2Mol, DecompDiff, and VoxBind. In \cref{fig:result-drug-posecheck-interaction}, we present the protein--ligand interaction regarding contacts and H-bonds. \ours{} does not degrade the performance compared to VoxBind in terms of both contacts and H-bonds.

\input{ICLR2027/resource/figure/appx/result-drug-posecheck-strain-rot}
\input{ICLR2027/resource/figure/appx/result-drug-posecheck-interaction}

\newpage
Additionally, we plot a qualitative example of re-docking in \cref{fig:result-pc-redock-all}. For each method, we show the ligand whose pose changes least under re-docking, i.e., the ligand with the lowest re-docking RMSD among the ligands re-docked per pocket, where the RMSD is computed over heavy atoms in place, without superposition. As this pocket is one where \ours{} attains the lowest re-docking RMSD, the example illustrates rather than quantifies; the comparison over all pockets is given in \cref{fig:result-drug-posecheck-redock}.

\input{ICLR2027/resource/figure/appx/result-pc-redocking}

\newpage
\subsection{Molecular structure}
We report the detailed Jensen-Shannon distance \citep[JSD]{lin1991divergence} for molecular structures, such as atom types, bond distances, atom pair distances, and ring statistics. 

\begin{itemize}
    \item Heavy atom types: \cref{fig:result-drug-jsd-atom-type}
    \item Atom pair distance under 12\AA{}: \cref{fig:result-drug-jsd-pair}
    \item Bond distance under 2\AA{}: \cref{fig:result-drug-jsd-bond-distance}
    \item Overall ring statistics: \cref{fig:result-drug-jsd-rings}
    \item Ring size: \cref{fig:result-drug-jsd-ring-size}
    \item Number of rings: \cref{fig:result-drug-jsd-n-rings}
    \item Aromatic rings: \cref{fig:result-drug-jsd-aromatic}
\end{itemize}

\input{ICLR2027/resource/figure/appx/result-jsd-atom-type}
\input{ICLR2027/resource/figure/appx/result-jsd-pair}

\newpage
\input{ICLR2027/resource/figure/appx/result-jsd-bond-distance}
\input{ICLR2027/resource/figure/appx/result-jsd-rings}

\newpage
\input{ICLR2027/resource/figure/appx/result-jsd-ring-size}
\input{ICLR2027/resource/figure/appx/result-jsd-n-rings}
\input{ICLR2027/resource/figure/appx/result-jsd-aromatic}

\newpage
\subsection{Rigid fragment consistency}
We present extended results from \cref{fig:result-drug-rigid-consistency}, regarding the number of rigid fragments and the distribution of generated molecules for each method.

\input{ICLR2027/resource/figure/appx/result-rigid-expanded}
\input{ICLR2027/resource/figure/appx/result-rigid-distribution}

\newpage
\subsection{PoseBusters}
\label{appx:pb}

We present PoseBusters failure cases across 22 tasks in \texttt{dock} mode, with an upset plot showing correlations between categories.

\input{ICLR2027/resource/figure/appx/result-pb-all}
\input{ICLR2027/resource/figure/appx/result-pb-upset}

\textbf{Ring pucker failure analysis.}
The \textit{Pucker} column of \cref{fig:result-drug-posebusters} is the single PoseBusters test \texttt{non-aromatic\_ring\_non-flatness}. It fails a molecule when a non-aromatic six-membered ring has no atom at least $0.05$\,\AA{} from the ring's best-fit plane; five-membered rings are not tested. VoxBind, \ours{}, TargetDiff, and FuncBind generate atoms without bonds, and all four assign bonds afterwards with the same OpenBabel-based construction \citep{o2011open}, which does not use aromaticity. To assess the contribution of molecular reconstruction, we classify failing molecules into four diagnostic categories. These categories describe geometric and reconstruction properties; they do not establish a unique underlying cause for each failure.

\begin{itemize}
    \item \textbf{Flat non-aromatic ring.} A genuinely planar saturated or partially saturated ring, which is the failure the test is designed to catch.
    \item \textbf{Signed-distance rule.} A ring atom lies at least $0.05$\,\AA{} from the plane, so the ring is puckered; PoseBusters nevertheless fails it because it compares the largest \emph{signed} distance with the threshold, and the orientation of the plane normal is arbitrary.
    \item \textbf{Benzene-like, no valid aromatic assignment.} The failing ring has benzene-like geometry (ring bonds between $1.20$ and $1.60$\,\AA{}, mean internal angle between $116^\circ$ and $124^\circ$, every atom within $0.1$\,\AA{} of the plane), but re-assigning it as aromatic cannot be kekulized with the rest of the molecule.
    \item \textbf{Valid aromatic reassignment.} The failing ring meets the benzene-like geometric criteria above, and an alternative aromatic bond assignment yields a valid, kekulizable molecule after adjusting attached hydrogens. Such cases are consistent with a mismatch between the generated geometry and the reconstructed molecular graph. However, the availability of an aromatic alternative does not establish that the original ring was intended to be aromatic.
\end{itemize}

For \ours{}, 87.3\% of the failing molecules fall into the valid-aromatic-reassignment category, while 11.6\% have benzene-like geometry without a valid aromatic assignment under our diagnostic procedure. The flat non-aromatic category accounts for 0.9\% of its failing molecules, or 0.20\% of all generated molecules, compared with 0.23\% of all generated molecules for VoxBind. These results suggest that molecular reconstruction contributes substantially to the observed ring non-flatness failures. The categories are diagnostic, however, and do not establish the physical validity of the original molecular graphs.

\input{ICLR2027/resource/figure/appx/result-pb-analysis}

\Cref{fig:qual-diene-vs-benzene} shows that alternative bond-order and hydrogen assignments can change the ring non-flatness outcome for the same heavy-atom geometry. It reduces the ring non-flatness failure rate of \ours{} (\cref{tab:result-pb-rearomatize-change}), with modest changes in the aggregated molecular properties. Also, in the evaluated docking subset, the Vina score changes are small (\cref{tab:result-pb-rearomatize-vina}). We interpret these results as sensitivity to molecular reconstruction, rather than as validation of the original molecular graphs.

\input{ICLR2027/resource/table/appx/appx-pb-change}
\input{ICLR2027/resource/figure/appx/result-pb-example}

%% file: ICLR2027/resource/table/result-binding-affinity-cl12.tex
\begin{table}[!ht]
    \centering
    \caption{
        \textbf{LP-PDBBind binding-affinity prediction on the \textit{CL1-filtered} and \textit{CL2-filtered} test sets.} 
        Results are averaged over five seeds, with zero-shot scorers reported with a single prediction
    }
    \label{tab:result-binding-affinity-cl1-cl2}
    \setlength{\tabcolsep}{4.5pt}
    \renewcommand{\arraystretch}{1.06}
    \setlength{\heavyrulewidth}{1pt}
    \setlength{\lightrulewidth}{0.4pt}
    \resizebox{.98\textwidth}{!}{%
        \begin{tabular}{llccccccc}
            \toprule
            & & & \multicolumn{3}{c}{\textbf{CL1 filtered (N=1166)}} & \multicolumn{3}{c}{\textbf{CL2 filtered (N=1149)}} \\
            \cmidrule(lr){4-6}\cmidrule(lr){7-9}
            \textbf{Input} & \textbf{Method} & \textbf{Pre.} & \textbf{Pearson \textit{r}} & \textbf{Spearman $\rho$} & \textbf{RMSE $\downarrow$} & \textbf{Pearson \textit{r}} & \textbf{Spearman $\rho$} & \textbf{RMSE $\downarrow$} \\
            \midrule
            \multirow{4}{*}{Seq.}
            & DeepDTA & \xmark & 0.554\std{0.008} & 0.508\std{0.009} & 1.493\std{0.008} & 0.541\std{0.013} & 0.495\std{0.006} & 1.491\std{0.018} \\
            & MolTrans & \xmark & 0.499\std{0.036} & 0.467\std{0.034} & 1.564\std{0.051} & 0.513\std{0.027} & 0.478\std{0.026} & 1.528\std{0.040} \\
            & HonestAffinity & \cmark & 0.477\std{0.025} & 0.453\std{0.025} & 1.598\std{0.020} & 0.483\std{0.033} & 0.450\std{0.034} & 1.580\std{0.032} \\
            & Nesso-1$^\dagger$ & \cmark & 0.703 & 0.694 & 1.367 & 0.707 & 0.698 & 1.354 \\
            \midrule
            \multirow{13}{*}{Struc.}
            & HBGSA & \xmark & 0.554\std{0.029} & 0.537\std{0.017} & 1.628\std{0.169} & 0.567\std{0.021} & 0.541\std{0.015} & 1.953\std{0.653} \\
            & EGNN & \xmark & 0.610\std{0.006} & 0.590\std{0.005} & 1.413\std{0.005} & 0.621\std{0.009} & 0.597\std{0.008} & 1.384\std{0.013} \\
            & EGNN\textsubscript{\scriptsize TargetDiff} & \xmark & 0.609\std{0.005} & 0.589\std{0.004} & 1.412\std{0.011} & 0.616\std{0.005} & 0.591\std{0.005} & 1.389\std{0.006} \\
            & GET & \xmark & 0.611\std{0.003} & 0.591\std{0.004} & 1.410\std{0.005} & 0.619\std{0.004} & 0.595\std{0.002} & 1.389\std{0.004} \\
            & CheapNet & \xmark & 0.644\std{0.007} & 0.609\std{0.009} & \underline{1.409\std{0.025}} & 0.645\std{0.012} & 0.608\std{0.013} & \underline{1.378\std{0.029}} \\
            & IPNet & \xmark & 0.603\std{0.010} & 0.567\std{0.010} & 1.424\std{0.014} & 0.610\std{0.022} & 0.578\std{0.020} & 1.406\std{0.033} \\
            & IPNet$^\dagger$ & \cmark & 0.594\std{0.015} & 0.557\std{0.013} & 1.455\std{0.015} & 0.607\std{0.012} & 0.563\std{0.011} & 1.421\std{0.014} \\
            & AEV-PLIG & \xmark & 0.595\std{0.013} & 0.562\std{0.015} & 1.479\std{0.022} & 0.589\std{0.015} & 0.556\std{0.013} & 1.466\std{0.026} \\
            & DSMBind & \cmark & 0.448 & 0.445 & N/A & 0.449 & 0.445 & N/A \\
            & BindNet & \cmark & 0.533\std{0.024} & 0.504\std{0.022} & 1.577\std{0.104} & 0.545\std{0.025} & 0.516\std{0.022} & 1.626\std{0.094} \\
            & ProFSA & \cmark & \underline{0.645\std{0.005}} & \underline{0.612\std{0.005}} & 1.473\std{0.021} & \underline{0.651\std{0.007}} & \underline{0.615\std{0.007}} & 1.445\std{0.053} \\
            \cmidrule(lr){2-9}
            \rowcolor{AffAccent!12} \cellcolor{white} & & \xmark & 0.583\std{0.024} & 0.567\std{0.012} & 1.498\std{0.059} & 0.591\std{0.025} & 0.573\std{0.013} & 1.478\std{0.059} \\
            \rowcolor{AffAccent!12} \cellcolor{white} & \multirow{-2}{*}{\oursC{}} & \cmark & 0.635\std{0.001} & 0.606\std{0.000} & \textbf{1.418\std{0.048}} & 0.646\std{0.001} & 0.612\std{0.000} & 1.396\std{0.031} \\
            \midrule
            \rowcolor{AffAccent!28} Struc.\,, & & \xmark & 0.587\std{0.018} & 0.571\std{0.011} & 1.476\std{0.042} & 0.594\std{0.019} & 0.576\std{0.012} & 1.458\std{0.044} \\
            \rowcolor{AffAccent!28} Density & \multirow{-2}{*}{\ours{} (Ours)} & \cmark & \textbf{0.671\std{0.002}} & \textbf{0.652\std{0.001}} & \textbf{1.380\std{0.054}} & \textbf{0.671\std{0.002}} & \textbf{0.650\std{0.001}} & \textbf{1.336\std{0.022}} \\
            \bottomrule
        \end{tabular} %
    }
\end{table}

%% file: ICLR2027/resource/table/result-binding-affinity-casf.tex
\begin{table}[!ht]
    \centering
    \caption{
        \textbf{CASF-2016 held-out binding-affinity prediction.} 90\% confidence intervals (CI) using BCa are reported.
    }
    \label{tab:result-binding-affinity-casf}
    \setlength{\tabcolsep}{4.5pt}
    \renewcommand{\arraystretch}{1.06}
    \setlength{\heavyrulewidth}{1pt}
    \setlength{\lightrulewidth}{0.4pt}
    \resizebox{.98\textwidth}{!}{%
        \begin{tabular}{llccccccc}
            \toprule
            & & & \multicolumn{6}{c}{\textbf{CASF-2016 clean held-out (N\,=\,92)}} \\
            \cmidrule(lr){4-9}
            \textbf{Input} & \textbf{Method} & \textbf{Pre.} & \textbf{Pearson \textit{r}} & \textbf{90\% CI} & \textbf{Spearman $\rho$} & \textbf{90\% CI} & \textbf{RMSE $\downarrow$} & \textbf{90\% CI} \\
            \midrule
            \multirow{4}{*}{Seq.}
            & DeepDTA & \xmark & 0.520\std{0.054} & {\scriptsize $[0.39,\,0.63]$} & 0.488\std{0.037} & {\scriptsize $[0.35,\,0.61]$} & 1.735\std{0.058} & {\scriptsize $[1.57,\,1.93]$} \\
            & MolTrans & \xmark & 0.529\std{0.026} & {\scriptsize $[0.41,\,0.63]$} & 0.500\std{0.035} & {\scriptsize $[0.37,\,0.61]$} & 1.732\std{0.034} & {\scriptsize $[1.57,\,1.93]$} \\
            & HonestAffinity & \cmark & 0.477\std{0.071} & {\scriptsize $[0.36,\,0.58]$} & 0.432\std{0.074} & {\scriptsize $[0.30,\,0.55]$} & 1.777\std{0.085} & {\scriptsize $[1.63,\,1.95]$} \\
            & Nesso-1$^\dagger$ & \cmark & 0.621 & {\scriptsize $[0.47,\,0.73]$} & 0.631 & {\scriptsize $[0.49,\,0.74]$} & 1.721 & {\scriptsize $[1.50,\,2.04]$} \\
            \midrule
            \multirow{13}{*}{Struc.}
            & HBGSA & \xmark & 0.598\std{0.021} & {\scriptsize $[0.48,\,0.68]$} & 0.586\std{0.023} & {\scriptsize $[0.45,\,0.69]$} & 1.800\std{0.189} & {\scriptsize $[1.63,\,2.01]$} \\
            & EGNN & \xmark & 0.634\std{0.021} & {\scriptsize $[0.51,\,0.73]$} & 0.629\std{0.017} & {\scriptsize $[0.50,\,0.73]$} & 1.591\std{0.024} & {\scriptsize $[1.42,\,1.79]$} \\
            & EGNN\textsubscript{\scriptsize TargetDiff} & \xmark & 0.643\std{0.019} & {\scriptsize $[0.52,\,0.74]$} & 0.636\std{0.016} & {\scriptsize $[0.51,\,0.73]$} & 1.588\std{0.027} & {\scriptsize $[1.42,\,1.79]$} \\
            & GET & \xmark & 0.632\std{0.017} & {\scriptsize $[0.51,\,0.73]$} & 0.615\std{0.011} & {\scriptsize $[0.48,\,0.72]$} & 1.584\std{0.026} & {\scriptsize $[1.41,\,1.80]$} \\
            & CheapNet & \xmark & \underline{0.681\std{0.035}} & {\scriptsize $[0.57,\,0.77]$} & \textbf{0.649\std{0.037}} & {\scriptsize $[0.52,\,0.75]$} & 1.539\std{0.067} & {\scriptsize $[1.36,\,1.76]$} \\
            & IPNet & \xmark & 0.568\std{0.024} & {\scriptsize $[0.45,\,0.66]$} & 0.548\std{0.016} & {\scriptsize $[0.41,\,0.66]$} & 1.714\std{0.029} & {\scriptsize $[1.54,\,1.94]$} \\
            & IPNet & \cmark & 0.672\std{0.021} & {\scriptsize $[0.57,\,0.75]$} & 0.623\std{0.025} & {\scriptsize $[0.50,\,0.72]$} & 1.540\std{0.019} & {\scriptsize $[1.38,\,1.75]$} \\
            & AEV-PLIG & \xmark & 0.592\std{0.021} & {\scriptsize $[0.49,\,0.69]$} & 0.535\std{0.028} & {\scriptsize $[0.41,\,0.64]$} & 1.737\std{0.054} & {\scriptsize $[1.58,\,1.93]$} \\
            & DSMBind & \cmark & 0.658 & {\scriptsize $[0.55,\,0.74]$} & 0.653 & {\scriptsize $[0.52,\,0.75]$} & N/A & \textendash \\
            & BindNet & \cmark & 0.573\std{0.030} & {\scriptsize $[0.43,\,0.68]$} & 0.557\std{0.022} & {\scriptsize $[0.40,\,0.68]$} & 1.696\std{0.061} & {\scriptsize $[1.54,\,1.90]$} \\
            & ProFSA & \cmark & \textbf{0.712\std{0.011}} & {\scriptsize $[0.61,\,0.78]$} & \underline{0.672\std{0.009}} & {\scriptsize $[0.55,\,0.76]$} & 1.693\std{0.018} & {\scriptsize $[1.52,\,1.91]$} \\
            \cmidrule(lr){2-9}
            \rowcolor{AffAccent!12} \cellcolor{white} & & \xmark & 0.610\std{0.048} & {\scriptsize $[0.51,\,0.69]$} & 0.581\std{0.044} & {\scriptsize $[0.47,\,0.68]$} & 1.712\std{0.130} & {\scriptsize $[1.57,\,1.89]$} \\
            \rowcolor{AffAccent!12} \cellcolor{white} & \multirow{-2}{*}{\oursC{}} & \cmark & 0.676\std{0.008} & {\scriptsize $[0.57,\,0.76]$} & 0.623\std{0.008} & {\scriptsize $[0.49,\,0.73]$} & \underline{1.528\std{0.016}} & {\scriptsize $[1.37,\,1.69]$} \\
            \midrule
            \rowcolor{AffAccent!28} Struc.\,, & & \xmark & 0.617\std{0.031} & {\scriptsize $[0.51,\,0.71]$} & 0.580\std{0.026} & {\scriptsize $[0.46,\,0.68]$} & 1.637\std{0.092} & {\scriptsize $[1.50,\,1.82]$} \\
            \rowcolor{AffAccent!28} Density & \multirow{-2}{*}{\ours{} (Ours)} & \cmark & \textbf{0.720\std{0.025}} & {\scriptsize $[0.62,\,0.79]$} & \textbf{0.683\std{0.024}} & {\scriptsize $[0.56,\,0.77]$} & \textbf{1.435\std{0.039}} & {\scriptsize $[1.29,\,1.61]$} \\
            \bottomrule
        \end{tabular} %
    }
\end{table}

%% file: ICLR2027/resource/figure/appx/result-drug-pocketwise-full.tex
\begin{figure}[!h]
\centering
    \includegraphics[width=.78\textwidth]{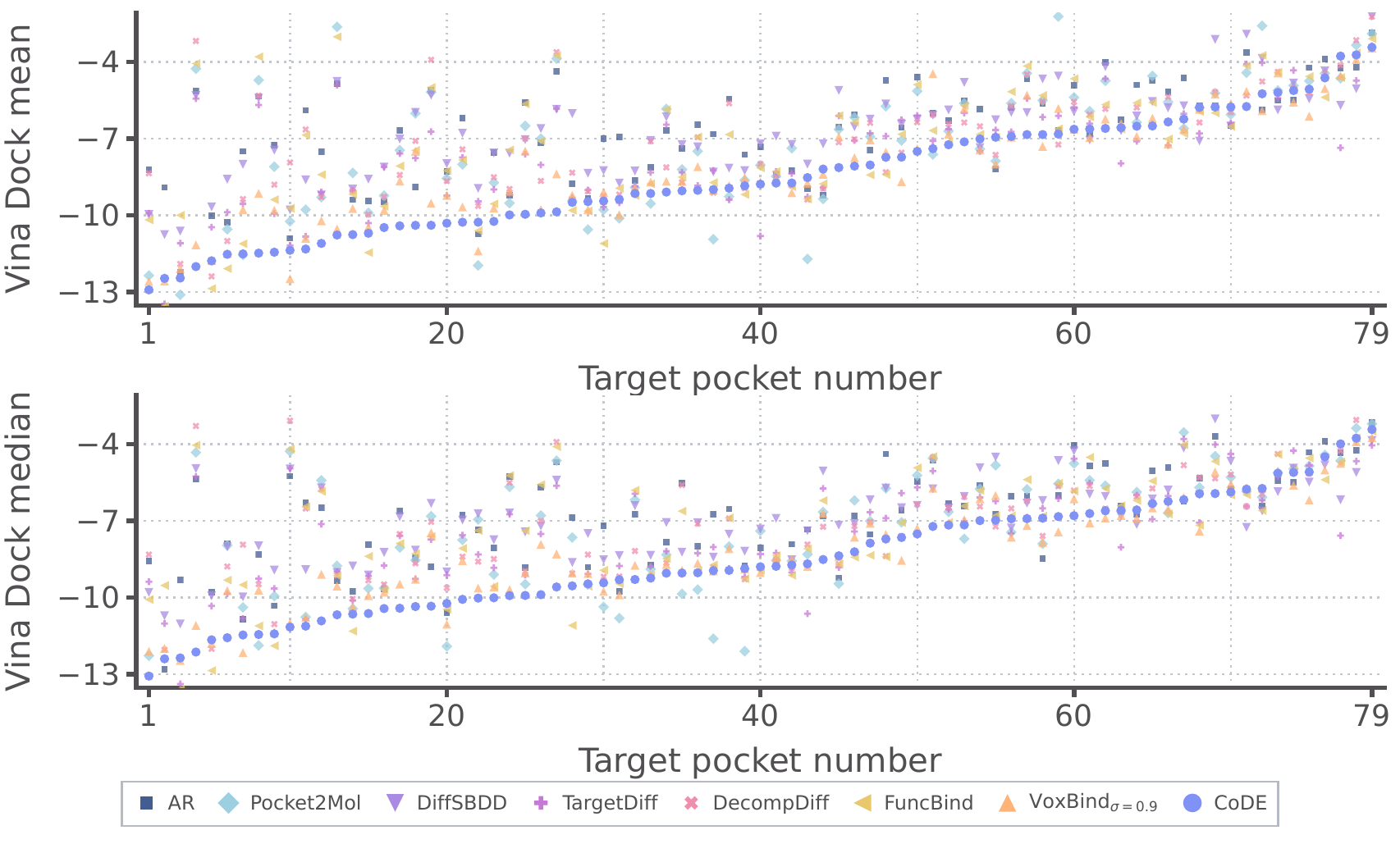}
    \caption{
        \textbf{Mean and median Vina docking scores per pocket.}
        of all generated molecules, with 79 test pockets having experimental density maps, sorted by the \ours{} Vina docking score.
    }
    \label{fig:result-drug-pocketwise-all}
\end{figure}

%% file: ICLR2027/resource/figure/appx/result-drug-atomwise-full.tex
\begin{figure}[!h]
\centering
    \includegraphics[width=.88\textwidth]{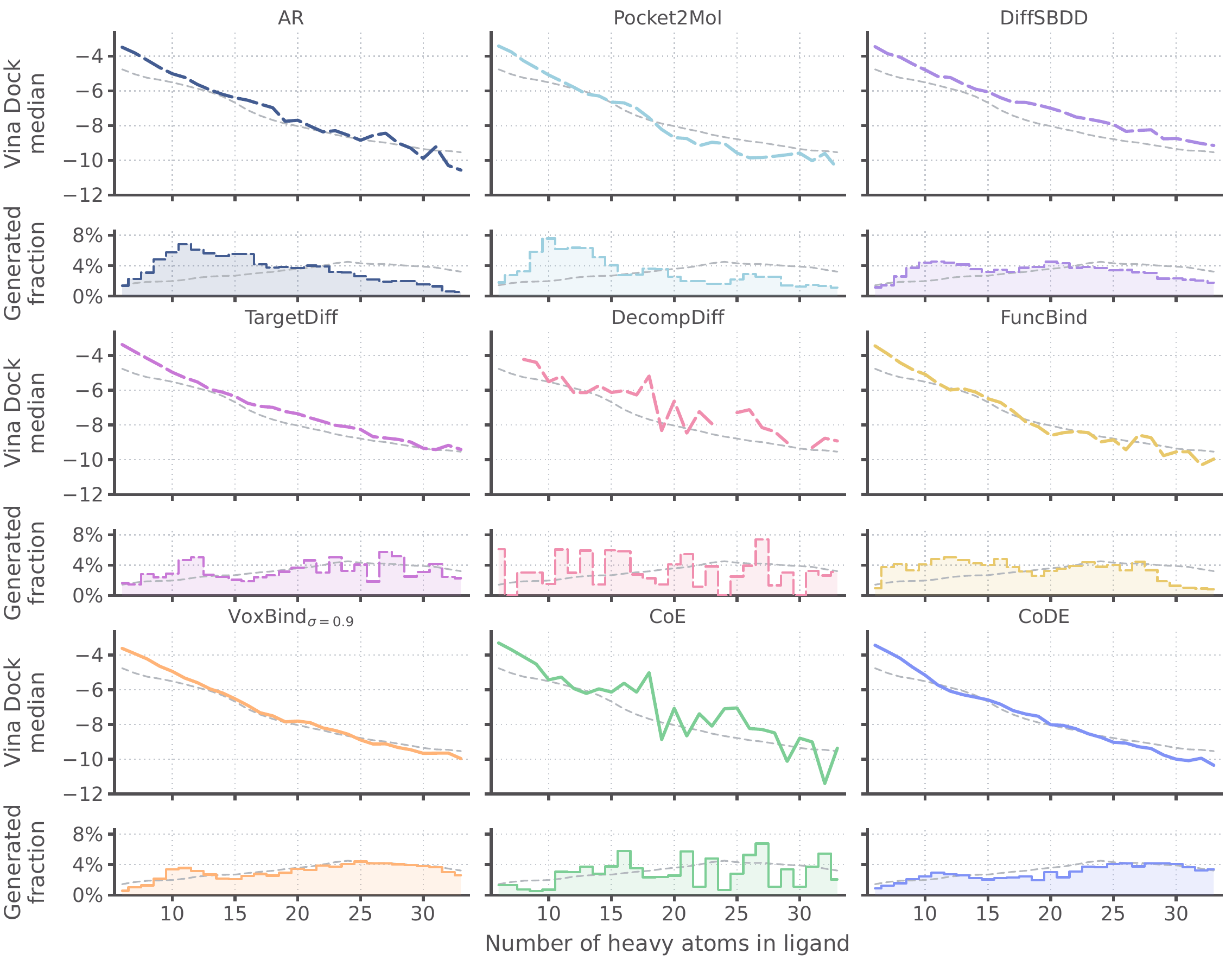}
    \caption{
        \textbf{Median Vina docking scores per number of heavy atoms.}
        of all generated molecules, with 79 test pockets having experimental density maps, sorted by the \ours{} Vina docking score.
    }
    \label{fig:result-drug-atomwise-all}
\end{figure}

%% file: ICLR2027/resource/figure/appx/result-drug-pocketwise-expanded.tex
\begin{figure}[!ht]
\centering
    \includegraphics[width=.98\textwidth]{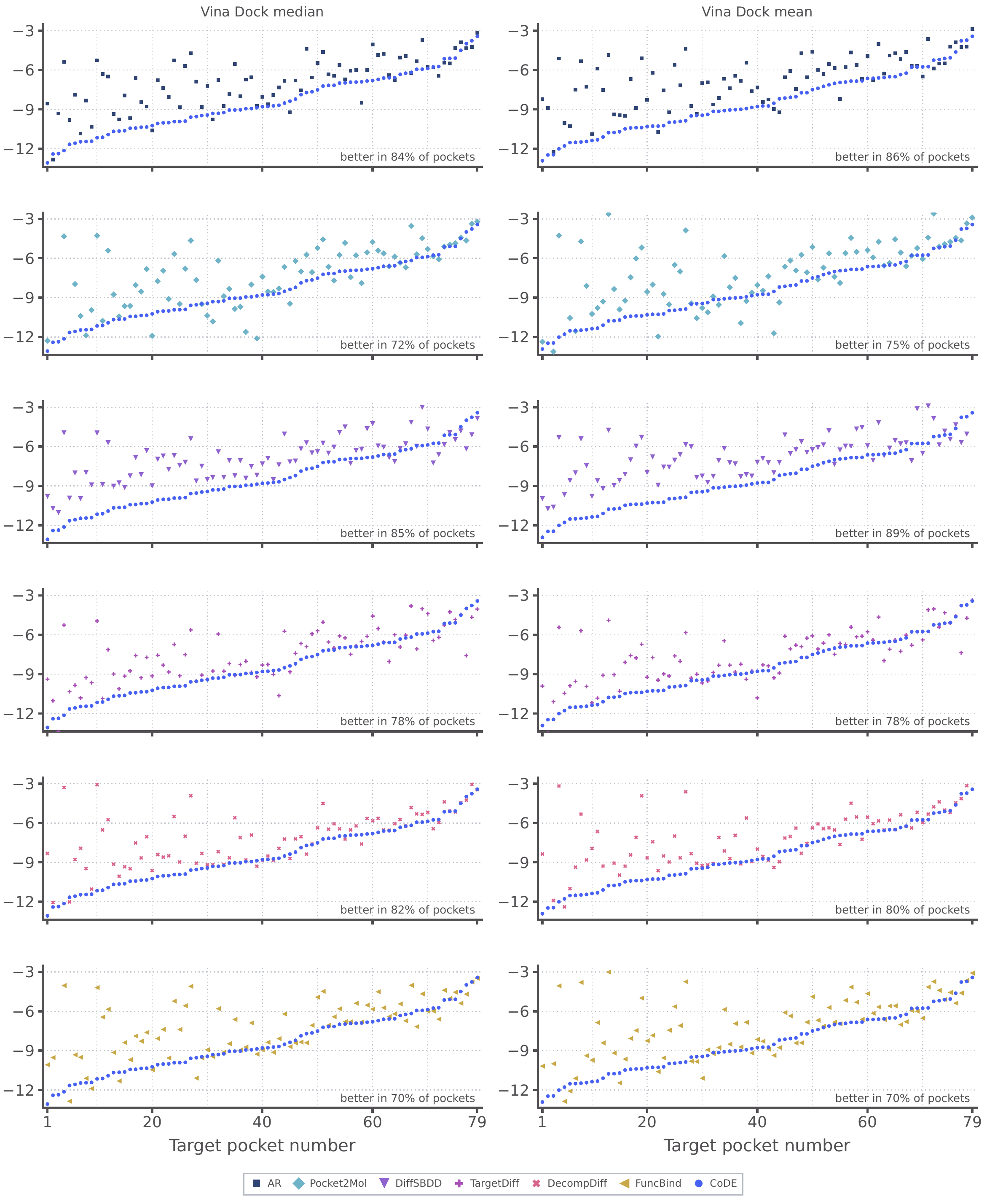}
    \caption{
        \textbf{Mean and median Vina docking scores per pocket compared against each baseline}
        with 79 test pockets having experimental density maps, sorted by the \ours{} Vina docking score.
    }
    \label{fig:result-drug-pocketwise-expanded}
\end{figure}

%% file: ICLR2027/resource/figure/appx/result-drug-qual-all.tex
\begin{figure}[!ht]
    \centering
    \begin{subfigure}[t]{0.49\textwidth}
        \centering
        \includegraphics[width=\linewidth]{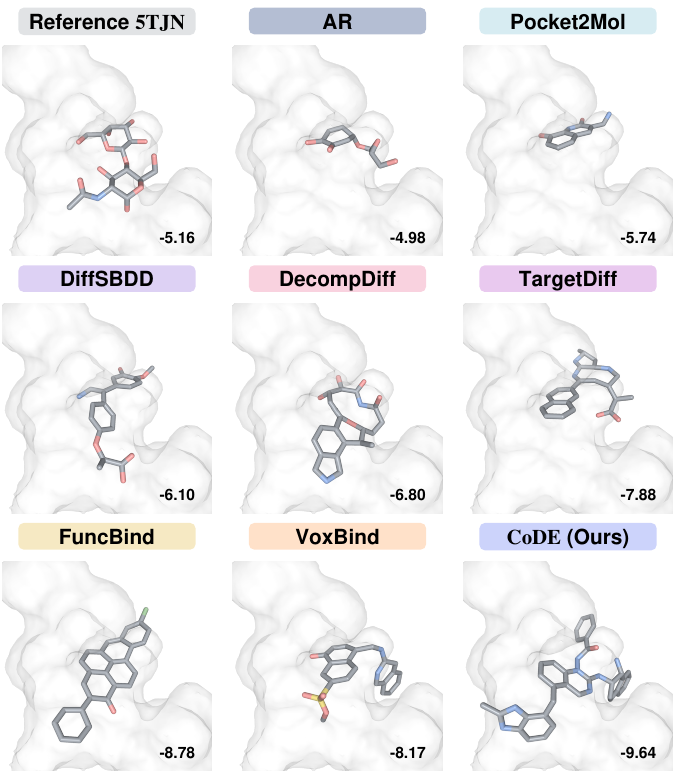}
    \end{subfigure}
    \hfill
    \begin{subfigure}[t]{0.49\textwidth}
        \centering
        \includegraphics[width=\linewidth]{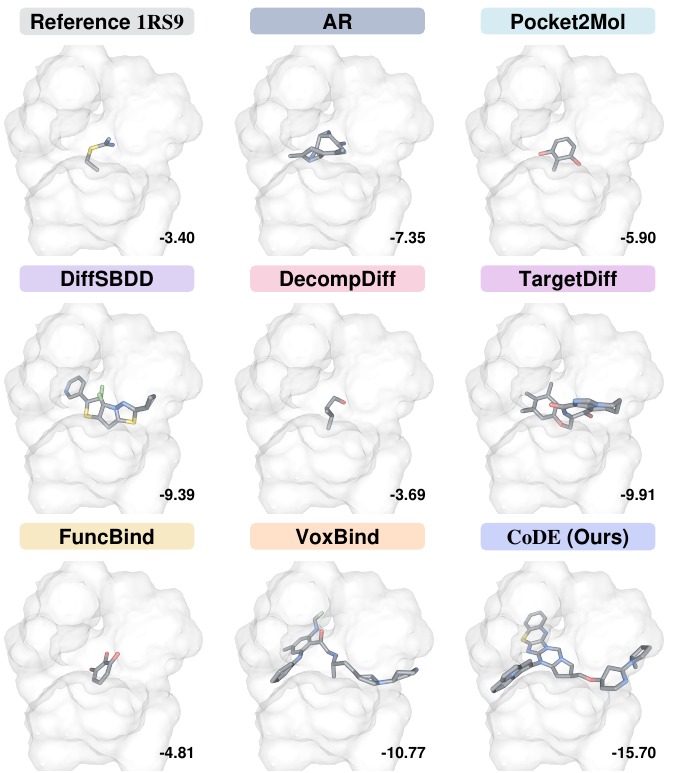}
    \end{subfigure}
    \par\vspace{.2in}
    \begin{subfigure}[t]{0.49\textwidth}
        \centering
        \includegraphics[width=\linewidth]{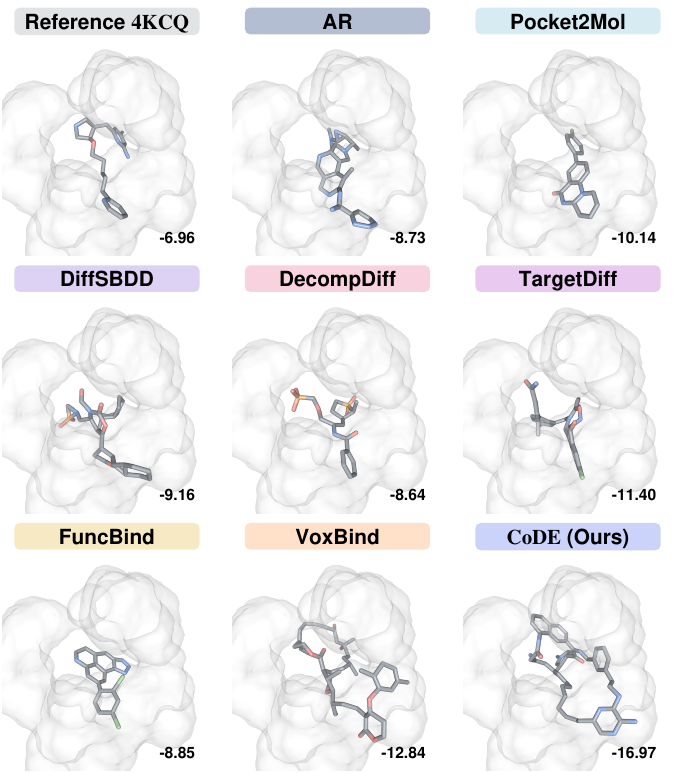}
    \end{subfigure}
    \hfill
    \begin{subfigure}[t]{0.49\textwidth}
        \centering
        \includegraphics[width=\linewidth]{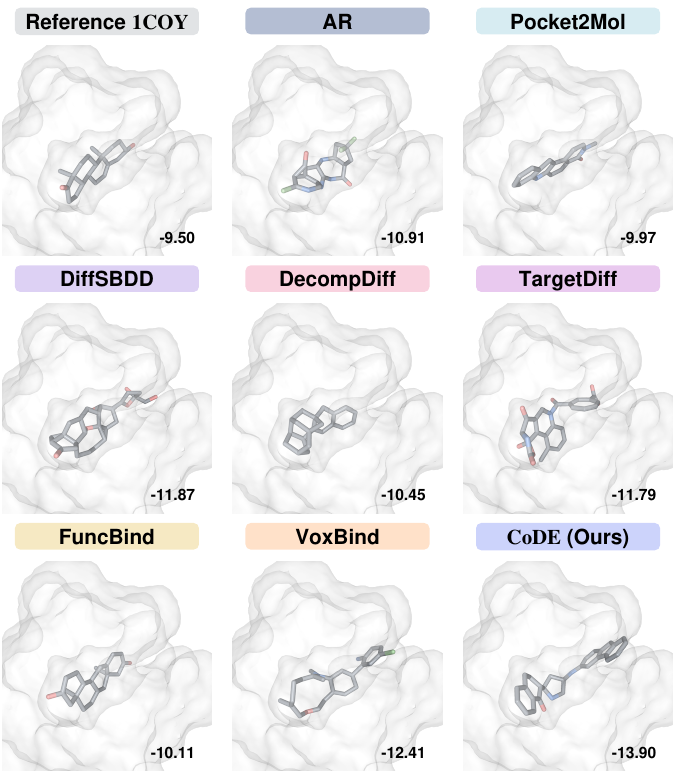}
    \end{subfigure}
    \caption{
        \textbf{Qualitative examples of the generated ligands.} The lowest Vina docking-score ligand generated is shown, with its docking score at the bottom right. Pockets \texttt{5TJN}, \texttt{1RS9}, \texttt{4KCQ}, and \texttt{1COY} are shown from left to right, from top to bottom.
    }
    \label{fig:result-drug-example-qual-all}
\end{figure}

%% file: ICLR2027/resource/figure/appx/result-drug-macrocycle.tex
\begin{figure}[!ht]
    \centering
    \begin{subfigure}[b]{0.45\textwidth}
        \centering
        \includegraphics[width=\linewidth]{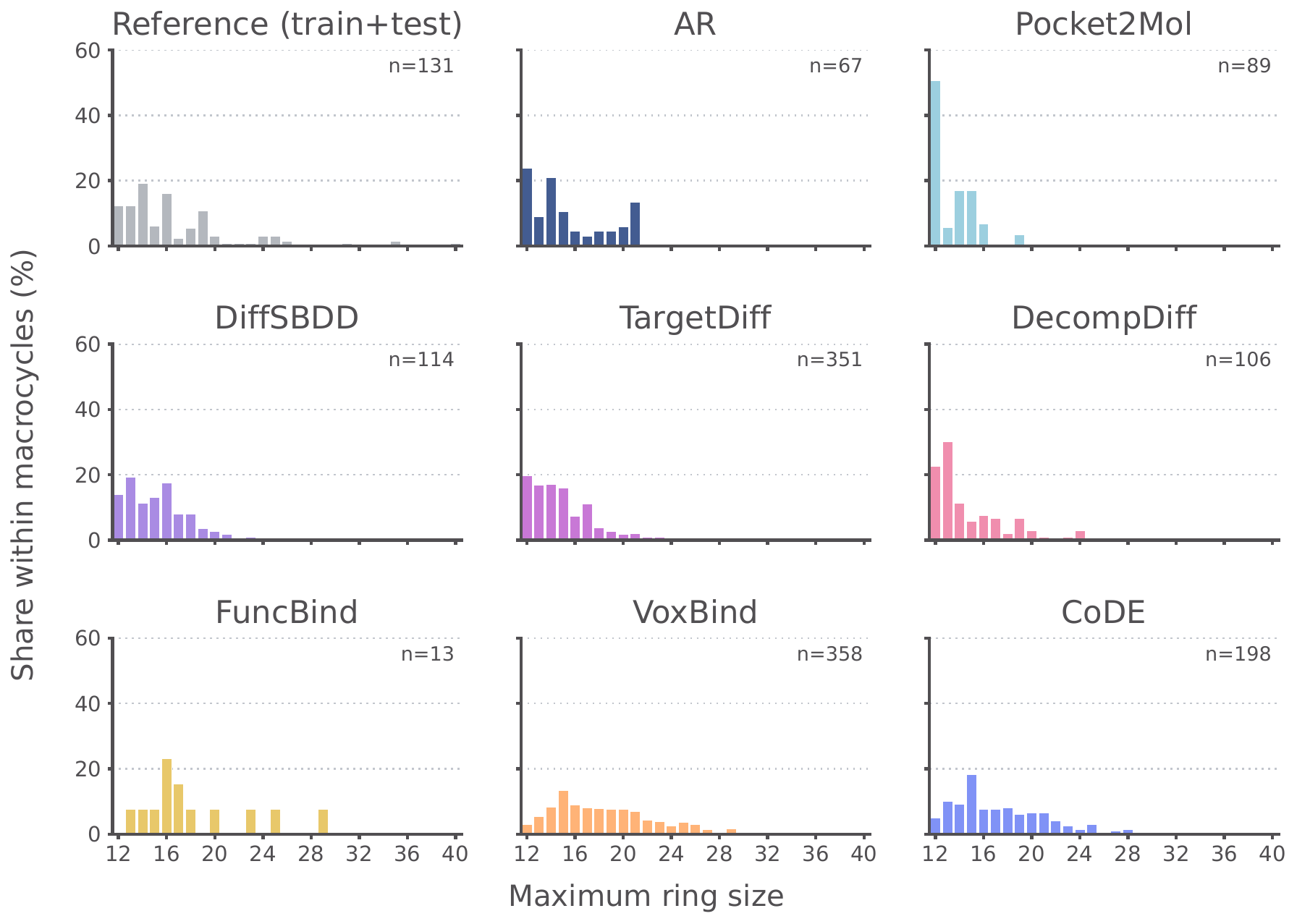}
        \caption{Ring size distribution}
        \label{fig:result-drug-macrocycle-dist}
    \end{subfigure}
    \hfill
    \begin{subfigure}[b]{0.52\textwidth}
        \centering
        \includegraphics[width=\linewidth]{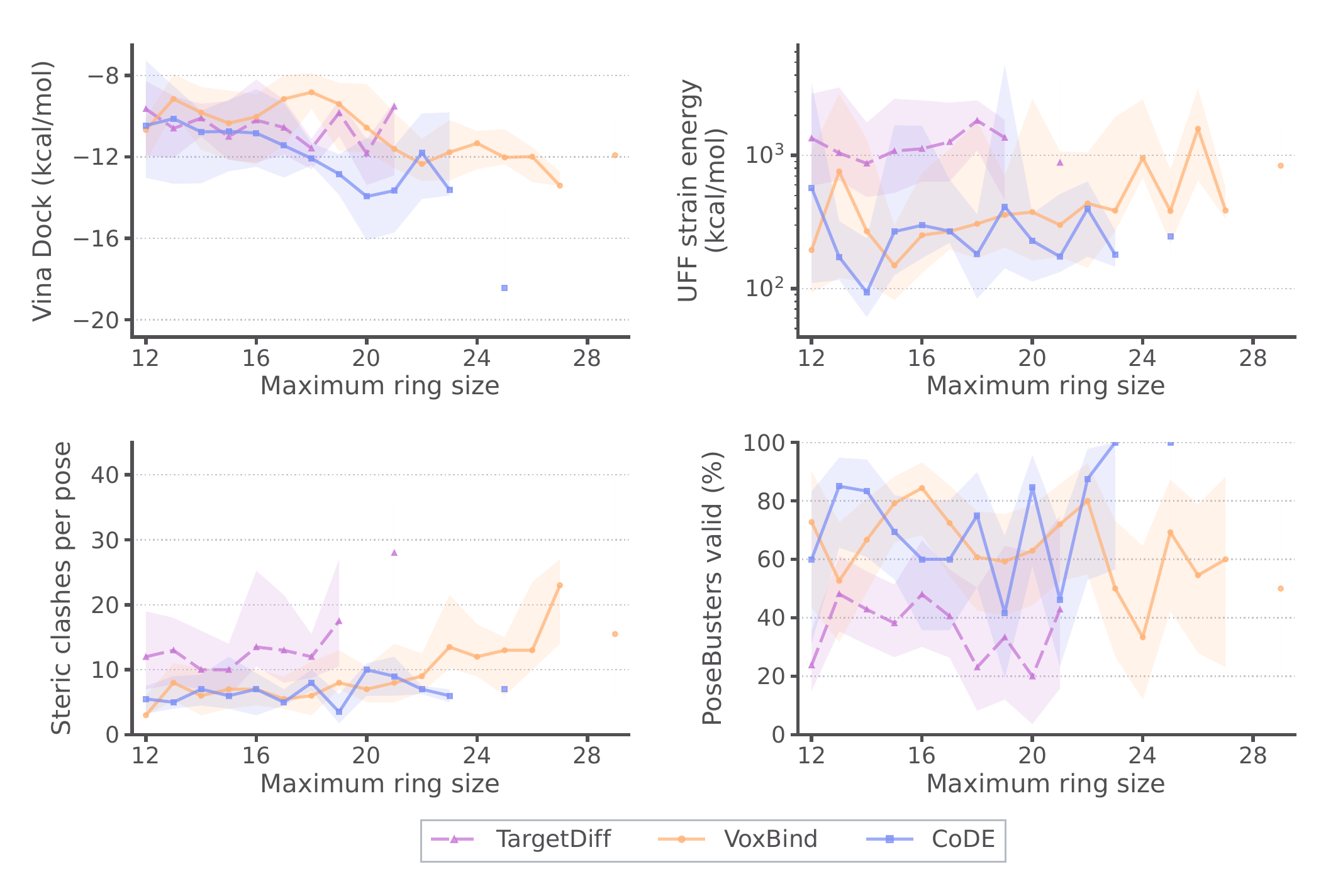}
        \caption{Quality by maximum ring size}
        \label{fig:result-drug-macrocycle-quality-panels}
    \end{subfigure}
    \caption{
        \textbf{Generated macrocycles by maximum ring size.}
        \textbf{\subref{fig:result-drug-macrocycle-dist}} Distribution of the largest SymmSSSR ring size per method.
        \textbf{\subref{fig:result-drug-macrocycle-quality-panels}} Median Vina Dock score, UFF strain energy (logarithmic axis), steric clash count, and PoseBusters pass rate at each exact ring size, without binning or smoothing.
        Shading shows the 25th--75th percentiles and error bars 95\% Wilson intervals.
        Sizes with fewer than five evaluations for a method and metric are omitted, breaking the lines.
    }
    \label{fig:result-drug-macrocycle-quality}
\end{figure}

%% file: ICLR2027/resource/figure/appx/result-drug-posecheck-strain-rot.tex
\begin{figure}[!h]
\centering
    \includegraphics[width=.63\textwidth]{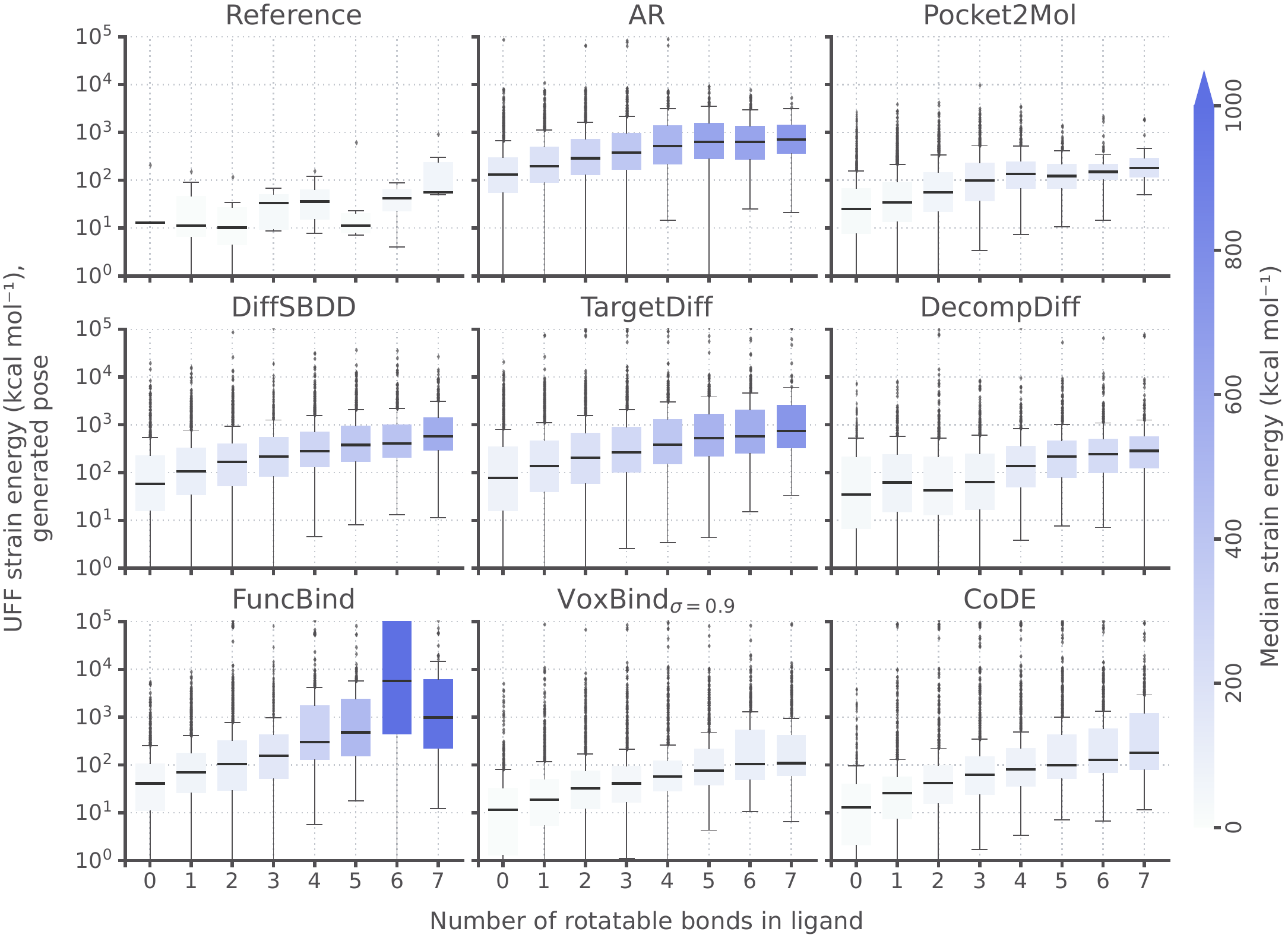}
    \hfill
    \includegraphics[width=.34\textwidth]{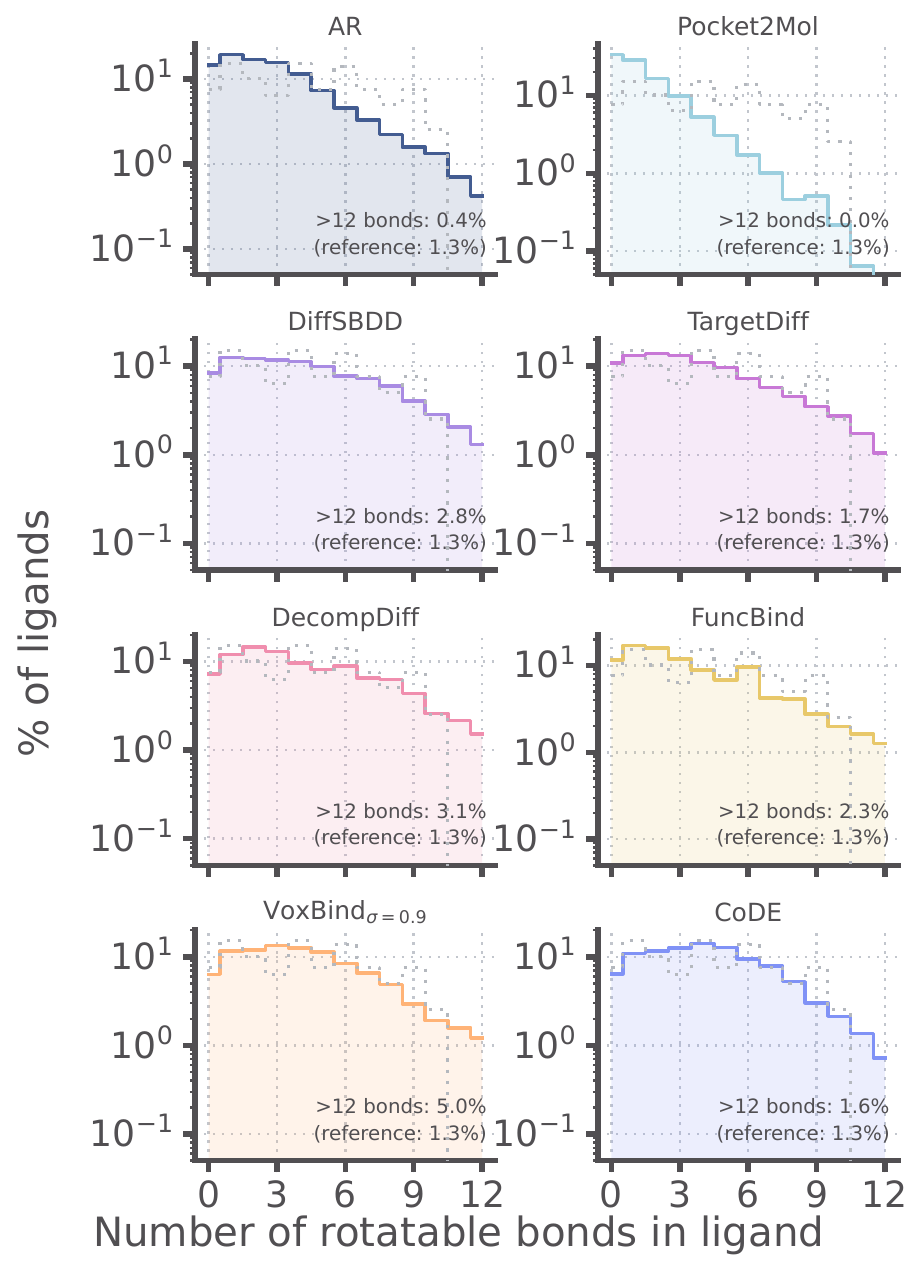}
    \caption{
        \textbf{Strain energy of rotatable bonds.} Strain energy is shown on a log scale, with the median value shown as a horizontal line and the box colored. The box represents the 25th-75th percentile, with outliers additionally plotted.
    }
    \label{fig:result-drug-posecheck-strain-rot}
\end{figure}

%% file: ICLR2027/resource/figure/appx/result-drug-posecheck-interaction.tex
\begin{figure}[!h]
\centering
    \includegraphics[width=.98\textwidth]{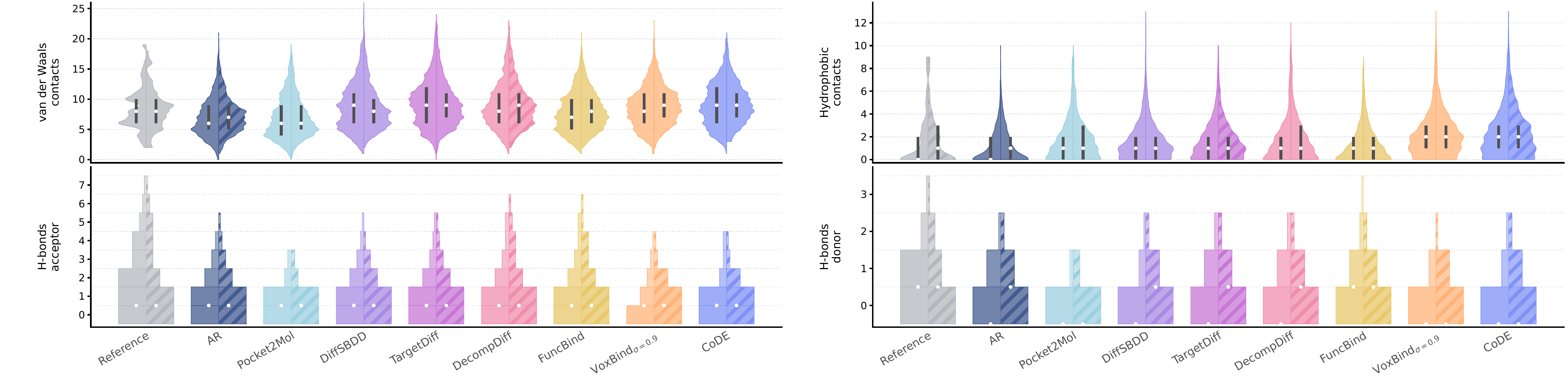}
    \includegraphics[width=.78\textwidth]{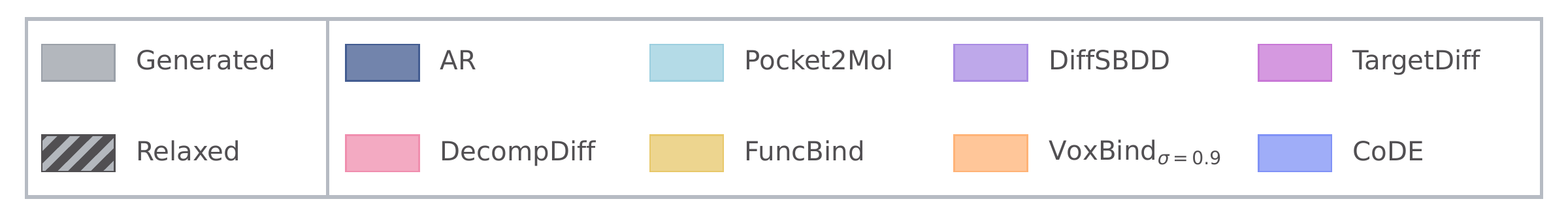}
    \caption{
        \textbf{Protein--ligand interaction.} Left side of each method represents metrics measured from the generated pose, while the right side of each method with slashes is measured from relaxed poses.
    }
    \label{fig:result-drug-posecheck-interaction}
\end{figure}

%% file: ICLR2027/resource/figure/appx/result-pc-redocking.tex
\begin{figure}[!ht]
    \centering
    \begin{subfigure}[t]{0.49\textwidth}
        \centering
        \includegraphics[width=\linewidth]{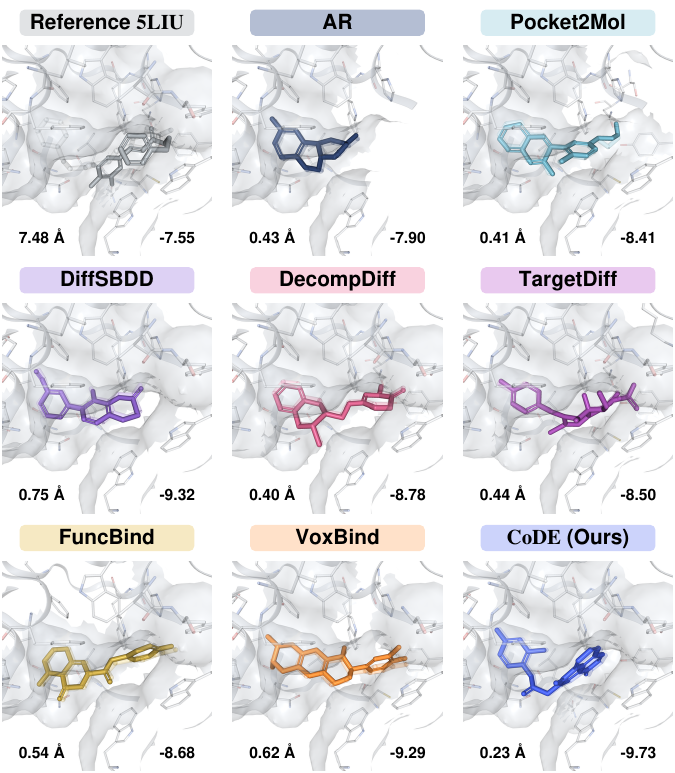}
    \end{subfigure}
    \hfill
    \begin{subfigure}[t]{0.49\textwidth}
        \centering
        \includegraphics[width=\linewidth]{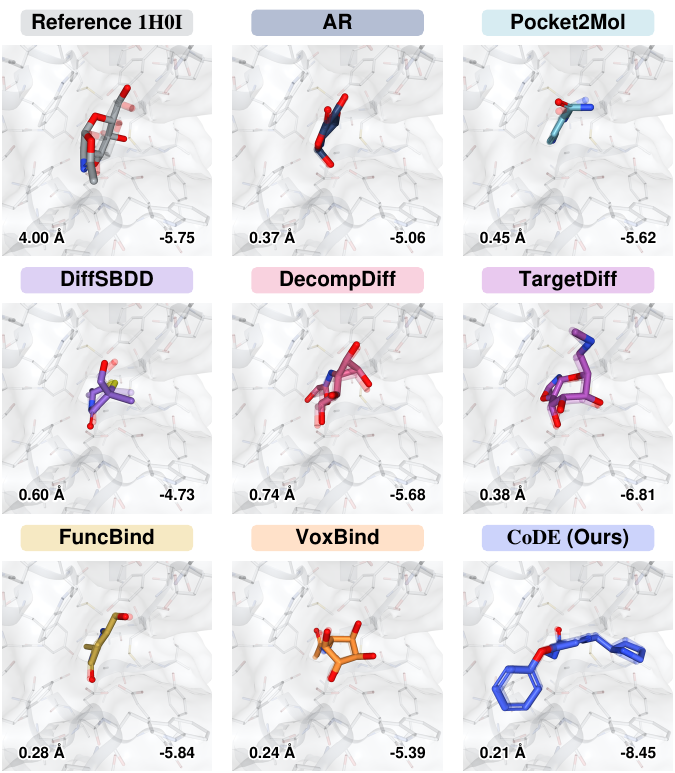}
    \end{subfigure}
    \par\vspace{.1in}
    \begin{subfigure}[t]{0.49\textwidth}
        \centering
        \includegraphics[width=\linewidth]{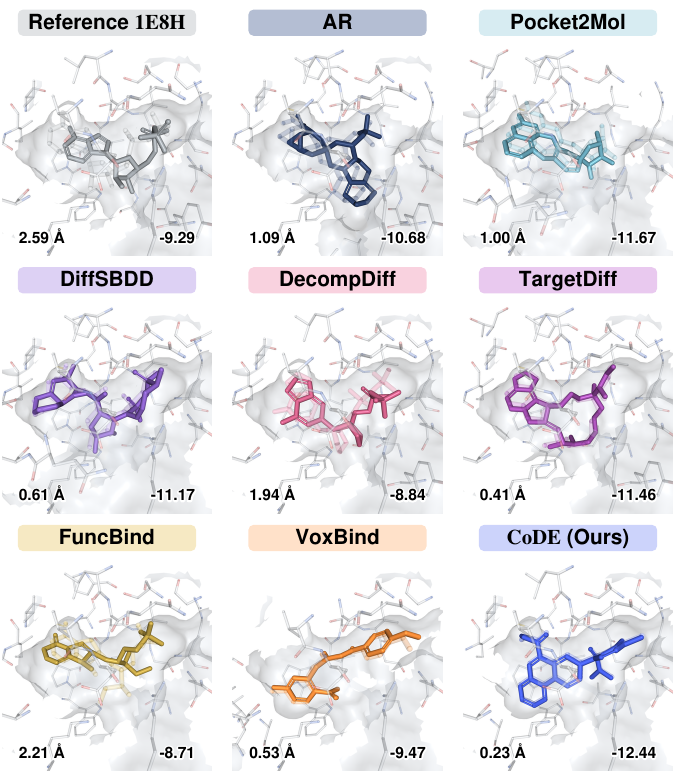}
    \end{subfigure}
    \hfill
    \begin{subfigure}[t]{0.49\textwidth}
        \centering
        \includegraphics[width=\linewidth]{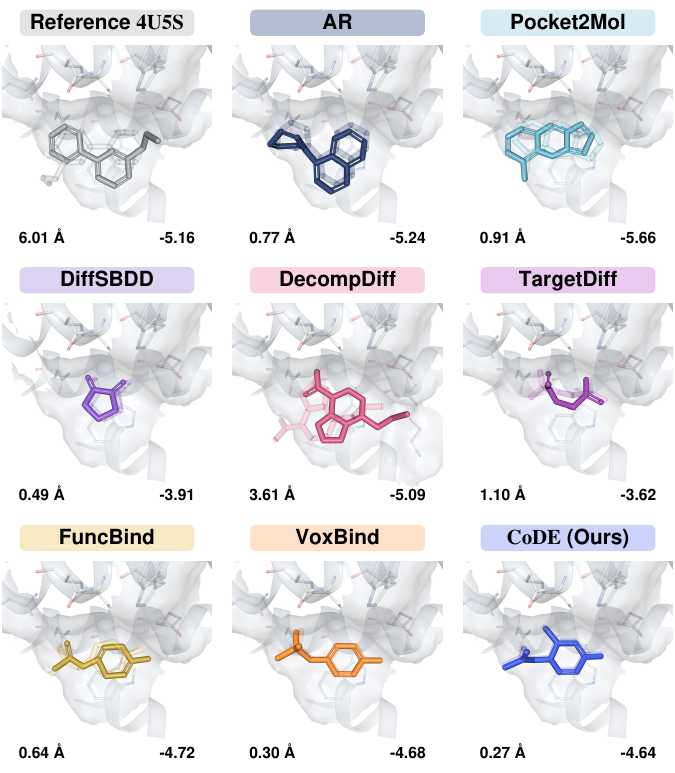}
    \end{subfigure}
    \caption{
        \textbf{Qualitative examples of the re-docked ligands.} Pockets \texttt{5LIU}, \texttt{1H0I}, \texttt{1E8H}, and \texttt{4U5S} are shown from left to right, from top to bottom. Carbon atoms are colored with the method color. Re-docking RMSD and docked Vina score are each visualized in the bottom left and bottom right.
    }
    \label{fig:result-pc-redock-all}
\end{figure}

%% file: ICLR2027/resource/figure/appx/result-jsd-atom-type.tex
\begin{figure}[!ht]
\centering
    \includegraphics[width=.98\textwidth]{ICLR2027/resource/image/jsd-legend.pdf}
    \includegraphics[width=.98\textwidth]{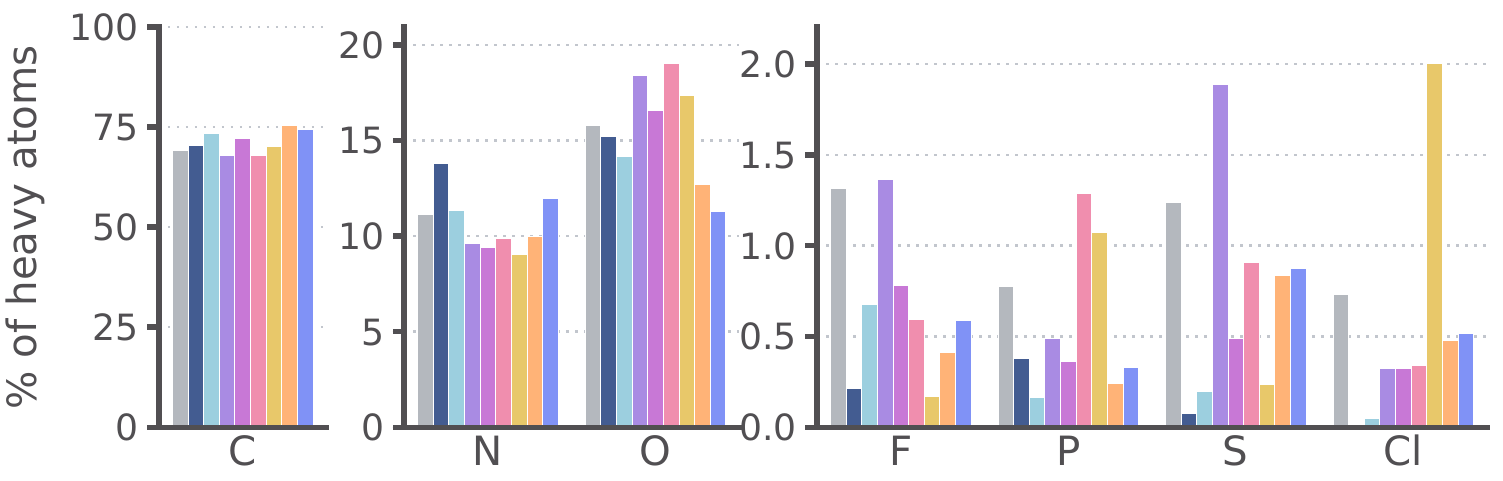}
    \caption{
        \textbf{Heavy-atom element shares.}
        Percentage of a set's heavy atoms at each element, split across three linearly scaled panels so that rare elements are not flattened against carbon: C, then N and O, then F, P, S, and Cl.
    }
    \label{fig:result-drug-jsd-atom-type}
\end{figure}

%% file: ICLR2027/resource/figure/appx/result-jsd-pair.tex
\begin{figure}[!ht]
\centering
    \includegraphics[width=.98\textwidth]{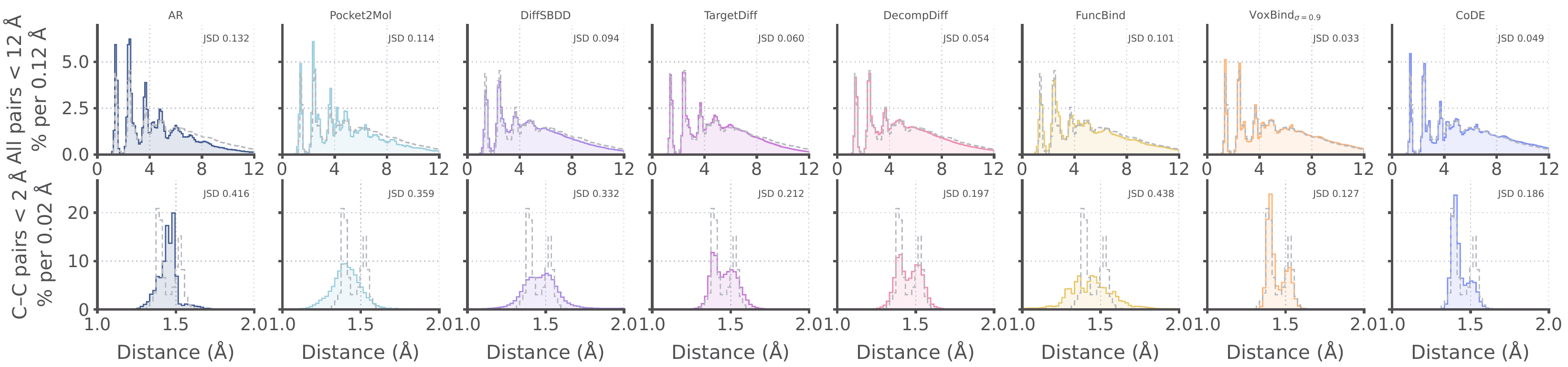}
    \caption{
        \textbf{Interatomic distance distributions.} All heavy-atom pairs within a molecule closer than 12\AA{} is shown above, and carbon--carbon pairs closer than 2\AA{} below. Distances are computed from coordinates and elements alone, without bond perception. The dashed gray outline is the reference ligand, the number in the corner shows the JSD, and the vertical scale is shared.
    }
    \label{fig:result-drug-jsd-pair}
\end{figure}

%% file: ICLR2027/resource/figure/appx/result-jsd-bond-distance.tex
\begin{figure}[!ht]
\centering
    \includegraphics[width=.98\textwidth]{ICLR2027/resource/image/jsd-legend.pdf}
    \includegraphics[width=.98\textwidth]{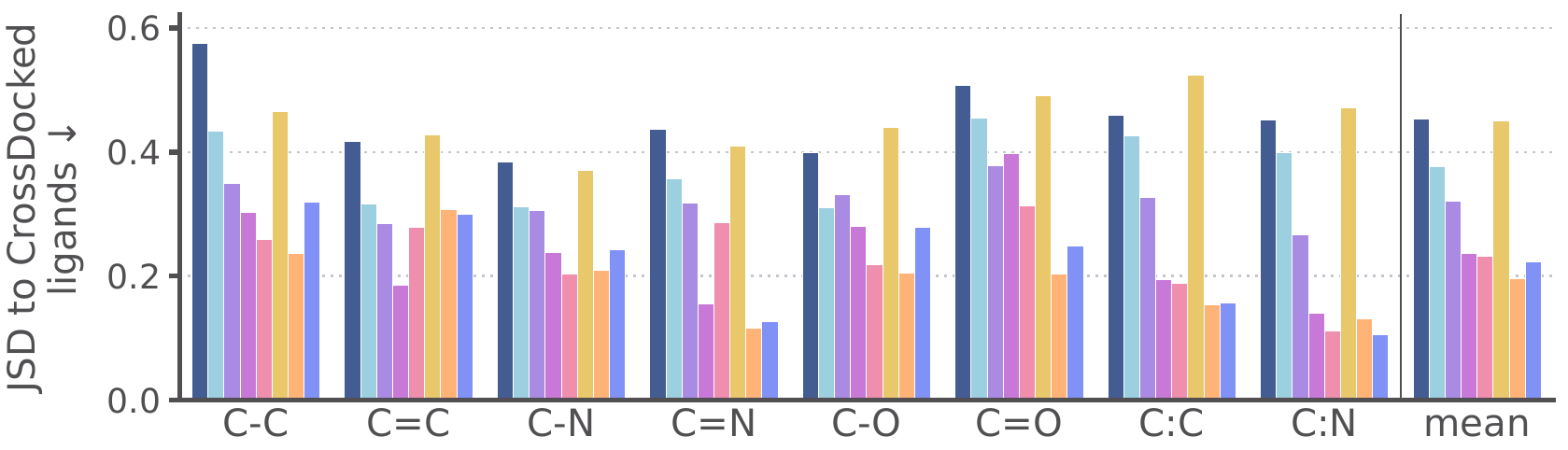}
    \includegraphics[width=.98\textwidth]{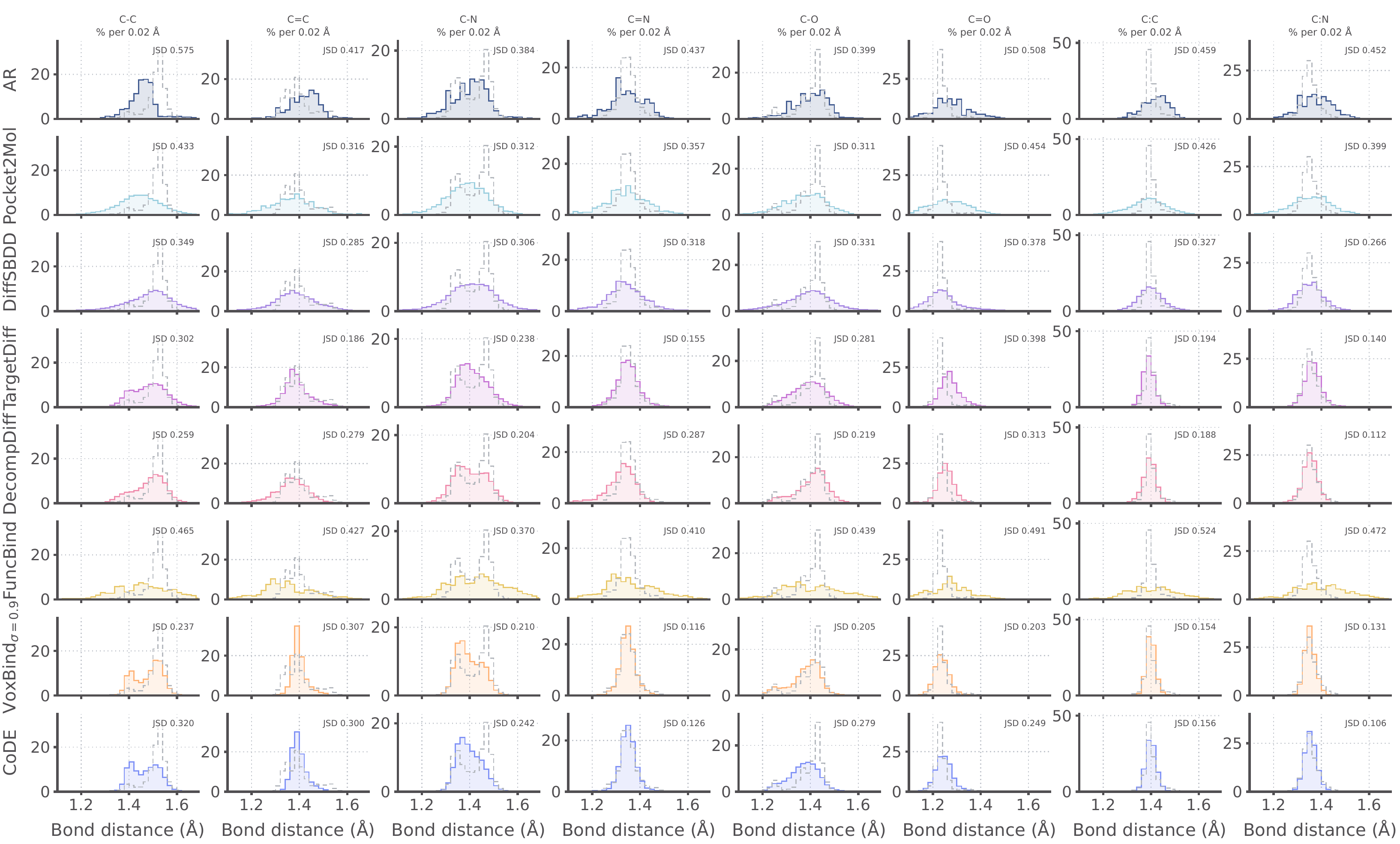}
    \caption{
        \textbf{Bond-length Jensen–Shannon distance per bond type (top) and detailed distribution (bottom)}, for the eight bond types, the lower the value, the closer to the reference distribution. The dashed gray outline is the reference distribution, and the number in the corner is that panel's JSD. Distances are scored in 0.005\AA{} bins and drawn four bins at a time, and the vertical scale is shared.
    }
    \label{fig:result-drug-jsd-bond-distance}
\end{figure}

%% file: ICLR2027/resource/figure/appx/result-jsd-rings.tex
\begin{figure}[!ht]
\centering
    \includegraphics[width=.98\textwidth]{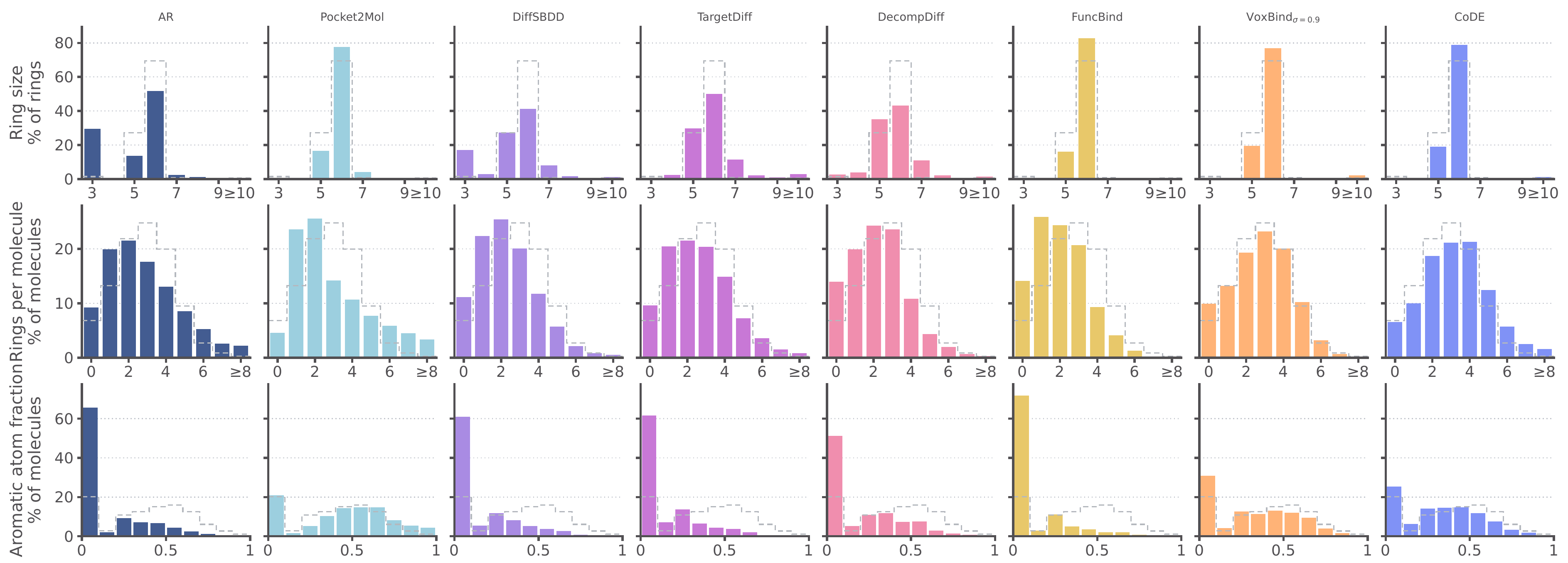}
    \caption{
        \textbf{Ring statistics, one column per method.}
        Ring sizes, rings per molecule, and the aromatic fraction of a molecule's atoms and of its rings, with the reference ligands repeated in gray.
    }
    \label{fig:result-drug-jsd-rings}
\end{figure}

%% file: ICLR2027/resource/figure/appx/result-jsd-ring-size.tex
\begin{figure}[!ht]
\centering
    \includegraphics[width=.88\textwidth]{ICLR2027/resource/image/jsd-legend.pdf}
    \includegraphics[width=.88\textwidth]{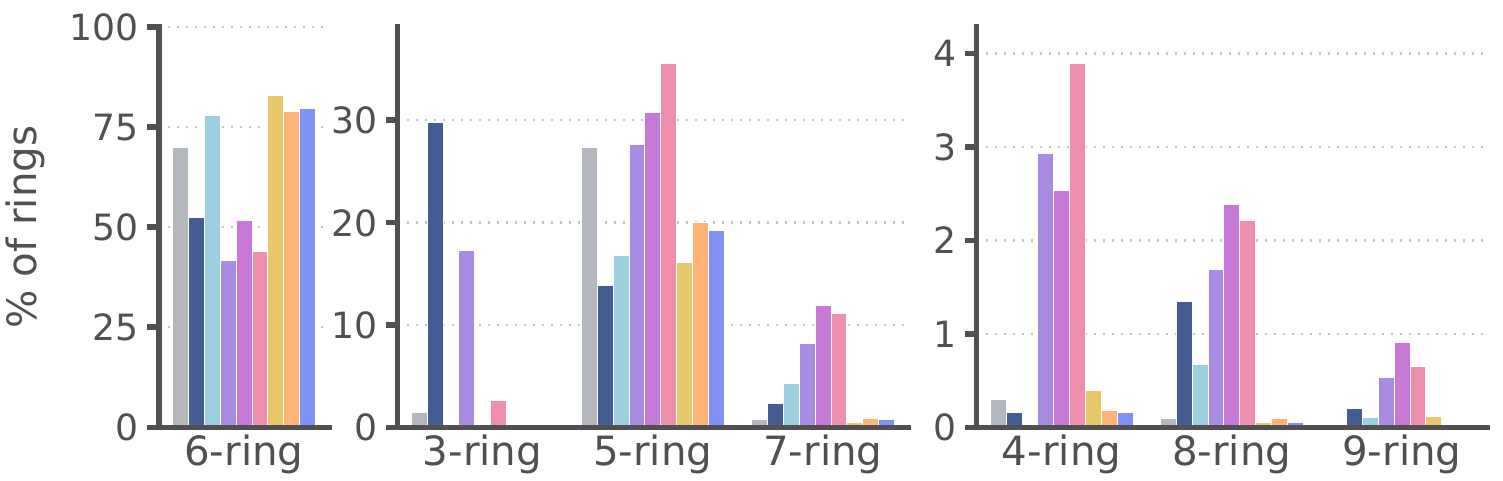}
    \caption{
        \textbf{Ring-size distribution.}
        The three panels are scaled independently: six-membered rings, then three-, five-, and seven-membered rings, then four-, eight-, and nine-membered rings.
    }
    \label{fig:result-drug-jsd-ring-size}
\end{figure}

%% file: ICLR2027/resource/figure/appx/result-jsd-n-rings.tex
\begin{figure}[!ht]
\centering
    \includegraphics[width=.88\textwidth]{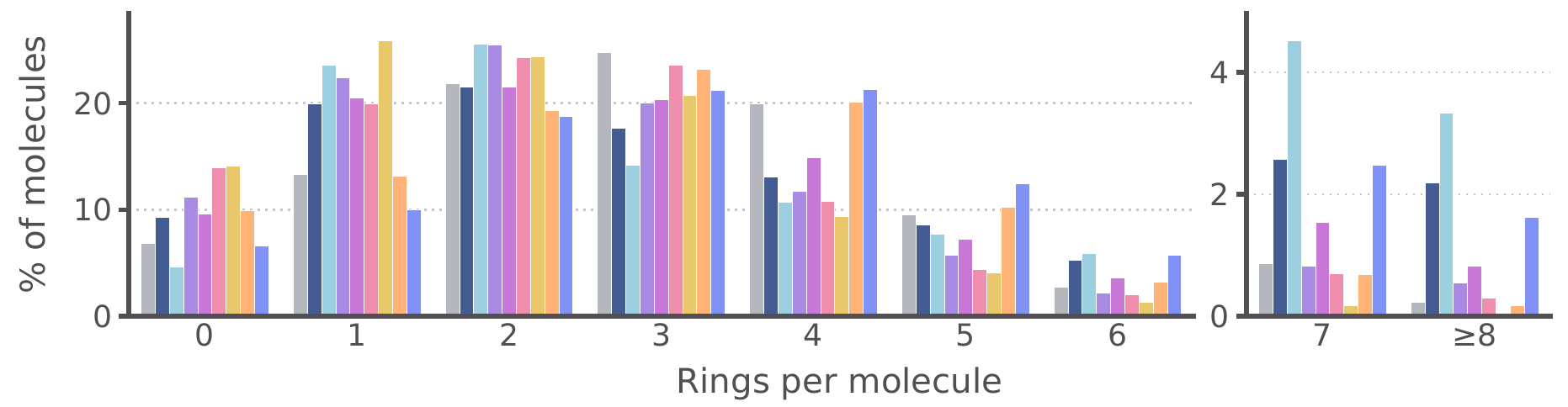}
    \caption{
        \textbf{Rings per molecule.}
        Percentage of a set's molecules containing each number of rings, with the final bar pooling molecules with eight or more rings.
    }
    \label{fig:result-drug-jsd-n-rings}
\end{figure}

%% file: ICLR2027/resource/figure/appx/result-jsd-aromatic.tex
\begin{figure}[!ht]
\centering
    \includegraphics[width=.98\textwidth]{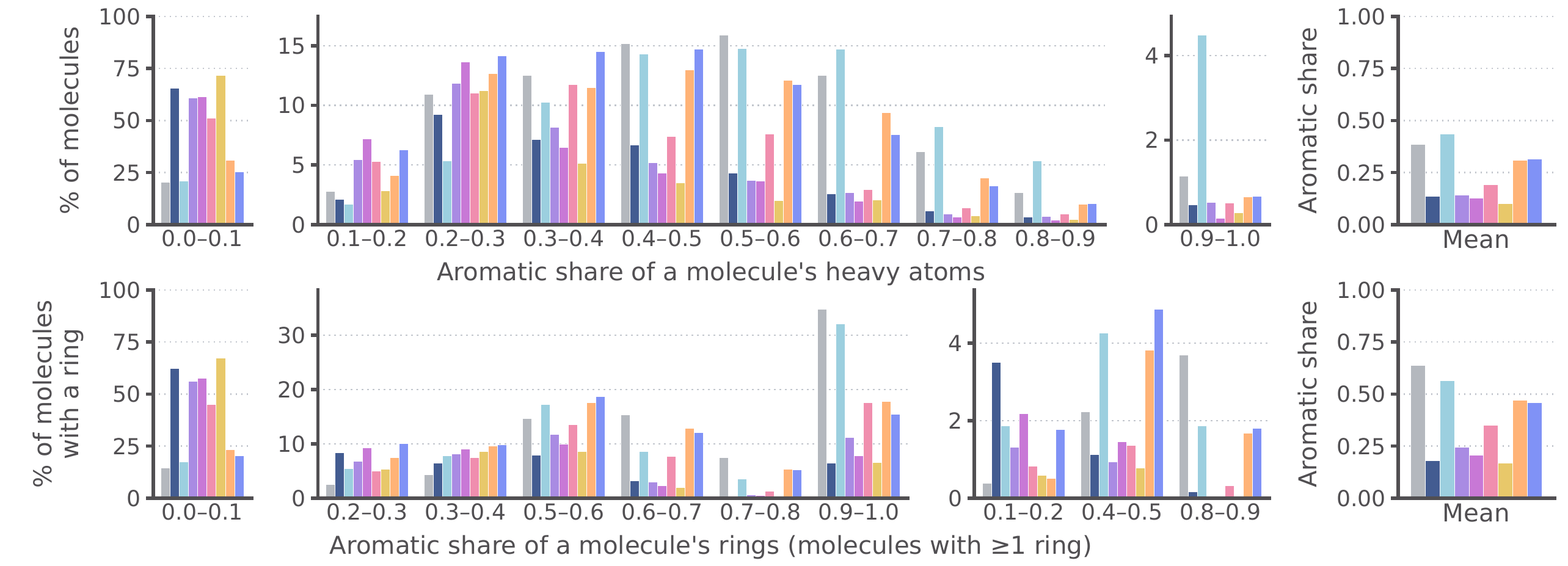}
    \caption{
        \textbf{Aromatic content.}
        Top: the fraction of a molecule's heavy atoms that are aromatic. Bottom: the fraction of a molecule's aromatic rings, meaning every bond of the ring is aromatic, computed over molecules with at least one ring. Both panels show the percentage of a set's molecules in each fraction bin, with the reference ligands drawn alongside the methods.
    }
    \label{fig:result-drug-jsd-aromatic}
\end{figure}

%% file: ICLR2027/resource/figure/appx/result-rigid-expanded.tex
\begin{figure}[!ht]
\centering
    \includegraphics[width=.98\textwidth]{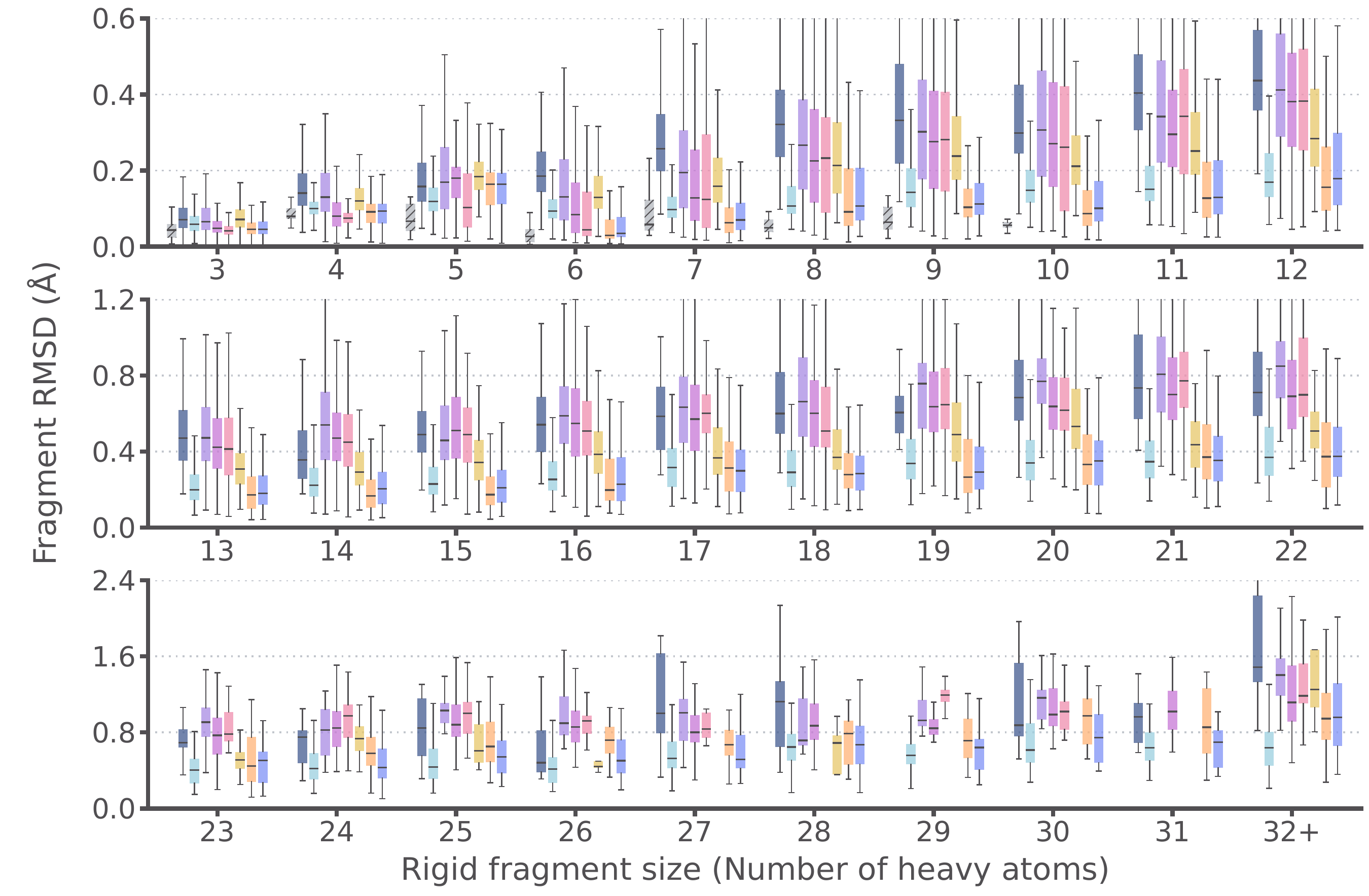}
    \caption{
        \textbf{Rigid fragment consistency expanded to more number of fragment size.}
    }
    \label{fig:result-rigid-all}
\end{figure}

%% file: ICLR2027/resource/figure/appx/result-rigid-distribution.tex
\begin{figure}[!ht]
\centering
    \includegraphics[width=.98\textwidth]{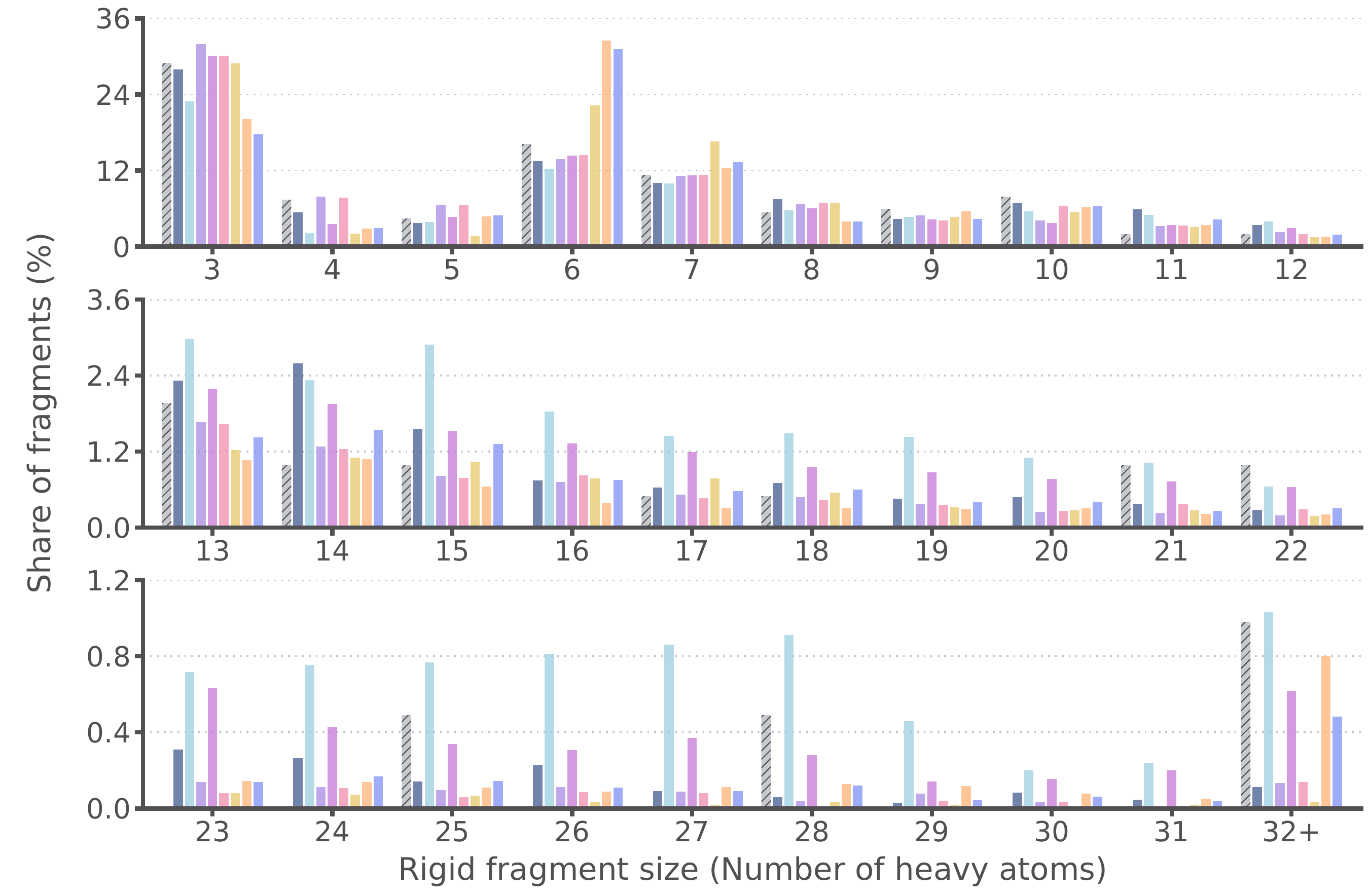}
    \caption{
        \textbf{Rigid fragment size distribution from generated molecules.}
    }
    \label{fig:result-rigid-dis}
\end{figure}

%% file: ICLR2027/resource/figure/appx/result-pb-all.tex
\begin{figure}[!ht]
\centering
    \includegraphics[width=.98\textwidth]{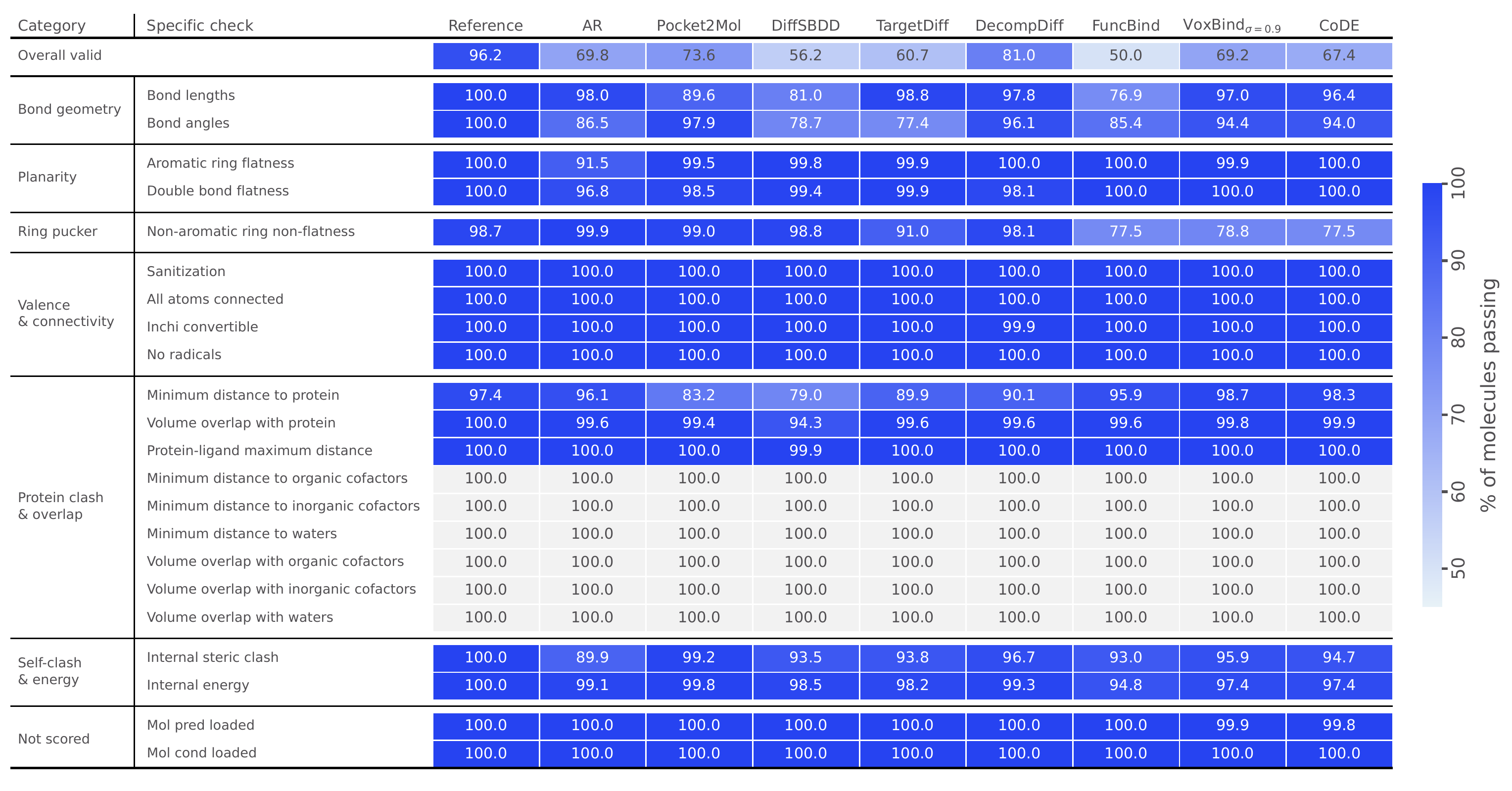}
    \caption{
        \textbf{PoseBusters validity per category.}
    }
    \label{fig:result-pb-expanded}
\end{figure}

%% file: ICLR2027/resource/figure/appx/result-pb-upset.tex
\begin{figure}[!ht]
\centering
    \includegraphics[width=.98\textwidth]{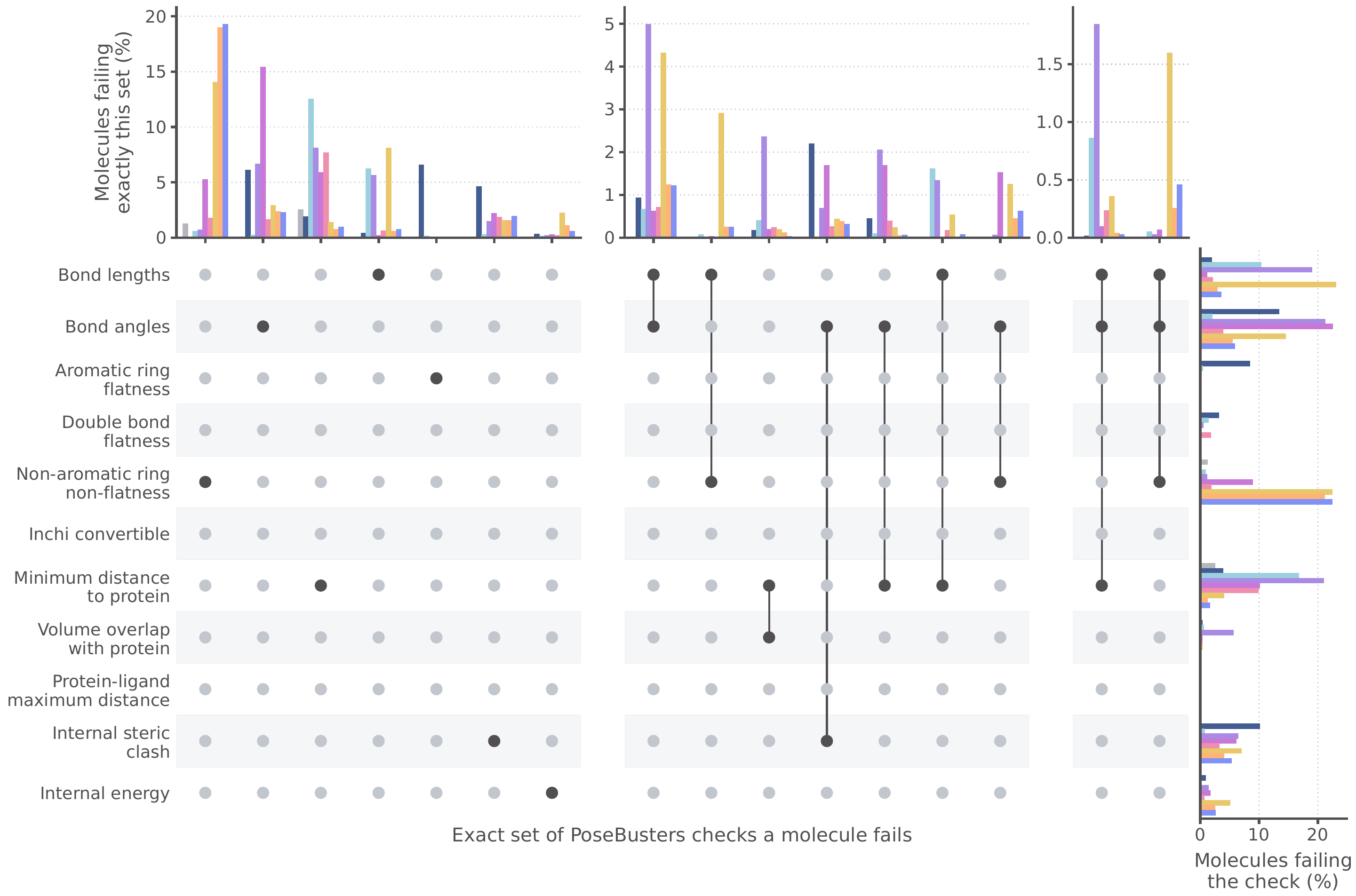}
    \includegraphics[width=.98\textwidth]{ICLR2027/resource/image/jsd-legend.pdf}
    \caption{
        \textbf{PoseBusters validity upset plot.} Relation between PoseBusters categories.
    }
    \label{fig:result-pb-all}
\end{figure}

%% file: ICLR2027/resource/figure/appx/result-pb-analysis.tex
\begin{figure}[!ht]
\centering
    \includegraphics[width=\textwidth]{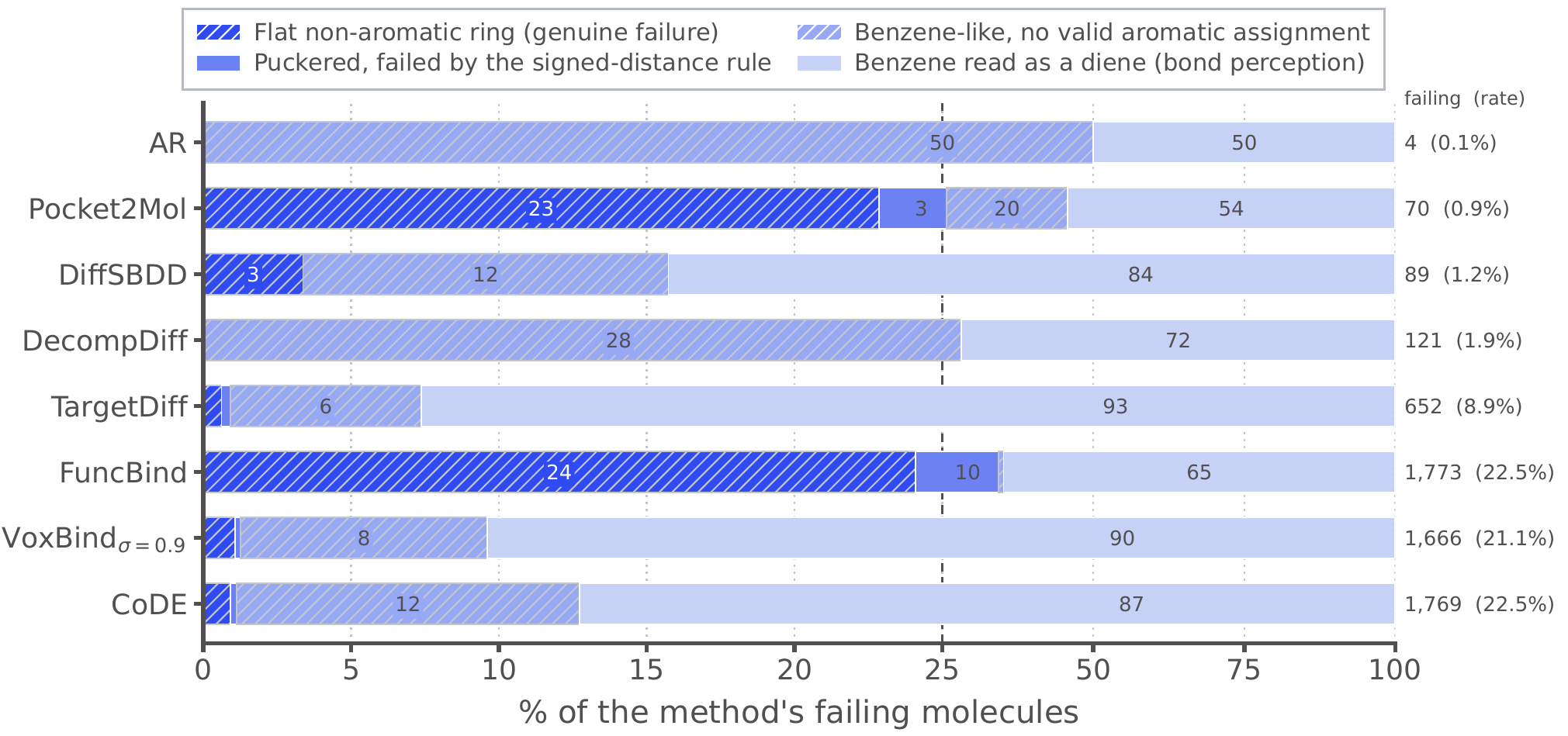}
    \caption{
        \textbf{Diagnostic categories of PoseBusters ring non-flatness failures.}
        Each bar splits the molecules of one method that fail \texttt{non-aromatic\_ring\_non-flatness} into the four categories, as a percentage of its failing molecules; segments narrower than about $5\%$ of the axis are left unlabeled. The horizontal axis is stretched below $25\%$ and compressed above it (dashed line), each part linear, so that the small causes remain legible. On the right are the number of failing molecules and the failure rate over all generated molecules; the AR bar rests on only four molecules.
    }
    \label{fig:result-pb-ring-nonflat-causes}
\end{figure}

%% file: ICLR2027/resource/table/appx/appx-pb-change.tex
\begin{table}[!ht]
    \centering
    \caption{
        \textbf{Evaluation metrics before and after alternative aromatic reassignment.}
        A failing six-membered ring with benzene-like geometry is reassigned as aromatic by modifying the ring bond orders and the attached hydrogen counts, while preserving the heavy-atom connectivity and coordinates. Rings bearing an exocyclic double bond are left unchanged in the procedure.
        \textit{Ring fail} is the share of molecules failing \texttt{non-aromatic\_ring\_non-flatness}, re-derived ring by ring (rings whose outcome depends on the sign of the plane normal count as failing). \textit{All rings} is the rate if rings with an exocyclic double bond are re-assigned as well. QED, SA, and logP are means over all molecules; diversity is averaged over pockets.
    }
    \label{tab:result-pb-rearomatize-change}
    \setlength{\tabcolsep}{4pt}
    \resizebox{.98\textwidth}{!}{%
        \begin{tabular}{lcccccccc}
            \toprule
            & \multicolumn{3}{c}{\textbf{Ring fail} (\%)} & \textbf{Changed} & & & & \\
            \cmidrule(lr){2-4}
            \textbf{Method} & Before & After & All rings & (\%) & \textbf{QED} & \textbf{SA} & \textbf{logP} & \textbf{Diversity} \\
            \midrule
            AR & 0.1 & 0.1 & 0.0 & 0.0 & 0.504 $\to$ 0.504 & 0.629 $\to$ 0.629 & 0.37 $\to$ 0.37 & 0.709 $\to$ 0.709 \\
            Pocket2Mol & 1.1 & 1.0 & 0.5 & 0.1 & 0.566 $\to$ 0.566 & 0.743 $\to$ 0.743 & 1.64 $\to$ 1.64 & 0.724 $\to$ 0.724 \\
            DiffSBDD & 1.4 & 1.4 & 0.2 & 0.0 & 0.495 $\to$ 0.495 & 0.628 $\to$ 0.628 & 1.07 $\to$ 1.07 & 0.796 $\to$ 0.796 \\
            DecompDiff & 1.9 & 1.8 & 0.5 & 0.2 & 0.494 $\to$ 0.494 & 0.666 $\to$ 0.666 & 1.12 $\to$ 1.12 & 0.711 $\to$ 0.711 \\
            TargetDiff & 9.5 & 3.7 & 0.7 & 5.9 & 0.493 $\to$ 0.497 & 0.608 $\to$ 0.615 & 1.36 $\to$ 1.36 & 0.747 $\to$ 0.748 \\
            FuncBind & 27.0 & 12.3 & 9.2 & 16.6 & 0.503 $\to$ 0.511 & 0.656 $\to$ 0.678 & 1.35 $\to$ 1.37 & 0.702 $\to$ 0.712 \\
            VoxBind\textsubscript{\scriptsize $\sigma$=0.9} & 21.3 & 8.7 & 2.1 & 14.0 & 0.540 $\to$ 0.544 & 0.704 $\to$ 0.720 & 2.65 $\to$ 2.66 & 0.763 $\to$ 0.764 \\
            \midrule
            VoxBind\textsubscript{\scriptsize $\sigma$=0.9}\,+\,\ours{} & 23.0 & 10.6 & 2.9 & 13.9 & 0.523 $\to$ 0.526 & 0.678 $\to$ 0.692 & 2.62 $\to$ 2.63 & 0.710 $\to$ 0.711 \\
            \bottomrule
        \end{tabular}
    }
\end{table}

\begin{table}[!ht]
    \centering
    \caption{
        \textbf{Vina before and after the re-assignment}: We randomly selected 50 molecules for each of two reassignment procedures: one that preserves exocyclic bond orders and one that also modifies them. Each molecule was docked before and after reassignment.$\Delta$ is after minus before in kcal/mol, mean (median); positive is worse. \textit{Same pose}: the two docked poses lie within $0.5$\,\AA{} RMSD. \textit{Gen.$\to$dock}: median RMSD between the generated and the docked pose, before $\to$ after.
    }
    \label{tab:result-pb-rearomatize-vina}
    \setlength{\tabcolsep}{4pt}
    \resizebox{.98\textwidth}{!}{%
        \begin{tabular}{lrccccc}
            \toprule
            \textbf{Fix} & $n$ & $\Delta$\textbf{Score} & $\Delta$\textbf{Min} & $\Delta$\textbf{Dock} & \textbf{Same pose} & \textbf{Gen.$\to$dock} (\AA{}) \\
            \midrule
            Ring bond-order / H assignment (used) & 49 & $+0.008$ ($+0.000$) & $-0.068$ ($+0.000$) & $-0.007$ ($+0.000$) & 40/49 & 3.37 $\to$ 3.41 \\
            Including exocyclic bond-order changes & 49 & $+0.187$ ($+0.241$) & $+0.242$ ($+0.252$) & $+0.188$ ($+0.231$) & 10/49 & 2.27 $\to$ 2.42 \\
            \bottomrule
        \end{tabular}
    }
\end{table}

%% file: ICLR2027/resource/figure/appx/result-pb-example.tex
\begin{figure}[!ht]
\centering
    \includegraphics[width=0.9\textwidth]{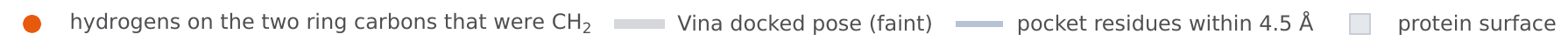}
    \begin{subfigure}[t]{0.49\textwidth}
        \centering
        \includegraphics[width=\linewidth]{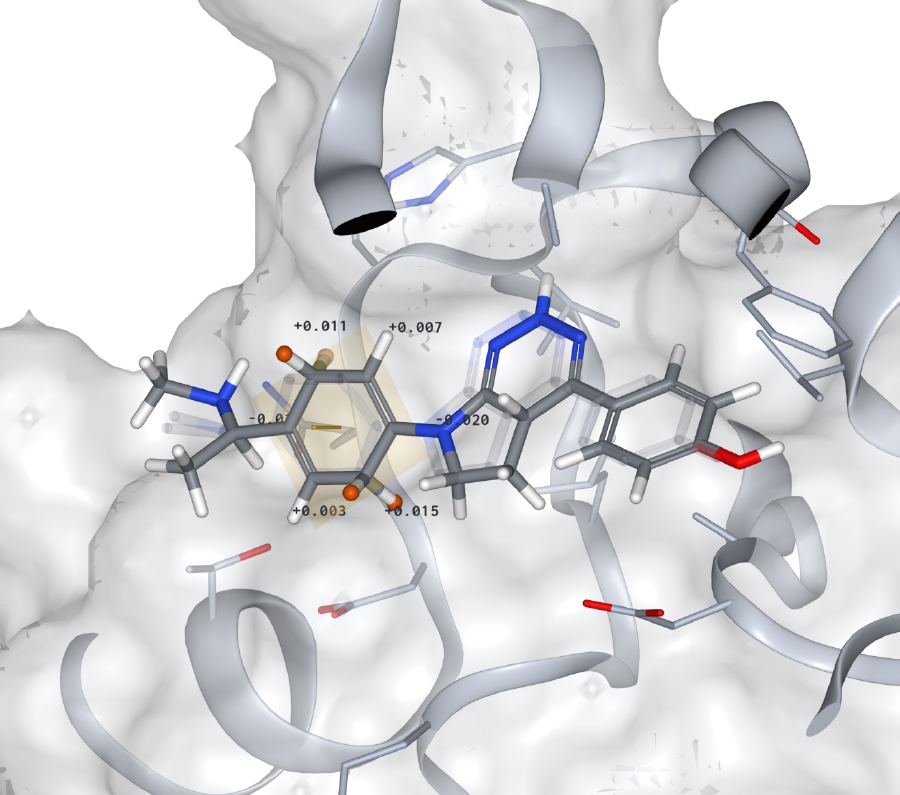}
        \caption{
            \textbf{Original reconstruction: 1,4-cyclohexadiene.} Two ring carbons are CH$_2$; the ring's largest distance to its plane is $0.020$\,\AA{}, so PoseBusters fails it as non-aromatic and flat. Vina score $-8.54$, dock $-8.73$\,kcal/mol.
        }
        \label{fig:qual-diene}
    \end{subfigure}
    \hfill
    \begin{subfigure}[t]{0.49\textwidth}
        \centering
        \includegraphics[width=\linewidth]{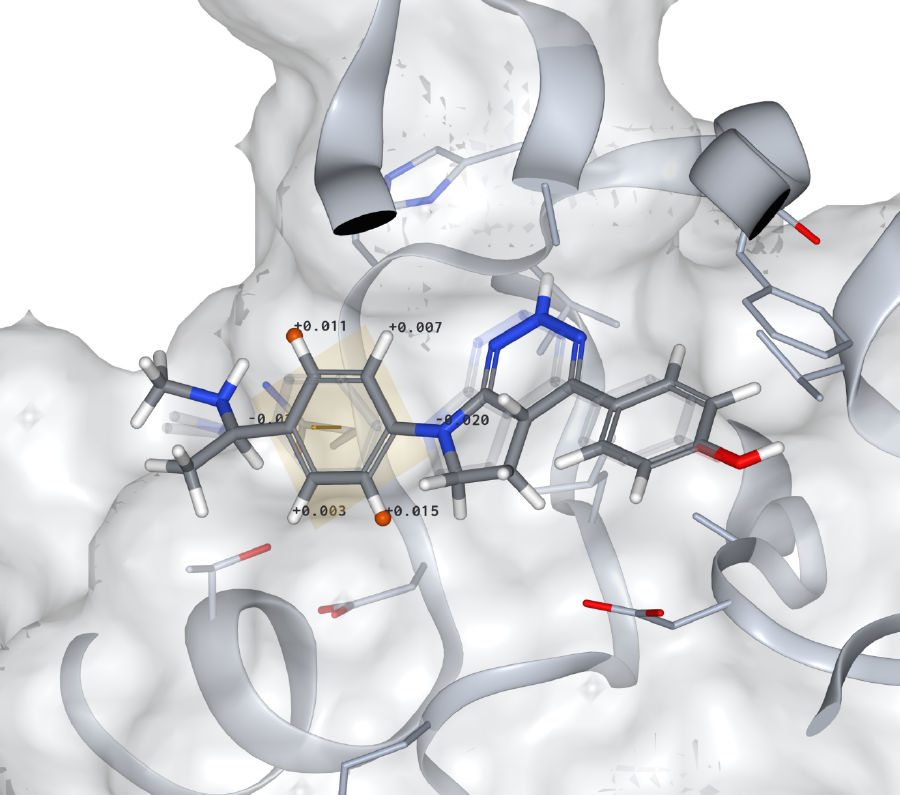}
        \caption{
            \textbf{Alternative aromatic reconstruction: benzene.} Same heavy atoms at the same coordinates; the ring is aromatic and not tested. Vina score $-8.54$, dock $-8.73$\,kcal/mol.
        }
        \label{fig:qual-benzene}
    \end{subfigure}
    \caption{
        \textbf{Alternative molecular reconstruction of the same generated heavy atom geometry.}
        A molecule generated by \ours{} for \texttt{CDK6}. The yellow square is the best-fit plane of the failing ring, with each ring atom's signed distance to it in \AA{}; orange marks the hydrogens on the two ring carbons that were CH$_2$. The faint sticks are the Vina-docked pose, $0.70$\,\AA{} RMSD from the generated pose and identical for both assignments. The two reconstructions differ in ring bond orders and hydrogen counts, while their heavy-atom coordinates are identical. For this example, the Vina-docked poses and the reported Vina scores are unchanged.
    }
    \label{fig:qual-diene-vs-benzene}
\end{figure}

%% file: ICLR2027/resource/table/appx/appx-mcp-result.tex
\begin{table}[!ht]
    \centering
    \caption{
        \textbf{Macrocyclic peptide design results on the FuncBind benchmark.} Structural metrics compare each generated peptide against its reference. TS is the whole-molecule fingerprint Tanimoto similarity; Residues-TS$\geq$0.5 is the percentage of residues with per-residue similarity at least 0.5; L-RMSD and I-RMSD are the global and interface backbone RMSDs, and Vina Dock is the percentage of peptides docking better than the reference.
    }
    \label{tab:result-mcp}
    \setlength{\tabcolsep}{4pt}
    \renewcommand{\arraystretch}{1.04}
    \setlength{\heavyrulewidth}{1pt}
    \setlength{\lightrulewidth}{0.4pt}
    \resizebox{.98\textwidth}{!}{%
        \begin{tabular}{lcccccc}
            \toprule
            \textbf{Method} & \textbf{TS $\uparrow$} & \textbf{Residues-TS$\geq$0.5\,\% $\uparrow$} & \textbf{L-RMSD $\downarrow$} & \textbf{I-RMSD $\downarrow$} & \textbf{TM-Score $\uparrow$} & \textbf{Vina Dock\,\% $\uparrow$} \\
            \midrule
            RFPeptide & 0.31 & 29 & 12.0 & 3.3 & 0.33 & 8.8 \\
            AfCycDesign & \underline{0.34} & \textbf{29} & \underline{7.6} & 3.7 & \underline{0.33} & \underline{29} \\
            FuncBind & 0.33 & 25 & \textbf{2.6} & \textbf{1.8} & \textbf{0.36} & \textbf{41} \\
            \midrule
            \rowcolor{AffAccent!12} \;+ \oursC{} & - & - & - & - & - & - \\
            \rowcolor{AffAccent!28} \;+ \ours{} & - & - & - & - & - & - \\ 
            \bottomrule
        \end{tabular}
    }
\end{table}